\documentclass[10pt]{article}
\usepackage{amsmath, amssymb}
\usepackage{amsthm}
\usepackage{natbib}
\usepackage[margin=1in]{geometry}
\usepackage{graphicx} 
\usepackage{url}      
\usepackage{cite}     
\usepackage{mathtools} 
\usepackage{hyperref}
\usepackage{comment} 
\hypersetup{
    colorlinks=true,
    linkcolor=cyan,
    filecolor=magenta,
    urlcolor=cyan,
    citecolor=cyan,
}

\usepackage{amsthm}
\usepackage{thmtools}
\usepackage{libertine}
\usepackage{cleveref}
\usepackage{authblk}
\usepackage{xcolor}
\usepackage{tikz}
\usepackage{wrapfig}
\usepackage{pgfplots}
\usepackage{enumitem}
\usepackage{quoting}

\newcommand{\E}{\mathop{\mathbb{E}}}

\theoremstyle{plain}
\newtheorem{theorem}{Theorem}
\newtheorem{lemma}[theorem]{Lemma}
\newtheorem{corollary}[theorem]{Corollary}
\newtheorem{proposition}[theorem]{Proposition}

\theoremstyle{definition}
\newtheorem{definition}[theorem]{Definition}

\theoremstyle{remark}
\newtheorem{remark}[theorem]{Remark}

\AddToHook{cmd/appendix/before}{%
    \crefalias{section}{appendix}%
    \crefalias{subsection}{appendix}
}

\usepackage{bbm}
\usepackage{mathrsfs}

\newcommand{\CX}{\mathcal{X}}
\newcommand{\CY}{\mathcal{Y}}
\newcommand{\CC}{\mathcal{C}}
\newcommand{\CD}{\mathcal{D}}

\newcommand{\CI}{\mathcal{I}}

\newcommand{\CP}{\mathcal{P}}

\newcommand{\CF}{\mathcal{F}}
\newcommand{\CA}{\mathcal{A}}

\newcommand{\CJ}{\mathcal{J}}

\newcommand{\N}{\mathbb{N}}
\newcommand{\R}{\mathbb{R}}
\newcommand{\Q}{\mathbb{Q}}

\newcommand{\eps}{\varepsilon}

\newcommand{\predictors}{\CF_\CX}
\newcommand{\x}{\mathbf{x}}
\newcommand{\y}{\mathbf{y}}

\newcommand{\normnoarg}[1]{\ell_{#1}}
\newcommand{\norm}[2]{\| #2 \|_{#1}}
\newcommand{\metric}[3]{\ell_{#1}(#2, #3)}
\newcommand{\metricres}[4]{\ell_{#1}|_{#2}(#3, #4)}

\renewcommand{\E}{\mathop{\mathbb{E}\mbox{}}}
\renewcommand{\Pr}{\mathop{\mathbb{P}\mbox{}}}

\newcommand{\on}[1]{\operatorname{#1}}
\newcommand{\wt}[1]{\widetilde{#1}}
\newcommand{\calset}{\on{cal}}
\newcommand{\mcalset}{\on{mcal}}
\newcommand{\accset}{\on{acc}}
\newcommand{\maccset}{\on{macc}}
\newcommand{\intcl}{\on{\CI}}

\newcommand{\dCE}{\on{dCE}}
\newcommand{\dMC}{\on{dMC}}
\newcommand{\dIMC}{\on{dIMC}}
\newcommand{\cdMC}{\wt{\dMC}}
\newcommand{\dMA}{\on{dMA}}
\newcommand{\wdMC}{\on{wdMC}}
\newcommand{\dAE}{\on{bias}} 
\newcommand{\wdMA}{\on{wdMA}} 

\newcommand{\dMCE}[2]{\dMC_{#1, #2}}
\newcommand{\dIME}[2]{\dIMC_{#1, #2}}
\newcommand{\dME}[2]{\cdMC_{#1, #2}}
\newcommand{\dC}[2]{\dCE_{#1, #2}}

\newcommand\restr[2]{{
  {#1}|_{#2}
  }}

\newcommand{\ul}{\underline}

\newcommand{\poly}{\mathsf{poly}}
\renewcommand{\exp}{\mathsf{exp}}
\newcommand{\polys}[1]{\CP_{#1}}

\title{
Auditability and the Landscape of Distance to Multicalibration
}

\newcommand{\email}[1]{\textsf{#1}}

\newcommand{\USC}{University of Southern California}

\author{Nathan Derhake}
\author{Siddartha Devic}
\author{Dutch Hansen}
\author{Kuan Liu}
\author{Vatsal Sharan}
\affil{\USC}
\affil{\email{\{derhake, devic, jmhansen, liukuan, vsharan\}@usc.edu}}

\date{\today}

\begin{document}
\date{}

\maketitle

\begin{abstract}
    Calibration is a critical property for establishing the trustworthiness of predictors that provide uncertainty estimates.
    Multicalibration is a strengthening of calibration which requires that predictors be calibrated on a potentially overlapping collection of subsets of the domain.
    As multicalibration grows in popularity with practitioners, an essential question is: \emph{how do we measure how multicalibrated a predictor is?}
    \citet{blasiok2022unifying} considered this question for standard calibration by introducing the \emph{distance to calibration} framework (dCE) to understand how calibration metrics relate to each other and the ground truth.
    Building on the dCE framework, we consider the auditability of the \emph{distance to multicalibration} of a predictor $f$.
    
    We begin by considering what are perhaps the two most natural generalizations of dCE to multiple subgroups: worst group dCE (wdMC), and distance to multicalibration (dMC).
    Using wdMC and dMC as a guiding path, we argue that there are two essential properties of any multicalibration error metric: 1) the metric should capture how much $f$ would need to be modified in order to be perfectly multicalibrated; and 2) the metric should be auditable in an information theoretic sense (i.e., with some finite sample complexity).
    We show that wdMC and dMC each fail to satisfy one of these two properties, and that similar barriers arise when considering the auditability of general distance to multigroup fairness notions (\textit{e.g.} multiaccuracy or low-degree multicalibration).
    We then propose two (equivalent) multicalibration metrics which do satisfy these requirements: 1) a continuized variant of dMC; and 2) a distance to \emph{intersection} multicalibration, which leans on intersectional fairness desiderata.
    
    Along the way, we shed light on the \emph{loss-landscape} of distance to multicalibration and the geometry of the set of perfectly multicalibrated predictors.
    We also demonstrate that the loss surface of any metric which captures how much $f$ would need to be modified to be perfectly multicalibrated often satisfies a \emph{local minima are global minima} property.
    Our findings may have implications for the development of stronger multicalibration algorithms, as well as multicalibration auditing more generally.  
\end{abstract}
\thispagestyle{empty}
\newpage

\tableofcontents
\thispagestyle{empty}

\newpage

\setcounter{page}{1}

\section{Introduction}
Model calibration is an essential requirement in settings where trustworthy machine-learned predictions are used \citep{dahabreh2017review, beque2017approaches}.
Calibration requires that among all samples given score $p \in [0,1]$ by a predictor $f$, exactly a $p$-fraction of those samples have positive label. However, an arbitrary predictor $f$ --- such as a neural network --- may not output the exact same prediction $p$ more than a handful of times, which makes directly measuring the true calibration error difficult.
Instead, for much of the modern history of machine and deep learning, the (binned) \emph{expected calibration error} (ECE) has been the most popular metric used to measure calibration \citep{guo2017calibration, wang2023calibration}. Nonetheless, numerous works have pointed to shortcomings of ECE \citep{blasiok2022unifying, kakade2004deterministic, brocker2009reliability, kull2015novel}.
For example, a small perturbation to the predictor $f$ can wildly change the ECE of $f$ with respect to a ground truth distribution $\CD$.
In response to this, the calibration community has developed a dizzying array of alternative metrics such as kernel calibration error \citep[kCE]{marx2023calibration, kumar2018trainable}, $\beta$-calibration error \citep{pmlr-v54-kull17a}, Smooth ECE \citep[smECE]{blasiok2023smooth}, and more \citep{naeini2015obtaining, niculescu2005predicting}.

In order to provide firmer theoretical grounding for alternatives to ECE, \citet{blasiok2022unifying} introduced an elegant unified framework for understanding the usefulness of calibration metrics.
They posit that, in lieu of the ultimate goal of obtaining \emph{better calibrated} predictors, the underlying desiderata for calibration error metrics should be: \emph{low error implies a predictor can easily be post-processed to be calibrated, and high-error the opposite.}
They do this by introducing the \emph{distance to calibration error} (dCE) of a predictor $f$, which measures the minimal distance $f$ would need to change (in a suitable $\ell_1$ metric space of predictors) in order to obtain a calibrated predictor.\footnote{This is non-trivial in part because the set of calibrated predictors is non-convex.}
Different calibration metrics such as ECE or kCE are then understood as approximations to the dCE of $f$ with varying degrees of usefulness.

Simultaneously and relatedly, model \emph{multicalibration} has emerged as a recent quantity and tool from the algorithmic fairness literature \citep{hebert2018multicalibration}.
Put simply, multicalibration posits that a predictor should be calibrated on a potentially rich, overlapping collection of subgroups of the data distribution. 
For example, a risk-predictor used by a bank to inform lending decisions \citep{beque2017approaches} should be calibrated when conditioned on legally protected subgroups of the population such as particular races, residents from certain geographic area, individuals older than 67, etc. \citep{eco}.

Although multicalibration originated as a tool within the fairness community, techniques and tools from the broader \emph{multi-group} viewpoint have recently received attention in the theoretical computer science community more generally \citep{gopalan2022low, gopalan2023characterizing, shabat2020sample}.
In particular, the multi-group literature has been used to obtain guarantees in \emph{omniprediction} \citep{gopalan2021omnipredictors, gollakota2024agnostically, okoroafor2025near}, robustness to distributions shift \citep{hu2024multigroup, wu2024bridging}, loss prediction \citep{gollakota2025loss}, and more \citep{devic2024stability, haghtalab2024unifying}.
On the empirical side, these algorithms  have been found to be useful in improving the calibration of text and image classifiers \citep{hansen2024multicalibration, bharti2025multiaccuracy, wu2024bridging}, and even LLMs \citep{detommaso2024multicalibration, liu2024multi}.

As we start applying and measuring the effectiveness of multi-group post-processing notions in the wild, the very question of \emph{measuring} multi-group calibration has become salient.
Practitioners applying standard (non-\emph{multi}) calibration metrics have a relatively good understanding of when different metrics such as ECE or smECE should be used based on the context and desired measurement properties \citep{blasiok2022unifying, blasiok2023smooth}. Practitioners of multicalibration, however, do not yet have much guidance on choice of measurement metrics.
For example, prior proposed theory utilizes worst group, discretized $\ell_1$ calibration errors \citep{hebert2018multicalibration, haghtalab2024unifying}.
That is, they consider the subgroup of the domain with the highest discretized $\ell_1$ calibration error as the multicalibration metric of interest.
The empirically-minded works of \citet{hansen2024multicalibration} and \citet{jin2025discretization} utilize worst-group unweighted smECE, while even others use kernel calibration error notions \citep{long2025kernel}.
Given the smorgasbord of options directly extending from the standard calibration literature, our goal is to provide firm theoretical footing to practical advances in multicalibration.
In particular, we ask:
\begin{center}\emph{
    \ (1) When can we measure the distance of a predictor to the nearest multicalibrated predictor?\\ 
    (2) More broadly, when are distance to multi-group calibration notions efficiently auditable?
    }
\end{center}
The first question (1) seeks to understand if the \emph{distance} to calibration notion of \citet{blasiok2022unifying} can directly be extended to multicalibration.
A positive result could potentially provide grounded recommendations to practitioners measuring and utilizing multicalibration algorithms in practice.
In addition, answering the question in any capacity also requires building a deeper understanding of the \emph{geometry} of the set of multicalibrated predictors.
This is because any distance to multicalibration notion is fundamentally a metric which measures the distance from a predictor $f$ to the \emph{set} of multicalibrated predictors for a ground truth distribution.
Natural follow-up questions abound: What does this set look like? Is measuring the distance to this set feasible, and does the set have nice properties relative to the ground truth distribution?

In addition, answering (1) may help uncover properties of the \emph{loss surface} of multicalibration in the metric space of all predictors.
Many multicalibration post-processing algorithms happen to be boosting algorithms which combine and collect predictors into complicated sequences of updates \citep{hebert2018multicalibration, haghtalab2024unifying}. Can other types of algorithms which operate more like gradient descent on the loss surface of predictors exist?
Boosting algorithms such as the one described in \citet{hebert2018multicalibration} can be viewed as a form of \emph{functional} gradient descent. We are instead asking: are there algorithms which apply gradient updates directly on the predictor $f$ (instead of some surrogate space), decreasing its multicalibration error at each step?

The second question (2) seeks to build upon three multi-group notions related to multicalibration: low-degree multicalibration \citep{gopalan2022low}, multiaccuracy \citep{kim2019multiaccuracy, long2025kernel}, and calibrated multiaccuracy \citep{casacuberta2025global, gopalan2023lossmin, okoroafor2025near}.
We may be able to extend distance to multicalibration to these weaker multi-group fairness notions, unlocking many possible questions about the geometry and auditability of each of these special sets of predictors.
Directions (1) and (2) are both important since they formally define the multi-group calibration evaluation problem, and may eventually allow for the development of more powerful algorithms for multi-group post-processing in a variety of domains.

\subsection{Notation and Background}
\label{subsec:cal-mcal-notation}
Fix a feature space $\CX$ with finite (but large) cardinality $|\CX| = n$, and define the label space as $\CY = \{0,1\}$. 
Define a predictor as a function $f:S \to [0,1]$ for an arbitrary set $S\subseteq \CX$.
Let $\predictors = [0,1]^\CX$ denote the set of all predictors on $\CX$.
We sample example-outcome pairs $(\x, \y)$ i.i.d. according to joint distributions $\CD$ over $\CX \times \CY$.
Equivalently, we can view this as first sampling $\x$ from a marginal distribution $\CD_\x$ over the domain $\CX$, and then sampling $\y \sim \text{Bernoulli}(p^*(\x))$ for some ground truth $p^* \in \predictors$.
We further assume that each $\CD_\CX$ is supported on all of $\CX$.
We will denote the set of all such distributions as $\Delta(\CX \times \CY)$.

\begin{definition}[Calibration]
    A predictor $f : \CX \to [0,1]$ is \emph{perfectly calibrated} with respect to a distribution $\CD$ over $\CX \times \CY$ if for every $v \in \text{Im}(f)$, we have: \begin{align*}
        \E\limits_{(\x, \y) \sim \CD} \Big[ \y \ | f(\x) = v\Big]  = v.
    \end{align*}
    We denote the set of all such predictors $\calset(\CD)$.
\end{definition}
Calibration is a well studied property of predictors which output uncertainty estimates in $[0,1]$. 
Calibrated predictions are useful in a broad variety of settings spanning online learning \citep{shenalgorithms}, collaboration \citep{collina2025collaborative}, conformal prediction \citep{jung2022batch}, and classification with neural networks \citep{guo2017calibration}.
The task of \emph{measuring} calibration has given rise to its own line of research (see \citet{silva2023classifier} for a survey).
In particular, a variety of calibration metrics now exist: kCE \citep{marx2023calibration}, smECE \citep{blasiok2023smooth}, ECE, etc.
In order to derive an understanding of this veritable smorgasbord, \citet{blasiok2022unifying} define the notion of \emph{distance to calibration error} ($\dCE$) of a predictor $f$.

Before proceeding, we establish notation. For a distribution $\CD$ over $\CX \times \CY$ and $f, g: \CX \to \R$, we let
\begin{align*}
    \norm{p}{f} := \left( \E_{\x \sim \CD_\x} \Big[ |f(\x)|^p \Big] \right)^{1/p}
    \quad
    \text{and} 
    \quad 
    \metric{p}{f}{g} := \norm{p}{f-g}.
\end{align*}

Given $H \subseteq [0,1]^\CX$, we define the $\normnoarg{p}$-distance to $H$ as follows.
\begin{align*}
    \metric{p}{f}{H} := \inf_{h \in H} \metric{p}{f}{h}
\end{align*}
Notice that in our context, the $\normnoarg{1}$ metric and norm differ slightly from the traditional norms and metrics on $\R^{|\CX|}$ because we weigh each coordinate by the marginal distribution $\CD_\x$.

\begin{definition}[Distance to Calibration Error \citep{blasiok2022unifying}]
\label{def:dCE}
    For $f : \CX \to [0,1]$ and a distribution $\CD$ over $\CX \times \CY$, we define
\begin{align*}
    \dCE_{\CD}(f) := \metric{1}{f}{\calset(\CD)} = \inf_{g \in \calset(\CD)} \metric{1}{f}{g}.
\end{align*}
\end{definition}

\noindent Note that $\calset(\CD)$ is finite by \Cref{lemma:calset-finite} and hence closed under the topology induced by the $\normnoarg{p}$ metric for any $p \geq 1$; we can therefore replace the $\inf$ with a $\min$ in the definition of $\dCE$.\footnote{Finiteness of $\calset(\CD)$ was also used and pointed out in \citep{blasiok2022unifying}.} As in \citet{blasiok2022unifying}, our presentation focuses on the $\normnoarg{1}$ metric.

The definition of $\dCE$ is based on measuring how far a predictor $f$ is from \emph{perfect} calibration.
This allows for \citet{blasiok2022unifying} to understand and characterize calibration metrics like ECE which are used in practice.
In particular, they show that ECE can be understood to give an \emph{upper} bound on $\dCE$, which implies that if ECE is small, the predictor is truly close to being perfectly calibrated (a soundness condition on the distance to the set of perfectly calibrated predictors).
On the other hand, large ECE does \emph{not} guarantee that the predictor is actually far from being calibrated (i.e., does not satisfy completeness).
Similar completeness and soundness guarantees can be derived for other proposed calibration metrics.

\paragraph{Generalizing to Multicalibration.} 
Let a collection $\CC = \{S_1, \dots, S_k\}$ with $S_i \subseteq \CX$ be given. 
Multicalibration is a \emph{strengthening} of calibration which requires a predictor to be simultaneously calibrated when conditioned on each $S_i \in \CC$.
To formalize this within the notation of $\dCE$, we require restrictions of predictors, distributions, and metrics. It will be useful to generalize some of the objects already introduced.
For $f : \CX \to [0,1]$ and $S \subset \CX$, we let $\restr{\CD}{S}$ denote the distribution $\CD$ conditioned on $\x \in S$; this is equivalent to conditioning $\CD_\x$ on $\x \in S$ and drawing $\y$ according to ground truth $\restr{p^*}{S}$. We will sometimes use a conditional version of the $\normnoarg{p}$ norm and metric:
\begin{align*}
    \norm{p,S}{f} := \left( \E_{\x \sim \restr{\CD}{S}} \Big[ |f(\x)|^p\Big] \right)^{1/p}
    \quad
    \text{and} 
    \quad 
    \metricres{p}{S}{f}{g} := \norm{p, S}{f-g}.
\end{align*}

We slightly abuse notation by writing $\metricres{1}{S}{f}{g}$ when the domain of $f$ and $g$ contain $S$.
We will sometimes say that a predictor $f \in \predictors$ is calibrated on (or with respect to) $S$; formally, we mean that $\restr{f}{S}$ is calibrated \textit{w.r.t.} $\restr{\CD}{S}$.

\begin{definition}[Multicalibration]
\label{def:perfect-multicalibration}
    Given a collection $\CC = \{ S_1 , \dots, S_k\}$ with each $S_i \subseteq \CX$, a predictor $f : \CX \to [0,1]$ is said to be \emph{perfectly $\CC$-multicalibrated} (or just multicalibrated) if $f$ is perfectly calibrated with respect to $S_i$ for all $i \in [k]$. We denote the set of such predictors as $\mcalset_\CC(\CD)$.
\end{definition}
We remark that $p^*$ is always in the set $\mcalset_\CC(\CD)$ for the distribution $\CD$ that it defines (the ground truth is perfectly calibrated on any subset).
However, $\mcalset_\CC(\CD)$ is most interesting when it contains predictors \emph{other} than $p^*$.
Lastly, note that multicalibration says nothing about how $f$ needs to behave on elements of $\CX$ not covered by $\CC$. 
Even even when $\CC$ does cover $\CX$, a predictor need not be calibrated over $\CX$ to be perfectly $\CC$-multicalibrated. 
It is therefore without loss of generality to assume that $\CC$ is a cover of $\CX$, \textit{i.e.} $\bigcup_i S_i = \CX$. 
We will nevertheless stipulate when this condition is required.

\begin{figure}
    \centering
    \includegraphics[width=0.48\linewidth]{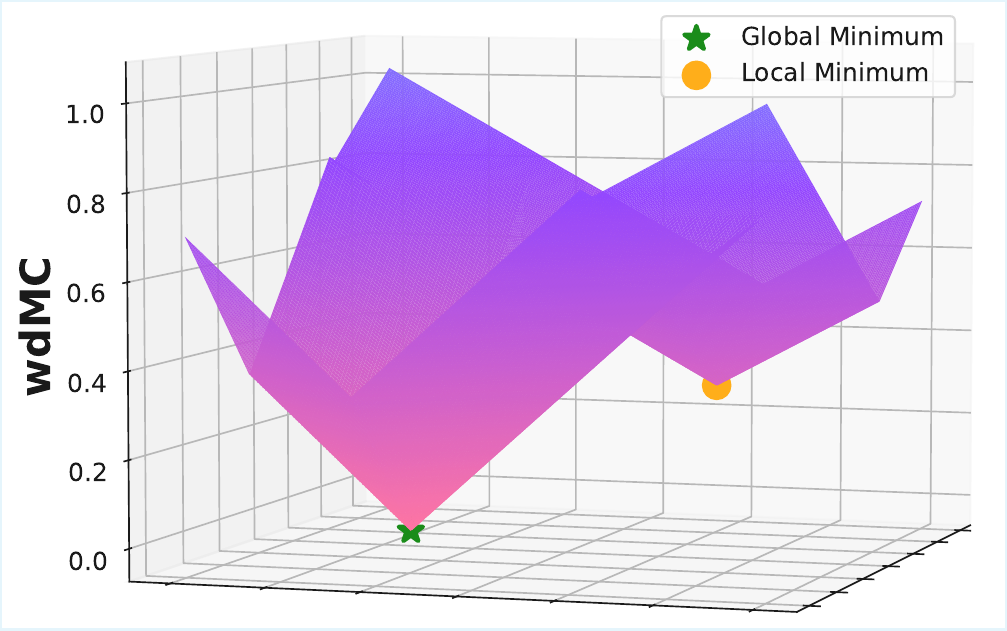}
    \includegraphics[width=0.43\linewidth]{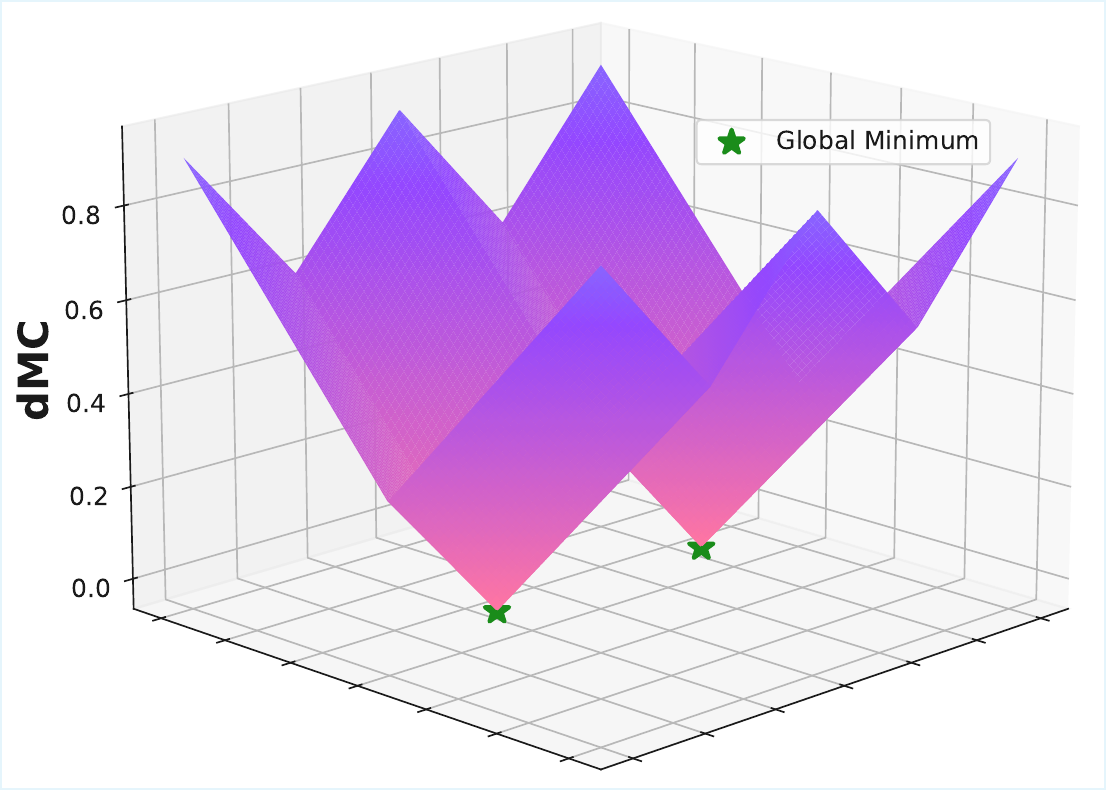}
    \vspace{-5pt}
    \caption{(\textbf{Left}): \Cref{lemma:improving-wdmc} demonstrates that the worst-group distance to calibration error function ($\wdMC$) contains \emph{local} minima which are not global minima in the space of all predictors. 
    That is, there are predictors $f$ which appear \emph{locally not improvable} in terms of worst-group calibration error, but which can be improved with a large enough change. (\textbf{Right}): Conversely, \Cref{thm:dMC-global-minima} demonstrates that all local minima are global for $\dMC$, even though the loss landscape may be non-convex.}
    \label{fig:dMC-loss-landscape}
\end{figure}

\subsection{Summary of Contributions}

\paragraph{A Practical Error Notion.}
Motivated by practitioners \citep{hansen2024multicalibration, jin2025discretization}, we begin by defining a straightforward ways to measure the distance of $f$ to multicalibration.
Let a subgroup collection $\CC$ be given.
In \Cref{subsec:measuring-multical}, we first introduce the worst-group distance to calibration error, denoted by $\wdMC_\CC(f)$.
This is defined as the maximum over all subgroups of how far $f$ is from being calibrated in terms of $\dCE$, when restricting $f$ and the distribution $\CD$ to only that subgroup.
Importantly, $\wdMC_\CC$ is also weighed by the ``size'' (mass) of each subgroup $S_i$ in the marginal distribution $\CD_\x$.
This weighing insures that a vanishingly small group will not have disproportionate impact on the multicalibration error, which intuitively would make it difficult to audit.\footnote{Consider a subgroup which appears with exponentially small probability in $\CD_\x$, but introduces a high calibration error for $f$. Such a subgroup would not be detectable in a polynomial number of samples, yet would maximize an unweighted $\wdMC$.}

We then discuss the \emph{loss-landscape} viewpoint of multicalibration error notions.
The metric $\wdMC: \predictors \to \R_{\geq0}$ maps from the space of predictors to real numbers; in particular, it defines a \emph{loss surface}.
Multicalibration algorithms can be viewed as traversing this loss surface.
In order for gradient-based multicalibration algorithms to be effective, a reasonable requirement on the loss surface is that local minima are global minima. If this were not the case, then any such algorithm could get stuck in a sub-optimal local minima and not be able to improve the multicalibration error.
Existing boosting-based multicalibration algorithms do not have this issue, and can drive the objective functions as low as desired with enough samples \citep{haghtalab2024unifying, hebert2018multicalibration}.

Unfortunately, in \Cref{lemma:improving-wdmc} we demonstrate that low $\wdMC$ \emph{does not} imply that a predictor is close to being perfectly multicalibrated.
In particular, we exhibit a multicalibration auditing instance --- a distribution $\CD$, subgroup collection $\CC$, and predictor $f$ to be audited --- for which $\wdMC(f)=\eps$, but $f$ is at a local minima which is not global.
That is, the closest predictor $\tilde{f}$ which has $\wdMC(\tilde{f}) < \eps / 2$ is $\Omega(1)$ distance away from $f$.
This is illustrated in \Cref{fig:dMC-loss-landscape}.

\paragraph{Local and Global Minima.} 
Using $\wdMC$, we have developed one natural requirement for any multicalibration error metric: all local minima should be global minima.
To directly address this, in \Cref{sec:dMC} we introduce another multicalibration error metric: \emph{distance to multicalibration} (denoted $\dMC_{\CC}(f)$).
This metric is defined exactly as the distance from $f$ to its projection onto the set of perfectly multicalibrated predictors defined in \Cref{def:perfect-multicalibration}.
In \Cref{thm:dMC-global-minima}, we show that $\dMC$ does indeed satisfy this property, since it is a function which computes the distance of $f$ to a set. The set of perfectly multicalibrated predictors is actually a union of finitely many affine subspaces of $\predictors$, and since $\dMC$ is defined with the $\normnoarg{1}$ metric, the loss landscape itself turns out to be piecewise-linear.

Next, we turn to auditing or computing $\dMC_\CC(f)$.
Throughout most of our presentation, we are concerned with \emph{information-theoretic} auditability, which asks whether we can compute the metric or quantity of interest with any finite number of samples.\footnote{We also sometimes consider statistical auditability, (i.e., can we compute the metric in a number of samples polynomial in $|\CC| = k$?) and will explicitly state when this is the case.}
In \Cref{prop:no-audit-dMC}, we demonstrate a pathology of $\dMC$ which makes it impossible to audit in this information theoretic sense.
We construct a simple instance for which determining whether $\dMC_\CC(f) = 0$ or  $\dMC_\CC(f) > C$ for a constant $C>0.2$ requires finding the \emph{exact} value of the ground truth predictor $p^*(x)$ on some point $x \in \CX$.
This is impossible with any (finite) number of samples, and hence, $\dMC$ is in general, inauditable.
Near the end of \Cref{sec:dMC} and in \Cref{appx:generalizing-dMC}, we show that similar instances exist for distance to low-degree multicalibration \citep{gopalan2022low} and calibrated multiaccuracy \citep{casacuberta2025global}, two recently proposed relaxations of full multicalibration which may be of interest to the community.

\paragraph{Stepping Towards Auditability.}
The pathology exhibited in \Cref{prop:no-audit-dMC} arises precisely because a minuscule change in the ground truth $p^*(x)$ on one particular point $x$ can cause the $\dMC_\CC(f)$ to rapidly change by a constant.
That is, $\dMC_\CC(f)$ is not \emph{Lipschitz} in the ground truth predictor $p^*$.
Armed with this intuition, we seek new multicalibration error notions which can simultaneously satisfy two conditions:
\begin{enumerate}
    \item[(1)] All local minima of the loss landscape induced by the metric are global minima;
    \item[(2)] When holding $f$, $\CD$, and $\CC$ fixed, the metric is Lipschitz in the ground truth function $p^*$.
\end{enumerate}
We note that (1) can be viewed as a form of a soundness condition relative to the distance of $f$ to the set of perfectly multicalibrated predictors.
On the other hand, (2) provides a \emph{necessary} condition for information-theoretic auditability of the metric using samples from $\CD$.
If (2) was violated by non-Lipschitzness or $C$-Lipschitzness for some very large $C$, then arbitrarily small changes in the ground truth $p^*$ could induce very large changes in the multicalibration metric.
This would then require approximating $p^*$ arbitrarily well in order to get any good estimate on the metric.

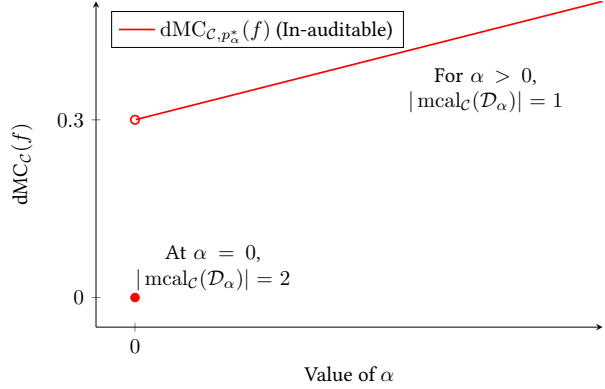
\begin{figure}
    \centering
    \begin{tikzpicture}[scale=0.8]
    \begin{axis}[
        xlabel={Value of $\alpha$},
        ylabel={$\text{dMC}_{\mathcal{C}}(f)$},
        xmin=-0.05, xmax=0.6,
        ymin=-0.05, ymax=0.5,
        axis lines=left,
        xtick={0},
        xticklabels={$0$},
        ytick={0, 0.3},
        legend pos=north west,
        legend cell align={left},
        width=10cm,
        height=7cm,
    ]


    \addplot[
        domain=0.001:0.6, 
        samples=100,
        color=red,
        thick,
    ] {0.3 + x/3};
    \addlegendentry{$\dMC_{\mathcal{C}, p^*_\alpha}(f)$ (In-auditable)}

    \addplot[only marks, mark=*, color=red, mark size=2pt] coordinates {(0, 0)};

    \addplot[only marks, mark=o, color=red, mark size=2pt, thick] coordinates {(0, 0.3)};

    
    \node (upper_text) at (axis cs:0.45, 0.35) [text width=2.8cm, align=center] {For $\alpha > 0$, \\ $|\mcalset_{\mathcal{C}}(\CD_\alpha)|=1$};

    \node (lower_text) at (axis cs:0.1, 0.05) [text width=2.8cm, align=center] {At $\alpha = 0$, \\ $|\mcalset_{\mathcal{C}}(\CD_\alpha)|=2$};

    \end{axis}
\end{tikzpicture}
    \caption{
    An illustration showing that the distance to multicalibration error ($\dMC$) of a predictor $f$ can be a non-continuous function of the ground truth predictor $p^*$.
    Here, $p^*$ is parameterized by a parameter $\alpha$.
    For small enough $\alpha>0$, any finite number of samples is not enough to determine whether $\alpha=0$ or $\alpha \neq 0$. 
    Since $\dMC_\CC(f)=0$ if $\alpha=0$, and $\dMC_\CC(f) \geq 0.3$ otherwise, this demonstrates an information-theoretic barrier to auditing $\dMC$. 
    This happens because the number of multicalibrated predictors $|\mcalset_\CC(p^*)|$ grows to two at $\alpha=0$, but for an $\alpha>0$, there is only a single multicalibrated predictor (given by $p^*$).
    Since $\dMC_\CC(f)$ measures the distance of $f$ to $\mcalset_\CC(p^*)$, it experiences an induced discontinuity.
    See \Cref{prop:no-audit-dMC} for details.}
    \label{fig:dmc-discont}
\end{figure}

With these two basic requirements in hand, in \Cref{subsec:continuized-dMC} we examine why the non-Lipschitzness used in the instance of \Cref{prop:no-audit-dMC} occurs.
Intuitively, there is a measure-zero set of ``rogue'' ground truth predictors which are the points of discontinuity of $\dMC_\CC(f)$.
This is illustrated in \Cref{fig:dmc-discont}, where a point-wise discontinuity causes $\dMC$ to change by a constant.
To rid ourselves of this, we introduce a \emph{continuized} distance to multicalibration error metric, which we denote by $\cdMC$ (\Cref{def:cdMC}).
At a high level, $\cdMC$ is defined by carefully \emph{smoothing} the $\dMC$ over a sequence of local neighborhoods around the ground truth predictor $p^*$.

\paragraph{Suitable Metric(s).}
In \Cref{prop:cdmc-lipschitz}, we show that $\cdMC_{\CC}(f)$ satisfies properties (1) and (2).
Our proof of this is presented in \Cref{sec:proof-of-cdmc-lipschitz} with lemmas developed in \Cref{sec:dIME}.
Interestingly, the proof relies on yet another multicalibration error metric: the distance to \emph{intersection} multicalibration error, denoted by $\dIMC_{\CC}(f)$.
This metric is simply defined as $\dIMC_{\CC}(f) := \dMC_{\CI(\CC)}(f)$, where $\CI(\CC)$ is the set of all intersections of sets from $\CC$.
Measuring notions of fairness for \emph{intersections} of subgroups is an important and difficult problem in the fairness community more broadly \citep{gohar2023survey, himmelreich2024intersectionality}, and hence, $\dIMC$ may be of independent interest.

In \Cref{thm:dimc-continuized-dmc}, we demonstrate that for all subgroup collections $\CC$ (which are covers of the domain $\CX$), distributions $\CD$, and predictors $f$, we have that $\dIMC_\CC(f) = \cdMC_\CC(f)$.
This reveals the surprising fact that \emph{smoothing} $\dMC$ by removing the point-wise discontinuities is \emph{equivalent} to measuring a strong, intersectional distance to multicalibration error notion.
Using this fact, we can think of $\cdMC_\CC(f)$ as exactly the $\normnoarg{1}$ distance from $f$ to the nearest predictor which is multicalibrated over $\CC$ \emph{and} all intersections of groups in $\CC$.

In \Cref{lemma:PIE-dIMC-definition}, we show that there is a \emph{disjoint} partition $\CJ(\CC)$ of the domain $\CX$ (defined in \Cref{def:partition-generated}) for which $\dMC_{\CJ(\CC)}(f) = \dMC_{\CI(\CC)}(f) = \dIMC_{\CC}(f)$.
Why is such a disjoint partition helpful? 
In \Cref{lemma:dMC-Lipschitz-if-groups-disjoint}, we show that if the collection $\CC$ is disjoint, then $\dMC_\CC(f)$ reduces to a weighted sum of $\dCE$ (\Cref{def:dCE}) with respect to each subset in the collection $\CC$.
It follows that $\dIMC_\CC(f)$ is a weighted sum of $\dCE$ with respect to each set in $\CJ(\CC)$.
Since $\dCE$ satisfies property (2) in that it is Lipschitz with respect to the ground truth function $p^*$ (\Cref{lemma:dCE-Lipschitz}), the disjointness of $\CJ(\CC)$ implies this carries through to $\dIMC$.
Using this interpretation, in \Cref{prop:dimc-lipschitz} we show that $\dIMC_\CC(f)$ --- equivalently, $\cdMC_\CC(f)$ --- satisfies both desiderata (1) and (2) of a multicalibration error metric, completing the proof of \Cref{prop:cdmc-lipschitz}.

\paragraph{(In)auditability of $\cdMC$ and $\dIMC$.}
In \Cref{subsec:auditability-cdMC}, we discuss auditability of $\dIMC$ (equivalently, $\cdMC$) using the perspective developed in \Cref{sec:dIME}.
In particular, we make use of the fact that $\dIMC$ is equivalent to a weighted average of $\dCE$ over the disjoint partition of $\CX$ defined by $\CJ(\CC)$.

By \emph{statistical} auditability of $\dCE$ (\citet{blasiok2022unifying}), $\dIMC$ should be statistically feasible to audit.
Nonetheless, the partition $\CJ(\CC)$ which we chose can turn out to have too many sets.
In particular, it could be the case that $|\CJ(\CC)| = \Theta(2^{|\CC|})$, meaning that auditing $\dIMC$ could require checking the $\dCE$ of exponentially many restrictions of $f$.
Since a majority of the density in $\CD_\x$ could be placed on sets that we never sample, in \Cref{prop:no-audit-dimc} we demonstrate an example where we cannot distinguish whether $\dIMC_\CC(f) = 0$ or $\dIMC_\CC(f) = 0.5$ using $\poly(|\CC|)$ samples.
On the flip side, \Cref{prop:dIMC-auditing} shows that when $\CJ(\CC)$ satisfies certain ``nice'' conditions --- such as sufficient mass on a small number of sets in $\CJ(\CC)$, or $\CC$ itself being relatively small or constant --- there is hope for a statistical (and computational) auditability result of $\dIMC$ in $\poly(|\CC|)$ samples.
This is noteworthy because in the practice of fair machine learning, we often expect there to be a small number of subgroups relative to the size of the domain $\CX$.

\paragraph{Distance to Multiaccuracy.}
Given the nature of the results presented in \Cref{subsec:auditability-cdMC}, in \Cref{sec:dMA-discussion} we ask: \emph{Is there a natural relaxation of full multicalibration for which a ``distance to'' notion is statistically auditable?}
To address this, we define and analyze the \emph{weakest} relaxation of multicalibration: multiaccuracy \citep{kim2019multiaccuracy}.
Multiaccuracy asks that a predictor $f$ be \emph{unbiased} over a collection $\CC$.
In \Cref{subsec:defining-accuracy-notions}, we begin by introducing $\wdMA_\CC(f)$ as a ``worst-group bias'' of $f$ on a collection $\CC$ (this is defined analogously to $\wdMC$ for multicalibration).
We also define $\dMA_\CC(f)$, which measures the distance of a predictor $f$ to closest \emph{perfectly multiaccurate} predictor (analogous to $\dMC$).

In \Cref{subsec:measuring-multiacc}, we demonstrate that the loss landscapes of both functions are \emph{convex}, which implies that all local minima are global minima.
Interestingly, we then show that $\wdMA$ and $\dMA$ are different quantities, and $\wdMA_\CC(f)$ fails to satisfy that ``low error implies $f$ is close to a multiaccurate predictor''.
In particular, \Cref{lemma:improving-wdma} exhibits an instance for which 
a predictor $f$ has $\wdMA_\CC(f) = \eps$, but the distance $\metric{1}{f}{\tilde f}$ from the predictor $f$ to \emph{any} predictor $\tilde f$ with $\wdMA_\CC(\tilde f) \leq \eps/6$ is $\Omega(\eps \cdot d^{|\CC|})$, where the constant $d>1$ is the square root of the golden ratio.
This emerges because the construction used in the proof uses a subgroup collection based on the Fibonacci recurrence.
By highlighting this undesirable property of $\wdMA$, \Cref{lemma:improving-wdma} further motivates the definition of $\dMA$.

Lastly, in \Cref{subsec:auditing-dma} we close by providing a linear program for computing $\dMA_\CC(f)$ exactly given perfect information on the average value of $p^*$ over each subgroup.
The program has $2\cdot |\CX|$ variables (to model the closest multiaccurate predictor to $f$), and $O(n+k)$ constraints (to ensure that the modeled predictor is unbiased on each group).
We leave open the question of statistically and computationally feasible auditing of $\dMA$.

\subsection{Related Work}
We remark that we are not the first to study distance to multicalibration error.
Given that it is a natural generalization of distance to \emph{calibration} error ($\dCE$, \citet{blasiok2022unifying}), the recent work \citet{dwork2024fairness} shows that achieving low distance to multicalibration in an online setting may be intractable.
This may translate to impossiblity results in the offline setting via online-to-batch reductions.
In general, we are interested in understanding whether we can even \emph{measure} notions of distance to multicalibration error, let alone achieve them.
This distinction between our work is discussed further in \Cref{rmk:auditability-of-dMC} within \Cref{sec:dMC}.

There is also recent empirical work on measuring multicalibration in practice.
In \citet{guy2025measuring}, the authors propose using a worst-group calibration error notion which weighs groups by the \emph{signal-to-noise} ratio of the predictor restricted to that group.
They demonstrate the superiority of their measurement method in a variety of practical settings.
Their framing stands in parallel to how we define $\wdMC$, which weighs groups by their size.
Indeed, \citep{guy2025measuring} emphasize the importance of not having multicalibration metrics being thrown off by a very small group for which $f$ is poorly calibrated or inaccurate on.
This further aligns with the results of \citet{hansen2024multicalibration}, who found that the worst-group calibration error is often determined by small groups in the data distribution.

\section{A Loss Landscape Approach to Multicalibration Measures}
In \Cref{subsec:measuring-multical}, we formally define the worst-group distance to calibration error notion which has been used in prior work ($\wdMC$).
In \Cref{sec:dMC}, we introduce our \emph{distance to multicalibration} notion, $\dMC$, which satisfies two properties that we argue are important for any multicalibration metric.
Finally, we close with an obstruction to auditing $\dMC$, and how this relates to auditability of the standard (non-multi) calibration metric of $\dCE$ \citep{blasiok2022unifying}.

\subsection{Measuring Multicalibration via Worst-Group Distance to Calibration}
\label{subsec:measuring-multical}

The most common way to measure multicalibration in both theory \citep{hebert2018multicalibration} and practice \citep{hansen2024multicalibration} is \emph{worst-group} calibration error.
To model this within the framework of \citet{blasiok2022unifying}, we consider an analogous definition: \emph{worst-group} distance to calibration, denoted as $\wdMC$.
\begin{definition}[Worst-group Distance to Calibration Error]
    \label{def:wdMC}
    Let a subgroup collection $\CC = \{S_1, \dots, S_k\}$ with $S_i \subseteq \CX$ and a predictor $f: \CX \to [0,1]$ be given. 
    We define the worst-group distance to calibration error $\wdMC_\CC(f)$ as:
    \begin{align*}
        \wdMC_\CC(f) := \max_{S \in \CC} \left(\Pr[S] \cdot \dCE_{\restr{\CD}{S}}(\restr{f}{S}) \right).
    \end{align*}
\end{definition}
The metric $\wdMC$ is a natural way to generalize $\dCE$ to multiple subgroups.
In particular, a small overall $\wdMC$ implies that on each subgroup $S \in \CC$, there is a perfectly calibrated predictor $g:S \to [0,1]$ for which $\metricres{1}{S}{f}{g}$ is small.\footnote{For technical reasons used to obtain \Cref{prop:wdmc-leq-dmc}, we define $\wdMC$ with a normalization term based on the subgroup size. We note that our main result for $\wdMC$, presented in \Cref{lemma:improving-wdmc}, can be extended to a variant of $\wdMC$ which is simply the non-weighted, absolute worst group $\dCE$. However, this non-weighted version has little hope of being auditable, since subgroups with very small mass in $\CD_\x$ may drive up the worst-group $\dCE$ in unpredictable ways.}
Put differently, $f$ has small (distance to) calibration error when restricted to each subgroup, which is precisely the optimization objective of most multicalibration post-processing algorithms.
Nontheless, these post-processing algorithms usually use notions like worst-group ECE or $\ell_p$ calibration error, instead of worst-group distance to calibration error \citep{haghtalab2024unifying, hebert2018multicalibration}.

We proceed by viewing $\predictors$ as a compact subset of a $|\CX|$-dimensional vector space and $\wdMC: \predictors \to \R_{\geq 0}$ as a continuous hyper-surface---or \emph{loss landscape}---over this space.
One can view multicalibration post-processing algorithms as optimizing over this loss landscape (or some parameterized representation of it).
Given the limited use of direct gradient-based methods in multicalibration algorithms, a natural question is whether this loss landscape presents any barriers to tried-and-true continuous optimization techniques.

A very basic property of this loss-landscape would be that there are no local minima which are not global.
Such a property would guarantee that if we could perform gradient steps, we would not stumble upon a sub-optimal solution.
Unfortunately, via a simple example, we first demonstrate that $\wdMC_\CC(f)$ does not satisfy this property. In fact, it's possible for a local minimum to be \emph{far} from any predictor improving on this local minimum by a constant fraction.
The proof is deferred to \Cref{appx:proof-wdMC}.

\begin{proposition}\label{lemma:improving-wdmc}
    Take $\eps > 0$ sufficiently small. There exists a distribution $\CD \in \Delta(\CX \times \CY)$, subgroups $\CC = \{S_1, S_2\}$, and predictor $f$ such that $f$ locally minimizes $\wdMC_\CC$ and $\wdMC_\CC(f) \leq \eps$, but for any $\tilde{f}$ satisfying $\wdMC_\CC(\tilde{f}) \leq \eps/2$, it holds that $\metric{1}{f}{\tilde{f}} \in \Omega(1)$.
\end{proposition}
\begin{figure}
    \centering
    \includegraphics[width=0.5\textwidth]{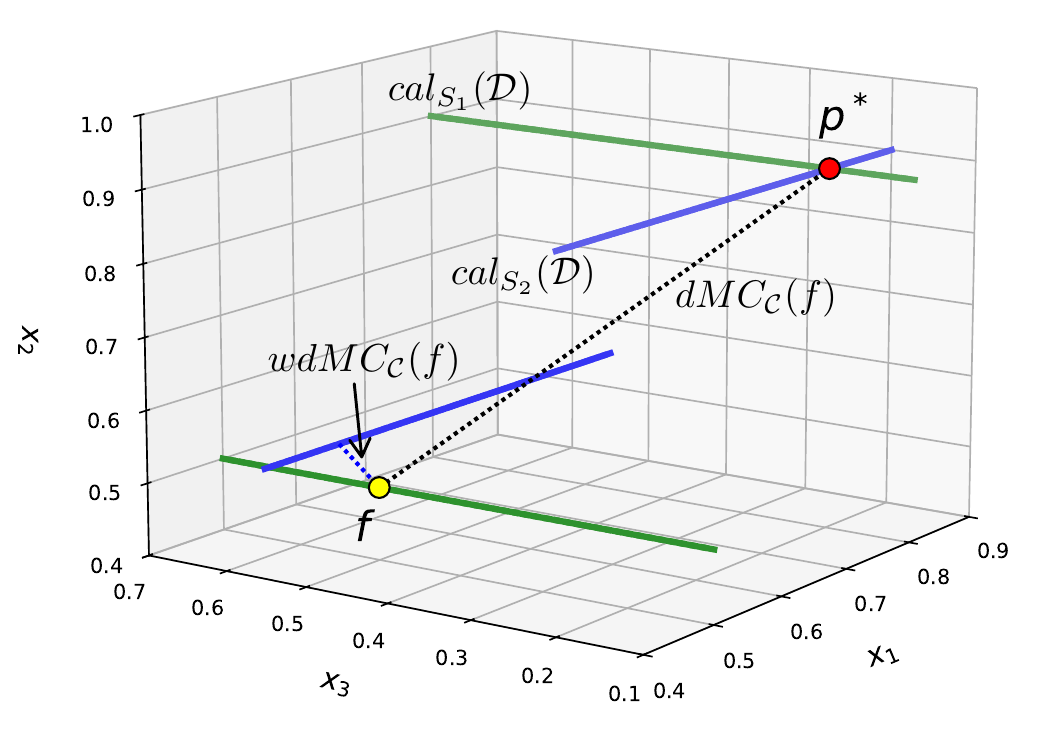}
    \caption{A visual illustration of the difference between $\wdMC$ and $\dMC$.
    The example pictured is detailed in the proof of \Cref{prop:no-audit-dMC}, setting $\alpha=0.1$.
    The green lines represent the set of predictors $\calset_{S_1} (\CD)$ calibrated with respect to $S_1$, and the blue lines are predictors from the set $\calset_{S_2}(\CD)$ calibrated with respect to $S_2$.
    In this example, the predictor $f$ is calibrated with respect to $S_1$, and has small worst-group calibration error $\wdMC$, represented by the dotted distance of $f$ to the closest predictor calibrated on $S_2$. However, $f$ is far from $p^*$, which is the only predictor which is simultaneously calibrated on $S_1$ \emph{and} $S_2$.
    Hence, $f$ has low $\wdMC$ but high distance to the nearest multicalibrated predictor, which is measured by $\dMC$.}
    
    \label{fig:difference-between-dMC-wdMC}
\end{figure}

The intuition for \Cref{lemma:improving-wdmc} is conveyed in \Cref{fig:difference-between-dMC-wdMC}. 
In particular, \Cref{lemma:improving-wdmc} exhibits a distribution $\CD$, collection $\CC = \{S_1, S_2 \}$, and predictor to be audited $f$ which is $\eps$-close to being calibrated on both groups $S_1$ and $S_2$ (and further, $f$ is a local minimum of $\wdMC_\CC(f)$).
However, the only place where the sets $\calset_{S_1}(\CD)$ and $\calset_{S_1}(\CD)$ intersect is actually very far from $f$.
As a bottom line, \Cref{lemma:improving-wdmc} demonstrates that measuring the worst-group calibration error does not tell us anything about the \emph{minimum} a predictor $f$ may be required to change---in the geometry of the $\ell_p$ metric space---in order to be converted into a predictor with lower error.

\Cref{lemma:improving-wdmc} may seem somewhat counter-intuitive given the results we have on achieving multicalibration via post-processing algorithms.
Works such as \citet{hebert2018multicalibration, haghtalab2024unifying} provide theorems which can drive down the worst-group calibration error of a predictor $f$ to $\eps$ with $O(\frac{1}{\poly(\eps)})$ examples for arbitrarily small $\eps$.\footnote{Note that these algorithms objective functions are often a discretized, $\ell_1$ worst-group calibration error notion, \emph{not} the worst-group distance to calibration error $\wdMC$ that we consider.} \Cref{lemma:improving-wdmc} does not contradict these results precisely because multicalibration post-processing algorithms do \emph{not} give guarantees on how much the input predictor $f$ may be required to be changed at different error thresholds. Importantly, it instead demonstrates that there exist certain thresholds $\eps$ at which the predictor $f$ may be required to change \emph{drastically} in order to achieve a worst-group calibration error below a constant fraction of $\eps$.

\subsection{A Nicely Behaved Multicalibration Error Loss Surface}
\label{sec:dMC}

\Cref{lemma:improving-wdmc} directly motivates our next question: 
Is there a multicalibration error metric whose loss surface has the property that \emph{any local minima are global minima?}
If such a metric existed, it would suggest the possibility of gradient-based algorithms that optimized this metric directly to find multicalibrated predictors. 

We will argue that such a metric exists, but at a cost. To start, we consider another multi-group analog of $\dCE$.
\begin{definition}(Distance to Multicalibration Error)
Let a predictor $f$, distribution $\CD \in \Delta(\CX \times \CY)$, and subgroup collection $\CC$ be given.
We define the distance to multicalibration $\dMC_\CC(f)$ as
\begin{align*}
\dMC_{\CC, \CD}(f) &= \inf_{g \in \mcalset_\CC(\CD)} \metric{1}{f}{g}.
\end{align*}
When clear from context, we omit the distribution subscript and write $\dMC_\CC(f)$.
\label{def:dMC}
\end{definition}

The main difference between $\wdMC$ and $\dMC$ is that the latter considers ``perfect'' predictors to be those that are calibrated on all subgroups in $\CC$ simultaneously. This is illustrated with an example in \Cref{fig:difference-between-dMC-wdMC}.
Furthermore, in \Cref{appx:wdmc-bounded-by-dmc} we show that $\wdMC$ is always upper bounded by $\dMC$.

Just as $\calset(\CD)$, $\mcalset_\CC(\CD)$ is generally not convex.
Neither is it true that $\dMC(f)$ is convex in $f$.
However, we demonstrate that $\dMC$ does satisfy the property that all local minima are global minima.
\begin{proposition}
\label{thm:dMC-global-minima}
    Let $\CD \in \Delta(\CX \times \CY)$, $\CC \subset 2^\CX$, and $f \in \predictors$. All local minima of $\dMC_{\CC}$ are global minima.
\end{proposition}

\noindent We note that this claim is a special case of a much more general fact about distances to sets in normed vector spaces; we leave this generalization to \Cref{appx:generalized-dmc-global-minima}.

\paragraph{On the Geometry of dMC and Perfectly Multicalibrated Predictors.} 
In addition to nice minima, $\dMC$ has a predictable continuous structure over $[0,1]^\CX$. Take $\CC = \{S_1, \ldots, S_k\}$, and note that
\begin{align*}
    \mcalset_\CC(\CD) = \bigcap_{S_i \in \CC} \Big\{f \in [0,1]^\CX : \restr{f}{S_i} \in \calset(\restr{\CD}{S_i})\Big\} = \bigcap_{S_i \in \CC} \bigcup_{g \in \calset(\restr{\CD}{S_i})} \Big( [0,1]^\CX \cap A_g \Big),
\end{align*}
where $A_g$ denotes the affine subspace of $\R^\CX$ given by the constraint that $f \in A_g$ satisfies $f(x) = g(x)$ for all $x \in S_i$. Since $\calset(\restr{\CD}{S_i})$ are all finite, we can take $\ell_i = |\calset(\restr{\CD}{S_i})|$,  and index these spaces $\{A_{ij}\}_{j=1}^{\ell_i}$ for $i \in [k]$. Distributing intersections, and taking multi-indices $\alpha = (\alpha_i)_i \in \prod_i [\ell_i]$, we have \begin{align*}
    \mcalset_\CC(\CD) &= \bigcap_{i=1}^{k} \bigcup_{j=1}^{\ell_i} \Big([0,1]^{\CX} \cap A_{ij} \Big)\\
    &= \bigcup_{\alpha \in \prod_i [\ell_i]} \bigcap_{i=1}^{k} \Big( [0,1]^\CX \cap A_{i \alpha_i} \Big) \\
    &= \bigcup_{\alpha \in \prod_i [\ell_i]} \left( [0,1]^\CX \cap \bigcap_{i=1}^{k} A_{i \alpha_i} \right).
\end{align*}
Since the intersection of affine subspaces is either the empty set or an affine subspace, $\mcalset_\CC(\CD)$ is a union of finitely many affine subspaces truncated to $[0,1]^\CX$. Denoting these truncated subspaces $C_\alpha$, we now have the decomposition \begin{align}\label{eq:dMC-decomposition}
    \dMC_{\CC}(f) = \inf_{g \in \mcalset_\CC(\CD)} \metric{1}{f}{g} = \min_{\alpha \in \prod_i [\ell_i]} \inf_{h \in C_\alpha} \metric{1}{f}{h}.
\end{align} 
That is, we take the minimum of several distances to orthogonal projections onto truncated affine spaces. It can be show that for all $\alpha \in \prod_i [\ell_i]$, $\inf_{h \in C_\alpha} \metric{p}{\cdot}{h}$ is convex. That is, $\dMC_{\CD, \CC}$ can be seen as the minimum of finitely many convex functions $[0,1]^\CX \to \R$. In the $\normnoarg{1}$ case, each projection is piecewise linear, giving $\dMC$ a piecewise linear structure.

From this perspective, one can observe where the difficulty of $\wdMC$ arises. Instead of taking a minimum over several distances to truncated affine spaces, we take the maximum over several such problems. \begin{align*}
    \wdMC_\CC(f) = \max_{S_i \in \CC} \min_{g \in \calset(\restr{\CD}{S})} \Pr[S] \cdot \metricres{1}{S}{f}{g}.
\end{align*}
The decomposition in \Cref{eq:dMC-decomposition} provides intuition for some of the results that follow, though we dispense with it here for concision.

\begin{remark}[Auditability of dMC]
\label{rmk:auditability-of-dMC}
While \Cref{thm:dMC-global-minima} and the decomposition in \Cref{eq:dMC-decomposition} suggests that we can optimize $\dMC$ to obtain good solutions, it has been shown in restricted contexts that achieving low $\dMC$ may be difficult.
\citet{dwork2024fairness} show that minimizing $\dMC$ --- which they refer to as ``strong'' distance to multicalibration error --- may be impossible in a general online setting, and online-to-batch results suggest that this may hold even in the offline setting. The hard instance that they present uses a number of groups $k$ on the order of the size of the domain $|\CX|$.
We find reason to be even more pessimistic; we demonstrate an instance on which $\dMC$ cannot even be \emph{audited} in a property testing sense, as studied in the calibration context by \citet{hu2024testing}. Even with only two groups, each with a size given as a constant fraction of the domain, we find an information-theoretic barrier: discontinuities w.r.t. $p^*$.
\end{remark}

\begin{proposition}\label{prop:no-audit-dMC}
    There exists a domain $\CX$, subgroups $\CC \subset 2^\CX$, a predictor $f \in \predictors$, and family of distributions $\CD_\alpha \in \Delta(\CX \times \CY)$ for $\alpha \in [0,.2]$ such that $\dMC_{\CC,\CD_\alpha}(f) > .3$ if $\alpha > 0$, and $\dMC_{\CC,\CD_0}(f) = 0$, but $\on{TV}(\CD_0, \CD_\alpha) \in O(\alpha)$.
\end{proposition}

\begin{proof}
    Let $\CX = \{x_1, x_2, x_3\}$ and $\CC = \{S_1, S_2\}$, where $S_1 = \{x_1, x_2\}$, and $S_2 = \{x_2, x_3\}$. Let the marginal distribution $\CD_\x$ be the uniform distribution. For a fixed $\alpha \in [0,0.2]$, we define the ground truth distribution $p^*$ as:
    \begin{align*}
          p^*(x_1) = 0.8,\ p^*(x_2)=0.2,\ p^*(x_3) = 0.8+\alpha.
    \end{align*}
    Define $f: \CX \to [0,1]$ by $f(x) = 0.5$ for all $x \in \CX$. 

    There are two predictors (up to restriction by $S_1$) which are perfectly calibrated with respect to $S_1$. The first one being $\restr{p^*}{S_1}$, and the second one being the predictor $g$ such that $g(x_1)=g(x_2)=0.5$. 
    Similarly, there are two predictors (up to restriction by $S_2$) which are perfectly calibrated with respect to $S_2$: $\restr{p^*}{S_2}$ and $h$ where $h(x_2) = h(x_3) = 0.5+\alpha/2$. 

    \textbf{Case 1.} If $\alpha > 0$, then $g(x_2) \neq h(x_2)$, thus $p^*$ is the only predictor which is both calibrated with respect to $S_1$ when restricted to $S_1$ and calibrated with respect to $S_2$ when restricted to $S_2$. It follows that $\dMC_\CC(f) = \metric{1}{f}{p^*} > C$ for some constant $C > 0.3$.
    
    \textbf{Case 2.} If $\alpha = 0$, then we have that $\restr{f}{S_1}=g$ and $\restr{f}{S_2} = h$.
    In this case, there are two multicalibrated predictors, $p^*$ and $f$, and therefore, $\dMC_\CC(f)=0$.

    Finally, note that \begin{align*}
        \on{TV}(\CD_0, \CD_\alpha) = \frac{1}{2}\left| \frac{1}{3} \Pr_{\CD_0}[\y = 1 | x_3] - \frac{1}{3} \Pr_{\CD_\alpha}[\y = 1 | x_3]\right|  + \frac{1}{2}\left| \frac{1}{3} \Pr_{\CD_0}[\y = 0 | x_3] - \frac{1}{3} \Pr_{\CD_\alpha}[\y = 0 | x_3]\right| = \frac{\alpha}{3}.
    \end{align*}
    The discontinuity which emerges due to this example is illustrated in \Cref{fig:dmc-discont}.
\end{proof}

The result above has immediate no-free-lunch implications. Let $\eps \in (0, 0.3]$, and suppose one has a hypothesis test which, given a finite i.i.d. sequence $(\x_1, \y_1), \ldots, (\x_m, \y_m)$ for $\x_i \sim \CD$ and $\y_i \sim \on{Ber}(p^*(\x_i))$ and unlimited oracle-query access to $f$, can determine whether $\dMC_\CC(f) = 0$ or $\dMC_\CC(f) \geq \eps$ with probability at least $1-\delta > 1/2$, for any $\CD$ chosen from $\{\CD_\alpha : \alpha \in [0,0.2]\}$. By \Cref{prop:no-audit-dMC}, this implies that for any $\alpha \in (0, 0.2]$, one can distinguish between $\CD_0$ and $\CD_\alpha$ with probability at least $1-\delta$. By Le Cam, however, we have that 
\begin{align*}
    2\delta > 1 - \on{TV}(\CD_0^{\otimes m}, \CD_\alpha^{\otimes m}) \geq (1-O(\alpha))^m.
\end{align*}
Since we can take any $\alpha > 0$, no such test exists. Crucial in this setup is the requirement that a multicalibration test must succeed with a finite number of samples dependent on the family of distributions from which $\CD$ is chosen. Fundamentally, this barrier arises from discontinuities that occur within such a family.

We remark that a similar counterexample exists to \Cref{prop:no-audit-dMC} when we place additional natural requirements on $\CC$ which at first glance may appear get around the counterexample. 
In particular, in \Cref{corr:extending-no-audit-dMC} we will demonstrate an instance where: (1) $\CX \setminus S \in \CC$ of each $S \in \CC$, (2) $\CX \in \CC$, and (3) $|\CX|$ is arbitrarily large but \Cref{prop:no-audit-dMC} still holds.

\Cref{prop:no-audit-dMC} showcases that an arbitrarily small change $\alpha$ in the ground truth labeling function $p^*$ can both (1) alter the size of the set $\mcalset_\CC(\CD)$ of $\CC$-multicalibrated predictors for the distribution $\CD$; and (2) add a new multicalibrated predictor in a totally different region of the prediction space $\predictors$ (see \Cref{fig:dmc-discont}).
This is unfortunate news, since it shows that even when the marginal $\CD_\x$ is held fixed, there are \emph{discontinuities} in $\dMC$ with respect to $p^*$.
This discontinuity highlights another key desirable property of any ``distance to'' notion: when holding the marginal $\CD_\x$ fixed, the metric should be \emph{continuous} with respect to changes in the ground truth $p^*$.
This is a core requirement for auditing to be possible, since it implies that we do not have to obtain the \emph{exact} ground truth $p^*$ to audit for the calibration notion.
It is relatively straightforward to see that the original $\dCE$ definition of \citet{blasiok2022unifying} satisfies this property.\footnote{Indeed, this also follows from the fact that $\dCE$ is approximately auditable in the ``prediction-access'' model of \citet{blasiok2022unifying}.}

It will be useful to establish additional notation. Let $\CD \in \Delta(\CX\times \CY)$. For a predictor $p \in \predictors$, we let $\CD_p$ denote the distribution induced by drawing $\x \sim \CD_x$ and $\y \sim \on{Ber}(p^*(\x))$, and define $\dC{\CD}{p}(f) := \dCE_{\CD_p}(f)$.
We may now consider continuity notions with respect to different ground truths.

\begin{lemma} \label{lemma:dCE-Lipschitz}
Let $\CD \in \Delta(\CX \times \CY)$ and $f \in \predictors$. Then $\dCE_{\CD}(f)$ is $1$-Lipschitz with respect to the ground truth $p^*$. That is, for $p_1, p_2 \in \predictors$, we have 
\begin{align*}
    |\dC{\CD}{p_1}(f)-\dC{\CD}{p_2}(f)| \leq \metric{1}{p_1}{p_2}.
\end{align*} 
\end{lemma}
\noindent We leave the proof to \Cref{appx:dCE-Liphcitz}.

\paragraph{Extension of \Cref{prop:no-audit-dMC} to Other Multi-group Notions.}
The works \citet{gopalan2022low, casacuberta2025global} showcase the utility of weakening the full multicalibration definition by relaxing the conditioning on $f(x)$ in the definition of calibration.
This is done by, \textit{e.g.}, restricting the dependence to various degree polynomials, and combining it with other notions like global calibration.
A natural question is whether we can \emph{circumvent} the pathology exhibited in \Cref{prop:no-audit-dMC} by weakening the notion of multicalibration considered. 
In \Cref{appx:generalizing-dMC}, we show that similar counter-examples exist for two weaker distance to multi-group notions: distance to \emph{low-degree multicalibration} and distance to \emph{calibrated multiaccuracy}.
Both remain impossible to audit due to discontinuities with respect to the ground truth.

\paragraph{Lipschitzness of wdMC.}
It is not difficult to see that $\wdMC$ is also 1-Lipschitz in $p^*$. By definition, of $\wdMC$, we have $\wdMC_\CC(f) = \max_i \Pr[S_i] \cdot \dCE_{\restr{\CD}{S_i}}(f)$. The observation follows from that fact that $\dCE_{\restr{\CD}{S_i}}(f)$ is 1-Lipschitz and the maximum function is 1-Lipschitz for the supremum norm.

\section{Continuized Distance to Multicalibration Error}
\label{sec:cdmc}

The example in \Cref{prop:no-audit-dMC} demonstrates that there are circumstances where $\dMC$ is impossible to audit or even approximate.
In \Cref{subsec:continuized-dMC}, we introduce a variant of $\dMC$, the \emph{continuized} distance to multicalibration error, $\cdMC$).
Carefully \emph{smoothing} the $\dMC$ over a sequence of local neighborhoods, $\cdMC$ is 1-Lipschitz in the ground truth labeling function $p^*$, satisfying a key property for the information-theoretic tractability discussed in \Cref{sec:dMC}.
We examine this notion over the course of Sections~\ref{sec:dIME} and \ref{sec:proof-of-cdmc-lipschitz}, via an equivalence to a new distance notion, the distance to \emph{intersection} multicalibration error.
Unfortunately, in \Cref{prop:no-audit-dimc}, we show that auditing $\cdMC$ (and $\dIMC$) remains \emph{statistically} hard with only a number of samples from $\CD$ which is polynomial in the domain size $n = |\CX|$ and subgroup collection size $k = |\CC|$.
However, for the common scenario in practice with a \emph{constant} number of groups $k$, we show that $\cdMC$ (and $\dIMC$) are both statistically and computationally feasible to audit, and provide a recipe for doing so via access to a $\dCE$ auditing oracle.

\subsection{Smoothing Away Discontinuities of $\dMC$}
\label{subsec:continuized-dMC}
The reason that $\dMC$ is not auditable is that the set $\mcalset_\CC(\CD)$ does not continuously vary with $p^*$.
In particular, recall the example given in \Cref{prop:no-audit-dMC} where $p^*$ is parameterized by $\alpha$.
Whenever $\alpha > 0$, $p^*$ ended up being the \emph{only} multicalibrated predictor, and $\dMC_\CC(f) > 0.3$.
However, when we change $\alpha$ to be \emph{exactly} zero, $f$ suddenly also becomes a multicalibrated predictor, and $\dMC_\CC(f)$ drops to zero by definition. 
This also demonstrates that even the cardinality of the set $\mcalset_\CC(\CD)$ is not constant with slightly changing $p^*$ (see \Cref{fig:dmc-discont}).

At a high level, we argue that $\alpha=0$ is a special case for the ground truth $p^*$ that induces \emph{agreement} between groups.
This is because at $\alpha=0$, $p^*$ looks identical when conditioning on the groups $S_1$ and $S_2$.
This in turns adds another predictor $f$ to the multicalibrated predictor set $\mcalset_\CC(\CD)$, which drops the distance $\dMC$ from $0.3$ to $0$.
We will eventually show that these discontinuities, for example at $\alpha=0$, can only ever \emph{lower} the $\dMC$.
Intuitively, this occurs because the $\mcalset_\CC(\CD)$ momentarily expands, which makes the $\normnoarg{1}$ distance of any predictor to the set shrink.

Working with the fact that all discontinuities have their image below the local neighborhood of $\dMC$, it is natural to introduce a \emph{continuized} version of $\dMC$, which we refer to as $\cdMC$.
In this continuized version $\cdMC$, we remove all discontinuities of $\dMC$ by taking the \emph{supremum} of $\dMC_{\CC}(f)$ over a sequence of neighborhoods around the ground truth $p^*$, effectively removing these lower discontinuities.

As above, we will need specify metrics with respect to varying ground truths and subgroup collections.  Let $\CD \in \Delta(\CX\times \CY)$. For a predictor $p \in \predictors$, recall that $\CD_p$ denotes the distribution induced by drawing $\x \sim \CD_x$ and $\y \sim \on{Ber}(p^*(\x))$. We let $\dMCE{\CC}{p}(f) := \dMC_{\CC, \CD_p}(f)$. When $p^*$ is clear from context, we omit the associated subscript.  We are now ready to formally introduce the continuized metric.

\begin{definition}[Continuized Distance to Multicalibration Error]
    Let $B_\eps(p)$ be the closed ball of $\normnoarg{1}$-radius $\eps$ around the predictor $p \in [0,1]^\CX$.\footnote{We remark that the open ball can work as well, at the expense of longer proofs.} 
    We define \emph{continuized distance to multicalibration error} as follows:
    \begin{align*}
    \dME{\CC}{p^*}(f) := \lim\limits_{\eps \to 0^+} \sup_{p \in B_\eps(p^*)} \dMCE{\CC}{p}(f).
    \end{align*}
    \label{def:cdMC}
\end{definition}
There are a few important things to note about the definition of $\cdMC$.
First, notice that since $p^* \in B_\eps(p^*)$ for all $\eps > 0$, it is always the case that $\dMCE{\CC}{p^*}(f) \leq \dME{\CC}{p^*}(f)$.
Second, as a sanity check, we note that $\cdMC$ is not simply measuring distance to the ground truth $p^*$.
Rather, we wish for $\cdMC$ to rule out ``bad'' multicalibrated predictors which induce discontinuities in the metric, and account for the remaining ``good'' multicalibrated predictors.
The following proposition, whose proof is deferred to \Cref{proof:cdMC-not-dist-to-p-star}, demonstrates that $\cdMC$ is not pathological in this sense, measuring distances to multicalibrated predictors other than $p^*$.
\begin{proposition}\label{prop:cdMC-not-dist-to-p-star}
    There exists distribution $\CD$ with ground truth $p^*$ and subgroup collection $\CC$, and predictor $f \neq p^*$ such that $\metric{1}{f}{p^*} = \Omega(1)$ and $\dME{\CC}{p^*}(f) = 0$.
\end{proposition}
Finally, and most importantly, $\cdMC$ seems impossible to compute a priori as its definition relies heavily on knowledge of $p^*$. Surprisingly, in \Cref{thm:dimc-continuized-dmc}, we show that this is actually not the case: computing $\cdMC$ is equivalent to computing distance to the nearest predictor which is calibrated with respect to all \emph{intersections} of groups in $\CC$.
This provides a path towards auditability of $\cdMC$ in some common scenarios discussed further in \Cref{subsec:auditability-cdMC}.

We are now prepared to discuss the properties of $\cdMC$.
In Sections~\ref{subsec:measuring-multical} and \ref{sec:dMC}, we argued that there were two essential requirements for multicalibration error metric: local minima are global, and lipschitzness in the ground truth labeling function $p^*$.
The main result in this section is that $\cdMC$ indeed satisfies both. 

\begin{theorem}
\label{prop:cdmc-lipschitz}
    Let $\CD \in \Delta(\CX \times \CY)$, subgroup collection $\CC$, and $f \in \predictors$ be given. Assume that $\CC$ covers the domain $\CX$. Then, $\cdMC_{\CC, p^*}(f)$ satisfies the following:
    \begin{enumerate}
        \item All local minima are global minima.
        \item $\cdMC_{\CC, p^*}(f)$ is 1-Lipschitz with respect to the ground truth $p^*$.
    \end{enumerate}
\end{theorem}

\noindent The proof of \Cref{prop:cdmc-lipschitz} is presented in \Cref{sec:proof-of-cdmc-lipschitz} and crucially relies on the equivalence between the continuized and \emph{intersection} distance notions, which we examine in \Cref{sec:dIME}.

\subsection{Distance to Intersection Multicalibration Error}
\label{sec:dIME}
As in the statement of \Cref{prop:cdmc-lipschitz}, we will assume that $\CC$ covers $\CX$ throughout this section.

\begin{definition}[Distance to Intersection Multicalibration Error]
\label{def:dIMC}
    Let $\CD \in \Delta(\CX \times \CY)$ with ground truth $p^*$, subgroup collection $\CC$, and predictor $f \in \predictors$ be given.
    Let $\intcl(\CC)$ be the intersection closure of the set $\CC$. Explicitly,
\begin{align*}
\intcl(\CC) = \left\{\bigcap\limits_{S \in \CA} S: \CA \subseteq \CC \ \land\  \CA \neq \emptyset \right\}.
\end{align*}
Then, we define the \emph{distance to intersection multicalibration error}, $\dIMC$, as follows:
\begin{align*}
    \dIME{\CC}{p^*}(f) := \dMCE{\intcl(\CC)}{p^*}(f).
\end{align*} 
\end{definition}

\noindent Before proceeding, we note that $\dIMC$ in fact satisfies the two desiderata of distance to multicalibration metrics.

\begin{theorem}
\label{prop:dimc-lipschitz}
    Let the distribution $\CD$, subgroup collection $\CC$, and predictor to audit $f$ be given. Assume that $\CC$ covers the domain $\CX$. Then, $\dIMC_{\CC, p^*}(f)$ satisfies the following:
    \begin{enumerate}
        \item All local minima are global minima.
        \item $\dIMC_{\CC, p^*}(f)$ is 1-Lipschitz with respect to the ground truth $p^*$.
    \end{enumerate}
\end{theorem}

\noindent Since $\dIMC$ is a special case of $\dMC$, the first criteria of \Cref{prop:dimc-lipschitz} comes directly from $\dMC$. The second comes from the nontrivial fact that multicalibration on $\intcl(\CC)$ is equivalent to multicalibration on a disjoint cover of $\CX$, defined in \Cref{def:partition-generated}; then by \Cref{lemma:dMC-Lipschitz-if-groups-disjoint}.

As a first step to proving \Cref{prop:cdmc-lipschitz}, we will show that surprisingly, $\dMC$ and $\dIMC$ are equivalent \emph{almost everywhere} in the space of ground truth predictors (\Cref{thm:dIMC-equals-dMC-almost-everywhere}).
That is, for most practical scenarios, we can treat them as equivalent quantities.

\begin{proposition} \label{thm:dIMC-equals-dMC-almost-everywhere}
    Let a subgroup collection $\CC = \{ S_1, \dots, S_k \}$ be given.
    Let $P \subset \predictors$ be the set of ground truth predictors $p^*$ such that $\dMCE{\CC}{p^*}(f) \neq \dIME{\CC}{p^*}(f)$ for some $f \in \predictors.$
    
    Then $P$ is contained in a set of Lebesgue-measure zero.

\end{proposition}
\begin{proof}
    Fix $p^*, f: \CX \to [0,1]$ such that $\dMCE{\CC}{p^*}(f) \neq \dIME{\CC}{p^*}(f)$. 
    Since $\mcalset_\CC(\CD)$ is finite (\Cref{lemma:mcalset-finite}), it must be the case that there is a predictor $g: \CX \to [0,1]$ such that $\dMCE{\CC}{p^*}(f) = \metric{1}{f}{g}$ and $g$ is not intersection multicalibrated. 
    In particular, there is some set $A \in \intcl(\CC)$ such that $g$ is not calibrated with respect to $A$. 

    It follows that there is some $x \in A$ such that $g(x) = v$ and the following holds: 
    \begin{align}\E\limits_{\x \sim \CD_\x}\Big[p^*(\x) \big| \x \in A \cap g^{-1}(v) \Big] \neq v.
    \label{eq:not-calibrated-wrt-intersection}
    \end{align}

    Let $S_i \in \CC$ be a set such that $A \subset S_i$.
    $S_i$ must exist since all elements of $\intcl(\CC)$ are intersections of groups in $\CC$.
    Let $V_i = S_i \cap g^{-1}(v)$. By $g$ being calibrated with respect to $S_i$, we have \begin{align*}
        v=\E\limits_{\x \sim \CD_\x}\Big[p^*(\x) \big| \x \in V_i \Big].
    \end{align*}
    Furthermore, we know $V_i \not \subseteq A$ because if it was, we would have $V_i = A \cap V_i= A \cap g^{-1}(v)$, which would contradict \Cref{eq:not-calibrated-wrt-intersection}.

    Using this fact, there exists some $x' \in S_i$ such that $x' \in V_i$, but $x' \notin A$.
    Note that $A = \bigcap_{j \in I} S_j$ for an index set $I \subseteq [k]$ for which $i \in I$ (since $A \subset S_i$). 
    Then, there must exist some $j \in I$ such that $x' \not \in S_j$ because if this was not the case, then we would have $x' \in S_j$ for all $j \in I$, meaning $x' \in \bigcap_{j \in I}S_j= A$.

    Let $V_j = S_j \cap g^{-1}(v)$. Importantly, we have $V_i \neq V_j$. By $g$ being calibrated with respect to $S_j$, we have \begin{align*}
        v=\E\limits_{\x \sim \CD_\x}\Big[p^*(\x) \big| \x \in V_j \Big].
    \end{align*}
    In particular, we have the following equality:
    \begin{align}
    \sum_{x \in V_i} p^*(x) \Pr[x]/\Pr[V_i] &= \sum_{x \in V_j} p^*(x) \Pr[x]/\Pr[V_j]. \label{eq:constraint-on-p-star}
    \end{align}

    \Cref{eq:constraint-on-p-star} can be rewritten as $\langle c, p^* \rangle = 0$ for some \emph{non-zero} vector $c \in \R^n$ for $n = |\CX|$.Therefore, the equation $\langle c, p^* \rangle = 0$ describes an $(n-1)$-dimensional subspace. In other words, the space of solutions for $p^*$ to \Cref{eq:constraint-on-p-star} is a measure-zero set in the ambient space of predictors $[0,1]^n$.

    If $g$ is multicalibrated but not intersection multicalibrated, we have shown that $p^*$ must satisfy at least one equation in the form of \Cref{eq:constraint-on-p-star}, and the solution space of each equation is a measure-zero subset of $[0,1]^\CX$. 
    
    Since $\CX$ is finite, there are at most $2^{2n}$ choices of $V_i$ and $V_j$ as subsets of $\CX$.
    
    The finite union of measure-zero sets is measure-zero.
    Therefore, the set of all values of $p^*$ which satisfy any equation in the form of \Cref{eq:constraint-on-p-star} is a measure-zero set.
\end{proof}

\noindent As results that follow will demonstrate, intuitively, the set of points where $\dIMC$ and $\dMC$ differ is exactly the ``bad'' ground truths $p^*$ that block auditability in \Cref{prop:no-audit-dMC}.
Viewed another way, \Cref{thm:dIMC-equals-dMC-almost-everywhere} says that these ``bad'' ground truths constitute a measure-zero subset in the space of all possible ground truths.
In other words, most of the time the multicalibrated predictor $g \in \mcalset_\CC(\CD)$ for which $\dMC_\CC(f) = \metric{1}{f}{g}$ is also multicalibrated on the \emph{intersections} of subgroups in the collection $\CC$.

\Cref{thm:dIMC-equals-dMC-almost-everywhere} also implies that any properties of $\dMC$ will also, for most ground truths $p^*$, be true for $\dIMC$.
In particular, our next result shows that $\dMC$ is 1-Lipschitz when restricted to \emph{disjoint covers} of $\CX$.
Combining this with a particular choice of disjoint cover then allows us to demonstrate that $\dIMC$ is 1-Lipschitz. 

\begin{lemma} \label{lemma:dMC-Lipschitz-if-groups-disjoint}
    Let $\CD \in \Delta(\CX \times \CY)$, subgroups $\CC$, and predictor $f \in \predictors$ be given.
    If $\CC$ is a disjoint cover of $\CX$, then $\dMC_{\CC, p^*}$ is $1$-Lipschitz with respect to $p^*$. That is, for any two ground truth labeling functions $p_1, p_2 \in \predictors$:
    \begin{align*}
        \left| \dMCE{\CC}{p_2}(f)-\dMCE{\CC}{p_1}(f) \right| \leq \metric{1}{p_1}{p_2}.
    \end{align*}
\end{lemma}
\begin{proof}
    Let $\CC = \{S_1, \cdots, S_k\}$ be a collection of disjoint sets covering $\CX$. Fix a function $f: \CX \to [0,1]$, and let $p_1, p_2: \CX \to [0,1]$ be ground truth predictors. Without loss of generality, we can assume $\dMCE{\CC}{p_1}(f) \leq \dMCE{\CC}{p_2}(f)$, and it suffices to show that $\dMCE{\CC}{p_2}(f)-\dMCE{\CC}{p_1}(f) \leq \metric{1}{p_1}{p_2}$.

    By \Cref{lemma:mcalset-finite}, we know there are finitely many multicalibrated predictors, so let $g: \CX \to [0,1]$ be a multicalibrated predictor with $p_1$ as the ground truth such that $\dMCE{\CC}{p_1}(f) = \metric{1}{f}{g}$. 
    For each $i \in [k]$, let $g_i= \restr{g}{S_i}$, and let $f_i = \restr{f}{S_i}$. Since $\CC$ is a disjoint cover of $\CX$, it follows that
    \begin{align*}
        \metric{1}{f}{g} = \sum_{i=1}^k \metricres{1}{S_i}{f_i}{g_i} \cdot \Pr[S_i].
    \end{align*}

    We claim that for every subgroup $S_i$, we have that $\dC{\restr{\CD}{S_i}}{p_1}(f_i) = \metricres{1}{S_i}{f_i}{g_i}$.
    To see this, suppose for contradiction that there was some $\rho_i:S_i \to [0,1]$ which was perfectly calibrated on $S_i$ with respect to $p_1$ for which $\metricres{1}{S_i}{f_i}{g_i} > \metricres{1}{S_i}{f_i}{\rho_i}$.
    Define a new predictor $\tilde g:\CX \to [0,1]$ in the following manner:
    \begin{align*}
        \tilde g(x) = \begin{cases}
            g(x) & \text{ if } x \notin S_i \\
            \rho_i(x) & \text{ if } x \in S_i
        \end{cases}.
    \end{align*}
    Notice that $\tilde g$ is perfectly multicalibrated with respect to $p_1$, since it is equal to $g$ everywhere except $S_i$ (and $g$ is multicalibrated with respect to $p_1$), and on  $S_i$, we know that $\rho_i$ is also perfectly calibrated with respect to $p_1$.
    However, since  $\metricres{1}{S_i}{f_i}{g_i} > \metricres{1}{S_i}{f_i}{\tilde g_i}$, we now have that $\metric{1}{f}{\tilde g}<\metric{1}{f}{g}=\dMCE{\CC}{p_1}(f)$, which is a contradiction since we chose $g$ as the closest multicalibrated predictor to $f$.
    
    For every subgroup $S_i$ over $i \in [k]$, applying \Cref{lemma:dCE-Lipschitz} with the distribution $\restr{\CD}{S_i}$ tells us that with the given ground truths $p_1, p_2$, 
    \begin{align*}
       \left| \dC{\restr{\CD}{S_i}}{p_1}(f_i) - \dC{\restr{\CD}{S_i}}{p_2}(f_i) \right| 
       \leq \metricres{1}{S_i}{p_1}{p_2}.
    \end{align*}
        
    First, using finiteness of the calibrated predictor set (\Cref{lemma:calset-finite}), for all $S_i$, there exists $h_i : S_i \to [0,1]$ which is calibrated on $S_i$ with respect to $p_2$ such that $\dC{\restr{\CD}{S_i}}{p_2}(f_i) = \metricres{1}{S_i}{f_i}{h_i}$.
    Combining this with the above claim that $\dC{\restr{\CD}{S_i}}{p_1}(f_i) = \metricres{1}{S_i}{f_i}{g_i}$, we have that for all subgroups $S_i$,
    \begin{align}
    \label{disjoint-error-bound}
        \Big|\metricres{1}{S_i}{f_i}{g_i}-\metricres{1}{S_i}{f_i}{h_i}\Big| 
        \leq 
        \metricres{1}{S_i}{p_1}{p_2}.
    \end{align}

    Using each of these $h_i$, construct the predictor $h:\CX \to [0,1]$ such that $\restr{h}{S_i} = h_i$. Since $\CC$ is a disjoint cover of $\CX$, $h$ is well-defined.
    Moreover, since each $h_i$ is calibrated with respect to $S_i$ under ground truth $p_2$, we know $h$ is multicalibrated under the ground truth $p_2$.
    Therefore, we have that:
    \begin{align*}
        \dMCE{\CC}{p_2}(f) \leq \metric{1}{f}{h} = \sum\limits_{i=1}^k \metricres{1}{S_i}{f_i}{h_i} \cdot \Pr[S_i].
    \end{align*}
    Finally, we have
    \begin{align*}
    \dMCE{\CC}{p_2}(f) - \dMCE{\CC}{p_1}(f) &\leq \metric{1}{f}{h}-\metric{1}{f}{g}\\
    & = \sum\limits_{i=1}^k \left( \metricres{1}{S_i}{f^i}{h^i}-\metricres{1}{S_i}{f^i}{g^i} \right) \cdot \Pr[S_i] \\ 
    & \leq \sum\limits_{i=1}^k  \metricres{1}{S_i}{p_1}{p_2}  \cdot \Pr[S_i] && \text{(Applying \eqref{disjoint-error-bound})}\\
    & = \metric{1}{p_1}{p_2}.
    \end{align*}    
\end{proof}

We still require two additional ingredients to show that $\dIMC$ is Lipschitz.
First, we need to have a way of converting $\CI(\CC)$ to a disjoint cover.
Then, we need to show that $\dMC$ on the chosen disjoint cover preserves its value even though we have changed the subgroup collection.
The following choice of partition is sufficient.

\begin{definition}
\label{def:partition-generated}
    Let subgroup collection $\CC = \{S_1, \dots, S_k\}$ be given. 
    Define the \emph{partition generated by $\CC$} as:
    \begin{align*}
        \CJ(\CC) = \left\{ \left( \bigcap\limits_{S \in \CA} S \right) - \left( \bigcup\limits_{S \in \CC-\CA} S\right): \CA \subseteq \CC \ \wedge\  \CA \neq \emptyset \right\}.
    \end{align*}
    
\end{definition}
\noindent We know that $\CJ(\CC)$ is in fact a partition of $\CX$ because for all $x \in \CX$, there is a unique $\CA \subseteq \CC$ such that for all $i \in [k]$ we have $x \in S_i$ if and only if $S_i \in \CA$. The following lemma shows that we can convert between $\CI(\CC)$ and $\CJ(\CC)$ without losing any information on the set of multicalibrated predictors.
This is key, since $\CI(\CC)$ is \emph{not} in general a partition.
\begin{lemma} 
\label{lemma:PIE-dIMC-definition}
    Let $\CC$ be a disjoint cover of $\CX$. 
    For any distribution $\CD$ with ground truth labeling distribution $p^*$, we have that $\mcalset_{\CI(\CC)}(\CD) = \mcalset_{\CJ(\CC)}(\CD)$.
    Furthermore, $\dIME{\CC}{p^*}(f) = \dMC_{\CI(\CC),p^*}(f) = \dMCE{\CJ(\CC)}{p^*}(f)$.
\end{lemma}
\noindent We reserve the proof for \Cref{appx:dIME-equivalence}. We may now prove \Cref{prop:dimc-lipschitz}.

\begin{proof}[Proof of \Cref{prop:dimc-lipschitz}]
    To see property (1), notice that $\dIMC$ is simply defined as $\dMC$ for the particular subgroup collection $\CI(\CC)$. Therefore, (1) is true via \Cref{thm:dMC-global-minima}.

    To prove Lipschitzness of $\dIMC$ (property (2) of \Cref{prop:dimc-lipschitz}), notice that by \Cref{lemma:PIE-dIMC-definition}, we can define $\dIME{\CC}{p^*}(f)$ as $\dMCE{\CJ(\CC)}{p^*}(f)$. Since $\CJ(\CC)$ is a collection of disjoint sets which cover $\CX$, by \Cref{lemma:dMC-Lipschitz-if-groups-disjoint}, we know $\dIMC$ is 1-Lipschitz in $p^*$.
\end{proof}

\subsection{Completing the Proof of \Cref{prop:cdmc-lipschitz}}
\label{sec:proof-of-cdmc-lipschitz}

In \Cref{prop:dimc-lipschitz}, we showed that $\dIMC$ satisfies both 1-Lipschitzness and local minima are global minima. The following proposition demonstrates that the two notions $\dIMC$ and $\cdMC$ are equivalent, which therefore implies \Cref{prop:cdmc-lipschitz} as a corollary.

\begin{proposition} \label{thm:dimc-continuized-dmc}
    For $\CD \in \Delta(\CX \times \CY)$ and $f, p^* \in [0,1]^\CX$, we have $\dIME{\CC}{p^*}(f) = \dME{\CC}{p^*}(f)$.
\end{proposition}

\begin{proof}
    Fix $f, p^*: \CX \to [0,1]$, and let $\eps > 0$. Note that $\dMCE{\CC}{p^*}(f) \leq \dIME{\CC}{p^*}(f)$ because $\CC \subseteq \intcl(\CC)$. By \Cref{thm:dIMC-equals-dMC-almost-everywhere}, since $\dIME{\CC}{p^*}(f) \neq \dMCE{\CC}{p^*}(f)$ for only a measure zero set of ground truth predictors, there must be some $p \in B_\eps(p^*)$ such that $\dIME{\CC}{p}(f) = \dMCE{\CC}{p}(f)$. Therefore, we have 
    \begin{align}
        \lim_{\eps \to 0} \inf_{p \in B_\eps(p^*)} \dIME{\CC}{p}(f) &\leq \lim_{\eps \to 0} \sup_{p \in B_{\eps}(p^*)} \dMCE{\CC}{p}(f) \leq \lim_{\eps \to 0} \sup_{p \in B_\eps(p^*)} \dIME{\CC}{p}(f). \label{eq:tilde-dmc-equivalence}
    \end{align}
    Notice that the middle term of \eqref{eq:tilde-dmc-equivalence} is exactly $\dME{\CC}{p^*}(f)$. For the left and right hand sides, we have the inequality that for all $\eps > 0$, \begin{align*}
        \inf\limits_{p \in B_\eps(p^*)}\dIME{\CC}{p}(f) \leq \dIME{\CC}{p^*}(f) \leq \sup\limits_{p \in B_\eps(p^*)}\dIME{\CC}{p}(f).
    \end{align*}
    However, by \Cref{prop:dimc-lipschitz}, we know that $\dIMC$ is $1$-Lipschitz in $p^*$, so we have \begin{align*}
        \sup\limits_{p \in B_\eps(p^*)}\dIME{\CC}{p}(f)-\inf\limits_{p \in B_\eps(p^*)}\dIME{\CC}{p}(f) \leq 2\eps.
    \end{align*} 
    By squeeze theorem, we have \begin{align*}
        \lim\limits_{\eps \to 0} \inf\limits_{p \in B_\eps(p^*)} \dIME{\CC}{p}(f) = \dIME{\CC}{p^*}(f) =\lim\limits_{\eps \to 0} \sup\limits_{p \in B_\eps(p^*)} \dIME{\CC}{p}(f).
    \end{align*} We can rewrite \Cref{eq:tilde-dmc-equivalence} as follows
    \begin{align*}
        \dIME{\CC}{p^*}(f) \leq \dME{\CC}{p^*}(f) \leq \dIME{\CC}{p^*}(f).
    \end{align*}
\end{proof}

\begin{proof}[Proof of \Cref{prop:cdmc-lipschitz}]
    This follows from the equivalence of $\cdMC$ and $\dIMC$ in \Cref{thm:dimc-continuized-dmc}, and the properties of $\dIMC$ given in \Cref{prop:dimc-lipschitz}.
\end{proof}

\paragraph{Discussion and Implications.}

The equivalence between $\cdMC$ and $\dIMC$ may have implications that go beyond our purposes. To start, because $\CJ(\CC)$ is a disjoint cover of $\CX$, $\mcalset_{\CJ(\CC)}(\CD)$ is precisely the set of predictors $f$ defined by $\restr{f}{S_i} \in \calset(\restr{\CD}{S_i})$ for each $i$. Applying total expectation, one obtains a decomposition of the continuized metric. 
\begin{align*}
    \cdMC_\CC(f) &= \dMC_{\CJ(\CC)}(f)\\
    &= \min_{g \in \mcalset_{\CJ(\CC)}(\CD)} \metric{1}{f}{g}\\
    &= \min_{g \in \mcalset_{\CJ(\CC)}(\CD)} \sum_{S \in \CJ(\CC)} \Pr[S] \cdot \metricres{1}{S}{f}{g}\\
    &= \sum_{S \in \CJ(\CC)} \Pr[S] \min_{g \in \calset(\restr{\CD}{S})} \metricres{1}{S}{f}{g}\\
    &= \sum_{S \in \CJ(\CC)} \Pr[S] \cdot \dCE_{\restr{\CD}{S}}(f).
\end{align*}
This property will be particularly useful for the purpose of auditing. This decomposition comes at a cost, however: $|\CJ(\CC)|$ may be exponential in $|\CC|$.

We also can use the notion of intersection multicalibration to show a strengthened version of \Cref{prop:no-audit-dMC}, which extends to settings where $\CC$ is closed under set compliments, $\CX \in \CC$, and $|\CX|$ is arbitrarily large. As in \Cref{prop:no-audit-dMC}, all groups are a constant fraction of the domain. 
\begin{corollary} \label{corr:extending-no-audit-dMC}
Let $N \in \N$. There exists a domain $\CX$ such that $|\CX| \geq N$, subgroups $\CC$, predictor $f \in \predictors$, and family of distributions $\CD_\alpha \in \Delta(\CX \times \CY)$ for $\alpha \in [0,0.1]$ such that $\dMC_{\CC, p^*}(f) = 0$, $\dMC_{\CC, p_\alpha}(f) \geq 0.2$, and $\on{TV}(\CD_0,\CD_{\alpha}) \leq \alpha$.
\end{corollary}
\begin{proof}
    Let $|\CX| = 4N$ and index the $x \in \CX$ as $x_{i,j}$ where $i \in [4], j \in [N]$. Assume the marginal distribution $\CD_\x$ is uniform with $\Pr[x] = \frac{1}{4N}$ for all $x \in \CX$. Define $S_{i, i'} := \{x_{i, j}: j \in [N]\} \cup \{x_{i', j}: j \in [N]\}$, and
    let $\CC = \{S_{1,2}, S_{2,3}, S_{3, 4}, S_{4,1}, \CX\}$. Let $f(x) = 0.5$ for all $x \in \CX$. Define $p^*(x)$ as follows:
    \begin{align*}
        p^*(x_{i,j}) = \begin{cases}
            0.8 & \text{ if } i \text{ is even} \\
            0.2 & \text{ if } i \text{ is odd.}
        \end{cases}
    \end{align*}
    For all $S \in \CC$, we have 
    \begin{align*}
        \E\limits_{\x \sim \restr{\CD}{S}}\big[p^*(\x)| f(\x) = 0.5\big] = \E\limits_{\x \sim \restr{\CD}{S}}\big[p^*(\x)\big] = 0.5.
    \end{align*}
    Thus $f$ is multicalibrated, so we have $\dMC_{\CC, p^*}(f) = 0$. Note, however, that $f$ is not intersection multicalibrated.
    We have $\CJ(\CC) = \{S_{1,1}, S_{2,2}, S_{3,3}, S_{4,4}\}$, and since $p^*$ is constant on each element of $\CJ(\CC)$, the only intersection multicalibrated predictor is $p^*$ itself. It follows that $\dIMC_{\CC, p^*}(f) = \metric{1}{f}{p^*}=0.3$.

    Let $\alpha \in [0, 0.1]$ and $B_\alpha(p^*)$ be the closed ball of $\normnoarg{1}$ distance $\alpha$ around $p^*$, restricted $[0,1]^\CX$. Since $B_\alpha(p^*)$ has positive Lebesgue measure, by \Cref{thm:dIMC-equals-dMC-almost-everywhere}, there is some $p_\alpha \in B_\alpha(p^*)$ such that $\dMCE{\CC}{p_\alpha}(f) = \dIME{\CC}{p_\alpha}(f)$. By \Cref{prop:dimc-lipschitz}, $\dIMC$ is Lipschitz in the ground truth predictor, and hence
    \begin{align*}
        \dMCE{\CC}{p_\alpha}(f) &= \dIMC_{\CC,p_\alpha}(f)\\
        &\geq \dIMC_{\CC, p^*}(f) - \alpha\\
        &= 0.3 - \alpha\\
        &\geq 0.2.
    \end{align*}
    To conclude, define $\CD_0$ by drawing $\y \sim \on{Ber}(p^*(\x))$ and define $\CD_\alpha$ by drawing $\y \sim \on{Ber}(p_\alpha(\x))$. Since the marginals over $\CX$ are identical, \begin{align*}
        \on{TV}(\CD_0, \CD_\alpha) &= \frac{1}{2} \sum_{x \in \CX} \Pr[x] \left( \Big|\Pr_{\CD_0}[\y = 1 | x] - \Pr_{\CD_\alpha}[\y = 1 | x]\Big| + \Big|\Pr_{\CD_0}[\y = 0 | x] - \Pr_{\CD_\alpha}[\y = 0 | x]\Big| \right)\\
        &= \sum_{x \in \CX} \Pr[x] \Big|p^*(x) - p_\alpha(x) \Big|\\
        &= \metric{1}{p^*}{p_\alpha} \leq \alpha.
    \end{align*} 
\end{proof}

\noindent \Cref{corr:extending-no-audit-dMC} demonstrates a discontinuity in $\dMC$ analogous to that of \Cref{prop:no-audit-dMC}, confirming that the information-theoretic barrier to auditing this measure do not disappear asymptotically, even with additional structural assumptions on $\CC$.

We close this section by relating all of the notions thus far introduced.
\begin{proposition}
    For $\CD \in \Delta(\CX \times \CY)$, $\CC \subset 2^\CX$, and $f \in \predictors$, we have the following hierarchy:
    \begin{align*}
        \wdMC_{\CD, \CC}(f) \leq \dMC_{\CD, \CC}(f) \leq \cdMC_{\CD, \CC}(f) = \dIMC_{\CD, \CC}(f) = \dMC_{\CD, \CJ(\CC)}.
    \end{align*}
\end{proposition}

\subsection{Auditability of $\cdMC$ and $\dIMC$}
\label{subsec:auditability-cdMC}

Though Theorems~\ref{prop:cdmc-lipschitz} and \ref{prop:dimc-lipschitz} show that testing $\cdMC$ may be tractable in restricted contexts, it is not hard to construct instances on which computing this notion --- or even distinguishing between the case where $\cdMC=0$ or $\cdMC \geq 0.5$ --- requires $\exp(k)$ samples. Such instances arise due to the same mechanisms that make \emph{intersectionality} guarantees in algorithmic fairness difficult to audit \citep{himmelreich2025intersectionality}.
Consider, for example, a case in which $\CJ(\CC)$ is exponentially large in $k = |\CC|$. On such an instance, auditing $\cdMC$ may require more knowledge of $p^*$ than is statistically tractable. We demonstrate this fact in the following result.

\begin{proposition}\label{prop:no-audit-dimc}
    Let $k \geq 2$. There exists a domain $\CX$, subgroup collection $\CC = \{S_1, \ldots, S_k\}$, predictor $f \in \predictors$, and distributions $\{\CD_i\}_0^N \subset \Delta(\CX \times \CY)$ such that: \begin{enumerate}
        \item $\cdMC_{\CC, \CD_0}(f) = 0$ and $\cdMC_{\CC, \CD_i}(f) = 0.5$ for all $i \geq 1$;
        \item For $m < 2^{k-2}$, we have $\on{TV}(\CD_0^{\otimes m}, \sum\limits_{i=1}^N \frac{1}{N} \CD_i^{\otimes m}) \in O(m^2/2^k)$.
    \end{enumerate}
\end{proposition}

\begin{proof}
    For chosen $k$, let $\CX = \{0,1\}^{k-1}$ and $\CC = \{S_i\}_1^{k-1} \cup \{\{0\}\}$ for $S_i := \{(x_1, \cdots, x_{k-1}) \in \CX : x_i = 1\}$. 
    Let $f(x) = 0.5$ everywhere.
    Let $\CD_\x$ be uniform over $\CX$. Define $\CD_0 = \CD_\x \otimes \on{Ber}(0.5)$. Next, for each $T \subset \CX$ of cardinality $2^{k-2}$, take $p_T(x) = \mathbbm{1}[x \in T]$ and define $\CD_S$ by drawing $\y \sim \on{Ber}(p_T(\x))$. We index these distributions $\CD_i$ for $i \in [N]$ where $N:=\binom{n}{n/2}$, and $n := 2^{k-1}$. One observes that $\CJ(\CC)$ is the set of all $2^{k-1}$ singletons, and hence, $\mcalset_{\CC}(\CD)$ consists of only the ground truth $p^*$. It follows that $\cdMC_{\CC, \CD_0}(f) = \metric{1}{f}{0.5} = 0$ and $\cdMC_{\CC, \CD_i}(f) = 0.5$ for all $i \geq 1$.
    
    Next, note that by birthday paradox, a sequence of samples $((\x_j, \y_j))_1^m$ admits repeats with probability at most $m(m-1)/2n$. Hence, \begin{align*}
        \on{TV}\left(\CD_0^{\otimes m}, \sum_{i} \frac{1}{N} \CD_i^{\otimes m}\right) &\leq \frac{m(m-1)}{2n} + \frac{1}{2} \sum_{\substack{((x_j, y_j))_1^m\\ x_j \ \text{distinct}}} \left| \prod_{j=1}^{m} \Pr_{\CD_0}[x_j, y_j] - \sum_{i=1}^{N} \frac{1}{N} \prod_{j=1}^{m} \Pr_{\CD_i}[x_j, y_j]\right|\\
        &= \frac{m(m-1)}{2n} + \frac{1}{2} \sum_{\substack{((x_j, y_j))_1^m\\ x_j \ \text{distinct}}} \left( \prod_{j=1}^{m} \Pr_{\CD_\x}[x_j] \right) \left| \prod_{j=1}^{m} \Pr_{\CD_0}[y_j|x_j] - \sum_{i=1}^{N} \frac{1}{N} \prod_{j=1}^{m} \Pr_{\CD_i}[y_j | x_j]\right|.
    \end{align*}
    Note that for a sequence $((x_j, y_j))_1^m$ with all distinct $x_j$, there are exactly $\binom{n-m}{n/2 - m}$ sets $S \subset \CX$ of cardinality $n / 2$ that contain all $x_j$. Hence, by definition of the $\CD_i$, \begin{align*}
        \sum_{i=1}^{N} \frac{1}{N} \prod_{j=1}^{m} \Pr_{\CD_i}[x_j, y_j] = \frac{\binom{n-m}{n/2-m}}{\binom{n}{n/2}} = \prod_{\ell = 0}^{m-1} \frac{\frac{n}{2} - \ell}{n-\ell}.
    \end{align*}
    Moreover, \begin{align*}
        \prod_{\ell = 0}^{m-1} \frac{\frac{n}{2} - \ell}{n-\ell} = 2^{-m} \prod_{\ell=0}^{m-1} \left(\frac{1 - 2\ell/n}{1-\ell/n}\right) \geq 2^{-m} \left(1- \frac{2(m-1)}{n}\right)^m.
    \end{align*}
    By the fact that the probability of all the $x_j$'s being distinct is trivially less than one, Fixed $((x_j, y_j))_1^m$ on the right hand side such thateach $x_j$ is distinct, we have the following inequality:
    \begin{align*}
    \frac{1}{2} \sum_{\substack{((x_j, y_j))_1^m\\ x_j \ \text{distinct}}} \left( \prod_{j=1}^{m} \Pr_{\CD_\x}[x_j] \right) \left| \prod_{j=1}^{m} \Pr_{\CD_0}[y_j|x_j] - \sum_{i=1}^{N} \frac{1}{N} \prod_{j=1}^{m} \Pr_{\CD_i}[y_j | x_j]\right| &\leq \left| \prod_{j=1}^{m} \Pr_{\CD_0}[y_j|x_j] - \sum_{i=1}^{N} \frac{1}{N} \prod_{j=1}^{m} \Pr_{\CD_i}[y_j | x_j]\right|.
    \end{align*}
    Since $\Pr\limits_{\CD_0}[\y = 1| \x] = 0.5$, \begin{align*}
        \on{TV}\left(\CD_0^{\otimes m}, \sum_{i} \frac{1}{N} \CD_i^{\otimes m}\right) &\leq \frac{m(m-1)}{2n} + \left|2^{-m} - \prod_{\ell = 0}^{m-1} \frac{\frac{n}{2} - \ell}{n-\ell}\right|\\
        &\leq \frac{m(m-1)}{2n} + 2^{-m} \left|1 - \left(1- \frac{2(m-1)}{n} \right)^m \right|\\
        &\leq \frac{m(m-1)}{2n} + 2^{-m+1} \frac{m(m-1)}{n}. &\text{(Bernoulli)}
    \end{align*}
    To conclude, there is an absolute constant $C > 0$ such that $\on{TV}(\CD_0^{\otimes m}, \sum_{i} \frac{1}{N} \CD_i^{\otimes m}) \leq Cm^2 / 2^k$.
\end{proof}

\noindent \Cref{prop:no-audit-dimc} exhibits a statistical barrier to auditing $\cdMC$ when limited assumptions are made on how the ground truth $p^*$ is chosen. Suppose $\CD$ is chosen with probability $1/2$ to be $\CD_0$, and with probability $1/2$ is chosen uniformly from $\{\CD_i\}_1^n$. The TV bound in \Cref{prop:no-audit-dimc} implies that auditing in this setting with constant advantage requires a number of samples exponential in $k$. We leave it to future work to examine reasonable constraints on $p^*$  under which auditing can be done with efficient sample complexity (\emph{e.g.} restriction to classes of bounded VC dimension). 

\begin{remark}
We remark that in the above proof, the difference between $\cdMC_{\CC, \CD_0}(f) = 0$ and $\cdMC_{\CC, \CD_i}(f) = 0.5$ corresponds to distinguishing between the cases where the constant predictor $f(x)=0.5$ correctly estimates that the ground truth $p^*$ is $0.5$ on every $x$, or $f$ is incorrect and $p^*$ is always 0 or 1 on each $x$.
In turn, these two cases model all uncertainty in $f$ being \emph{aleatoric} vs. \emph{epistemic} respectively.
Aleatoric uncertainty is \emph{irreducible} uncertainty in the data.
On the other-hand, epistemic uncertainty is --- also known as ``model uncertainty'' or reducible uncertainty --- captures deficiencies in the modeling capabilities of the predictor.
As discussed in \citet{ahdritz2024higherorder}, predictors trained from only samples of the form $(\x,\y)$ may not be able to distinguish between these two types of uncertainties.
\citet{ahdritz2024higherorder} therefore propose \emph{higher order predictors}, which \emph{are} able to distinguish between aleatoric and epistemic uncertainty since they are trained with ``$k$-snapshots'': samples of the form $(\x, \y_1, \y_2, \dots, \y_k)$ constructed by first sampling $\x \sim \CD_\x$, then sampling $\y_i \overset{\text{i.i,d}}{\sim} p^*(x)$ from the ground truth labeling function $p^*$.
This represents, for example, asking multiple doctors to label a patient X-Ray as containing a broken bone or not.
Such an approach may provide workarounds of roadblocks within multicalibration auditing.
We leave exploration to future work.
\end{remark}

Next, we exhibit a loose sample complexity upper bound for estimating $\cdMC$ when $\CJ(\CC)$ is not too large and none of its elements have too small of a probability mass.
We start by estimating a weighted version of $\dCE$ for conditionals $\restr{\CD}{S}$. The following lemma uses tools developed in \citet{blasiok2022unifying}, and we leave its proof to \Cref{appx:dCE-auditability}.

\begin{lemma} \label{lem:dCE-sample-guarentee}
    Let $\CD \in \Delta(\CX \times \CY)$, $S \in 2^\CX \setminus \emptyset$, $f \in \predictors$, and $\eps, \delta > 0$. There is an estimator $\hat \mu$ such that, given a sequence of $m = O(\eps^{-2} log(1/\delta))$ samples $((f(\x_j), \y_j))_{j=1}^m$ for i.i.d. $(\x_j, \y_j) \sim \restr{\CD}{S}$, it holds with failure probability at most $\delta$ that \begin{align}\label{eq:dCE-estimate}
        \hat{\mu} - \eps \leq \dCE_{\restr{\CD}{S}}(f) \leq 4\sqrt{\hat{\mu} + \eps}.
    \end{align}
\end{lemma}

\begin{proposition}
\label{prop:dIMC-auditing}
    Let $\CD \in \Delta(\CX \times \CY)$, $f \in \predictors$, and $\CC \subset 2^\CX \setminus \emptyset$. Suppose $\CJ(\CC) =\{S_1, \ldots, S_\ell\}$ and $\gamma = \min_{S \in \CJ(\CC)} \Pr[S] > 0$. Let $\eps \in (0, \gamma]$ and $\delta > 0$. There is an estimator $\hat{\theta}$ such that, given a sequence of $m = O( \ell^4 \eps^{-2} \gamma^{-1} \log(\ell/\delta))$ samples $(f(\x_j), \y_j, (\mathbbm{1}[\x_j \in S_i])_{i=1}^{\ell})_{j=1}^m$ for i.i.d. $(\x_j, \y_j) \sim \CD$, it holds with failure probability at most $\delta$ that \begin{align*}
        \hat{\theta} - \eps \leq \cdMC_\CC(f) \leq 4\sqrt{\ell \hat{\theta}} + \sqrt{\eps}.
    \end{align*}
\end{proposition}

\begin{proof}
    Let $w = C\eps^{-2}\log(\ell/\delta)$ for $C > 0$. Drawing $O((w + \log(\ell/\delta))/\gamma)$ independent samples from $\CD$ ensures we see at least $w$ samples from each conditional $\restr{\CD}{S_i}$ with failure probability at most $\sum_{i=1}^\ell \delta/\ell = \delta$. Choosing $C$ appropriately and invoking \Cref{lem:dCE-sample-guarentee}, we obtain estimates $\hat{\mu}_i$ of $\mu_i := \dCE_{\restr{\CD}{S_i}}(f)$ for $i \in [\ell]$, and with failure probability at most $\sum_{i=1}^\ell \delta / \ell = \delta$, all of these estimates satisfy \Cref{eq:dCE-estimate}.
    By Hoeffding, we obtain estimates $\hat{p}_i$ of $p_i := \Pr[S_i]$ to error $\eps$ for all $i \in [\ell]$, with failure probability $\sum_{i=1}^\ell \delta/\ell = \delta$ using $O(\eps^{-2}\log(\ell/\delta))$ samples. Hence, with failure probability at most $3\delta$, we have the following bound. Since $\eps \leq \gamma$,
    \begin{align*}
        \sum_i p_i \mu_i &\geq \sum_i (\hat{p}_i - \eps)(\hat{\mu}_i - \eps)\\
        &\geq \left(\sum_i \hat{p}_i \hat{\mu}_i\right) - 2\ell\eps.
    \end{align*}
    Moreover, taking $\hat{\mu}_i \in [0,1]$ without loss of generality, \begin{align*}
        \sum_i p_i \mu_i &\leq 4 \sum_i (\hat{p}_i + \eps) \sqrt{\hat{\mu}_i + \eps}\\
        &\leq 4 \sum_i \hat{p}_i \sqrt{\hat{\mu}_i} + 4\sum_i \left(\eps \sqrt{\hat{\mu}_i} + \sqrt{\eps}\hat{p}_i + \eps^{3/2} \right)\\
        &\leq 4 \sum_i \sqrt{\hat{p}_i \hat{\mu}_i} + 4 \sqrt{\eps} \sum_i \left( \sqrt{\eps \hat{\mu}_i} + \hat{p}_i + \eps \right)\\
        &\leq 4\sqrt{\ell} \sqrt{\sum_i \hat{p}_i \hat{\mu}_i} + 12\ell\sqrt{\eps}. &\text{(Cauchy-Schwarz)}
    \end{align*}
    Updating parameters, it suffices to take $O( \ell^4 \eps^{-2} \gamma^{-1} \log(\ell/\delta))$ samples so that with failure probability at most $\delta$, \begin{align*}
        \sum_i \hat{p}_i \hat{\mu}_i - \eps \leq \cdMC_\CC(f) \leq 4\sqrt{\ell}\sqrt{\sum_i \hat{p}_i \hat{\mu}_i} + \sqrt{\eps}.
    \end{align*}
\end{proof}

\section{Auditability of Distance to Multiaccuracy}
\label{sec:dMA-discussion}

As seen in \Cref{subsec:auditability-cdMC}, it can be \emph{statistically} hard to audit $\dIMC$ or $\cdMC$ in $\poly(k)$ samples.
The natural next question is: \emph{what is the minimal relaxation of multicalibration which is statistically feasible to audit?}
In this section, we consider whether the weakest multi-group notion, multiaccuracy \citep{kim2019multiaccuracy}, has an auditable ``distance to'' version.
In \Cref{subsec:defining-accuracy-notions}, we begin by defining the analogous versions of $\wdMC$ and $\dMC$ for multiaccuracy, denoted by $\wdMA$ and $\dMA$.
Following that, in \Cref{subsec:measuring-multiacc}, we ask if either of these notions satisfy both desiderata: (1) Local minima are global; and (2) Lipschitzness in the ground truth $p^*$.
Lastly, in \Cref{subsec:auditing-dma}, we provide a linear program which can be run to compute $\dMA$ under perfect group-wise bias estimates.

\subsection{Defining Worst-group and Distance to Multiaccuracy}
\label{subsec:defining-accuracy-notions}

Multiaccuracy is a weakening of multicalibration which asks a predictor to be \emph{unbiased} with respect to all subgroups in a collection $\CC$.
\begin{definition}[Unbiased Predictor]
    A predictor $f \in \predictors$ is \emph{perfectly unbiased} with respect to the distribution $\CD$ if:
    \begin{align*}
        \E_{(\x, \y) \sim \CD}\Big[\y - f(\x) \Big] = 0.
    \end{align*}
    We denote the set of perfectly unbiased predictors for a distribution $\CD$ as $\accset(\CD)$.\footnote{This is done mainly to unify notation with the set of multiaccurate predictors introduced shortly.} 
    We say that a predictor $f$ is unbiased with respect to a subset $S \subseteq \CX$ if $\restr{f}{S}$ is unbiased with respect to the distribution $\restr{\CD}{S}$.
\end{definition}
\begin{definition}[Multiaccuracy]
    Given a collection of groups $\CC = \{S_1 \cdots S_k\}$ and a distribution $\CD$, a predictor $f \in \predictors$ is \emph{multiaccurate} with respect to $\CC$ if for all $i \in [k]$, $f$ is perfectly unbiased with respect to $S_i$. 
    We denote the set of perfectly multiaccurate predictors as $\maccset_\CC(\CD)$.
\end{definition}
Similarly to multicalibration, perfect multiaccuracy of $f$ does not guarantee that $f$ is unbiased with respect to the entire distribution $\CD$.
However, if this is desired, one can simply append $\CX$ to the set $\CC$.

Next, we define the distance to multiaccuracy of a predictor $f$.
This is a natural relaxation of $\dMC$ (\Cref{def:dMC}), and is defined in the identical way.
\begin{definition}[Distance to Multiaccuracy Error]
    Let a collection $\CC$, distribution $\CD$, and predictor $f \in \predictors$ be given. The \emph{distance to multiaccuracy} of $f$ is the $\normnoarg{1}$ distance from $f$ to $\maccset_\CC(\CD)$:
    \begin{align*}
        \dMA_\CC(f) := \inf_{g \in \maccset_\CC(\CD)} \metric{1}{f}{g}.
    \end{align*}
\end{definition}

We will also define the ``worst-group'' version of $\dMA$, which is analogous to $\wdMC$ for multicalibration (defined in \Cref{def:wdMC} in \Cref{subsec:measuring-multical}).

\begin{definition} \label{def:acc-error}
    The \emph{bias} of a predictor is defined as follows:
    \begin{align*}
        \dAE_\CD(f) := \Big| \E_{(\x, \y) \sim \CD}\Big[\y - f(\x) \Big] \Big|.
    \end{align*}
\end{definition}

It is not difficult to see that $\dAE_\CD(f)$ is equivalent to the $\normnoarg{1}$-distance from $f$ to $\accset(\CD)$ (we defer a proof to \Cref{appx:wdma-proof}).
This justifies the following definition of a worst-group variation on $\dMA$.

\begin{definition}
    For a predictor $f$, subset collection $\CC$, and distribution $\CD$, the worst-group bias is defined as:
    \begin{align*}
        \wdMA_\CC(f) := \max_{S \in \CC} \left(\dAE_{\restr{\CD}{S}}(f) \cdot \Pr[S] \right).
    \end{align*}
\end{definition}

\subsection{Properties of $\wdMA$ and $\dMA$}
\label{subsec:measuring-multiacc}

We are still interested in the two properties outlined throughout \cref{sec:cdmc}: (1) Local minima of the multicalibration error metric are global minima; and (2) Lipschitzness in the ground truth.

First, we show that the set of perfectly multiaccurate predictors is convex.
\begin{lemma}
    For any collection $\CC$ and distribution $\CD$, $\maccset_\CC(\CD)$ is convex.
\end{lemma}
\begin{proof}
To prove that $\maccset_{\mathcal{C}}(\mathcal{D})$ is a convex set, we must show that for any two predictors $f_1, f_2 \in \maccset_{\mathcal{C}}(\mathcal{D})$ and any $\lambda \in [0, 1]$, we have that $f_\lambda = \lambda f_1 + (1-\lambda)f_2$ is also in $\text{macc}_{\mathcal{C}}(\mathcal{D})$.

Consider $f_\lambda$ conditioned on an arbitrary $S \in \mathcal{C}$.
By linearity of expectation and the fact that $f_1, f_2 \in \maccset$, we have the following.
\begin{align*}
    \E_{(\x, \y) \sim \CD} [f_\lambda(\x) \mid \x \in S]
    &= \lambda \E_{(\x, \y) \sim \CD} [f_1(\x) \mid \x \in S] 
    + (1 - \lambda) \E_{(\x, \y) \sim \CD} [f_1(\x) \mid \x \in S] \\
    &= \lambda \E_{(\x, \y) \sim \CD} [\y \mid \x \in S] 
    + (1 - \lambda) \E_{(\x, \y) \sim \CD} [\y \mid \x \in S] \\
    &= \E_{(\x, \y) \sim \CD} [\y \mid \x \in S]
\end{align*}
Therefore, $f_\lambda$ is unbiased for any subset $S \in \CC$, and is perfectly multiaccurate.
\end{proof}
Using this, it is simple to show that $\dMA$ is a convex function, satisfying requirement (1) that local minima are global minima.
\begin{proposition} \label{prop:wdma-global-minima}
    Let a distribution $\CD$ and collection $\CC$ be given.
    Then $\dMA_\CC(f)$ is a convex function of $f$.
    Furthermore, all local minima are global, and $\dMA_\CC$ is minimized at 0.
\end{proposition}
\begin{proof}
    This follows from the fact that $\dMA$ is defined as the $\normnoarg{1}$ distance from its input to a convex set, and the fact that $p^*$ is perfectly multiaccurate and has bias 0 everywhere.
\end{proof}

\begin{proposition}
The function $\text{wdMA}_{\mathcal{C}}(f)$ is a convex function of $f$.
Furthermore, all local minima are global, and $\wdMA_\CC$ is minimized at 0.
\end{proposition}
\begin{proof}
Fix $S \in \CC$. Consider the following.
\begin{align*}
    g_S(f) 
    &:= \dAE_{\restr{\CD}{S}}(f)\cdot \Pr[S]\\
    &= \left|\E_{(\x,\y)\sim \restr{\CD}{S}} \Big[\y - f(\x)\Big]\right| \cdot  \Pr[S]\\
    &= \left| \frac{\E\limits_{(\x,\y)\sim \CD}\Big[(\y - f(\x)) \cdot \mathbbm{1}_{\x \in S}\Big]}{\Pr[S]} \right| \cdot \Pr[S]\\
    &= \Big|\E_{(\x,\y)\sim \CD}[\y \cdot \mathbbm{1}_{\x \in S}] - \E_{(\x,\y)\sim \CD}[f(\x) \cdot \mathbbm{1}_{\x \in S}]\Big|
\end{align*}
In the above, $\mathbbm{1}_{\x \in S}$ is the indicator or set membership function.
Notice that $\E_{(\x,\y)\sim \CD}[\y \cdot \mathbbm{1}_{\x \in S}]$ is independent of $f$.
Furthermore, $\E_{(\x,\y)\sim \CD}[f(\x) \cdot \mathbbm{1}_{\x \in S}]$ is a linear function of $f$. 
Therefore, $g_S(f)$ is a composition of the convex absolute value function $|\cdot|$ and an affine function, so $g_S(f)$ is convex.
By definition, we have that $\wdMA_\CC(f) = \max_{S \in \CC} g_S(f)$, and since the pointwise maximum of convex functions is convex, $\wdMA$ is convex.
The other properties follow by convexity and due to the fact that $\wdMA_\CC(p^*) = 0$.
\end{proof}

Interestingly, despite the loss landscape of $\wdMC$ having the property that all local minima are global minima, a predictor $f$ with a worst group bias of $\eps$ \emph{does not} imply closeness to a predictor with worst bias of $\eps/c$ for some constant $c \geq 6$. In particular, we will show that the distance between $f$ and a predictor $\tilde{f}$ with $\wdMA(\tilde{f}) = \eps/c$ is dependent on $\eps$ but could be on the order of $\eps \cdot \exp(k)$. 
We show this in \Cref{lemma:improving-wdma} by constructing an interesting instance based on the Fibonacci recurrence.

For completeness, recall the definition of the $i$th Fibonacci number $F_i$, where $F_0 = 0, F_1 = 1$, and for $i > 1$, $F_i = F_{i-1}+F_{i-2}$.

\begin{proposition}
\label{lemma:improving-wdma}
    Let $k>0$, $\eps \in (0, \frac{1}{2(k+1)F_{k+1}})$.
    There exists a distribution $\CD\in \Delta(\CX \times \CY)$, set of groups $S_1, \cdots, S_{2k}$, and predictor $f$ such that $\wdMA_\CC(f) = \eps$, but for \emph{any} predictor $\tilde f$ with $\wdMA_\CC(\tilde f) \leq \eps/6$ it holds that $\metric{1}{f}{\tilde f} = \Omega(\eps \cdot \phi^k)$, where $\phi \approx 1.618$ is the golden ratio. 
\end{proposition}

To give intuition for \Cref{lemma:improving-wdma}, consider the construction depicted in \Cref{fig:wdMA-Fibonacci}. 
One can verify that $f$ is unbiased with respect to all groups in $\CC$ except for $W$, where the bias is $\delta/2$, making $\wdMA_\CC(f)=\eps$ where $\eps = \frac{\delta}{2(k+1)}$. 
Any perfectly multiaccurate predictor $g$ must satisfy $|g(x_1)-f(x_1)| \geq \delta$. Moreover, if we fix $g(x_0), g(x_1), g(x_2)$, there are no remaining degrees of freedom: for $i\geq 3$, $g(x_i)$ is completely determined by recurrence relations induced by the remaining groups in $\CC$. 
These recurrence relations force $|g(x_i)-f(x_i)|$ to scale with the Fibonacci sequence, enforcing an $\eps \cdot c^k$ distance from $f$ in $\normnoarg{1}$.

If we consider $\tilde f$ to be the nearest predictor to $f$ with worst group bias of $\leq \eps/c$, and if $c$ is too small, the errors are able to accumulate in such a way that the quantity $|\tilde f(x_i)-f(x_i)|$ stays constant. In this case, we would have $\metric{1}{f}{\tilde f} = O(\eps)$. However, $c=6$ turns out to be sufficient to avoid this accumulation.

    \begin{figure}
    \centering
    \includegraphics[width=0.8\textwidth]{figures/wdMA_Fibonacci.pdf}
    \caption{The setup of $f$, $p^*$ and the groups $\{U_i\}_{i \in [k]}, \{V_i\}_{i \in [k-1]}, W$ where $k=6$. Define $\eps = \frac{\delta}{2(k+1)}$. Note that $\wdMA_\CC(f) = \eps$, and $\metric{1}{f}{p^*} = \Omega(\eps \cdot \phi^k)$.}
    \label{fig:wdMA-Fibonacci}
    \end{figure}
\begin{proof}[Proof of \Cref{lemma:improving-wdma}]
We define the instance then prove the claim.

\paragraph{Instance definition.}
    The number of subgroups $k$ is given as input.
    We let $\CX = \{x_0, x_1, \cdots, x_{2k+1}\}$. The sets $S_1 \cdots S_{2k}$ are of three types. To avoid excessive subscripts, we use different letters to describe each type of set. Let $U_i = \{x_{2i-1}, x_{2i}, x_{2i+1}\}$ for each $i \in [k]$. Let $V_i = \{x_{2i-1}, x_{2i+2}\}$ for each $i \in [k-1]$. 
    Finally, let $W = \{x_0, x_1\}$. We can assign $S_i = U_i$ for $i \in [k]$, $S_{k+i} = V_i$ for $i \in [k-1]$ and $S_{2k} = W$.
    Let the marginal distribution $\CD_\x$ be uniform with $\Pr[x] = 1/(2k+2)$ for each $x \in \CX$. 

    Define $\delta = 2(k+1)\eps$. 
    Notice that $\eps$ is small enough such that $F_{k+1} \cdot \delta < 1$.
    We define $p^*$ as follows:
    \begin{align*}
        p^*(x_0) &= 0 \\
        p^*(x_2) &= 1 \\
        p^*(x_{2i-1}) &= \frac{1-(-1)^i}{2} + (-1)^i F_i \delta \quad &\forall i \in [k+1]\\
        p^*(x_{2i+2}) &= 1-p^*(x_{2i-1}) \quad &\forall i \in [k-1]
    \end{align*}
    This is to say, $p^*$ follows the following pattern on $x_j$ for $j>2$:
    \begin{align*}
        p^*(x_4) = \delta\cdot F_{1}, \quad
        p^*(x_6) = 1-\delta\cdot F_{2}, \quad
        p^*(x_8) = \delta\cdot F_{3}, \quad \dots \quad \quad & \text{if $j$ is even}\\
        p^*(x_3) = \delta\cdot F_{2}, \quad
        p^*(x_5)=1-\delta\cdot F_{3}, \quad
        p^*(x_7)=\delta\cdot F_{4}, \quad \dots \quad \quad & \text{if $j$ is odd.}
    \end{align*}
    We define $f$ as follows. For all $0 \leq i \leq k$:
    \begin{align*}
        f(x_{2i}) = \frac{1-(-1)^i}{2}, \quad f(x_{2i+1}) = 1-f(x_{2i}).
    \end{align*}
    Equivalently, $f(x_j) = 1$ whenever we have $j \equiv 1 \;\; (\text{mod}\; 4)$ or $j \equiv 2 \;\; (\text{mod}\; 4)$, and $f(x_j) = 0$ otherwise.

    \paragraph{Computing multiaccuracy error.}
    Firstly, we verify $f$ is unbiased on each $U_i$ for $i \in [k]$. We have the following calculation:
    
    \begin{align*}
        \frac{1}{3}\Big(f(x_{2i-1})+f(x_{2i})+f(x_{2i+1})\Big) &= \frac{1}{3}\Big(1-f(x_{2(i-1)}) + f(x_{2i})+1-f(x_{2i})\Big)\\
        &= \frac{1}{3}\Big(2-f(x_{2(i-1)}) \Big)\\
        &= \frac{1}{3}\Big(2-\frac{1+(-1)^i}{2} \Big)\\
        &= \frac{1}{3}\Big(1+\frac{1-(-1)^i}{2} \Big).
    \end{align*}

    Examining $p^*$, we have $\frac{1}{3}(p^*(x_1)+p^*(x_2)+p^*(x_3)) = \frac{1}{3}(1-\delta + 1 + \delta) = \frac{2}{3}$, confirming $\E[f(\x)|U_1] = \E[p^*(\x)|U_1]$. For $1 < i \leq k$, we have
    \begin{align*}
        \frac{1}{3}\Big(p^*(x_{2i-1})&+p^*(x_{2i})+p^*(x_{2(i+1)-1})\Big)\\ 
        &= \frac{1}{3}\Big(p^*(x_{2i-1})+1-p^*(x_{2(i-1) - 1})+p^*(x_{2i+1})\Big)\\
        &= \frac{1}{3}\Big(1+ \frac{1-(-1)^i + (-1)^{i-1} - (-1)^{i+1}}{2} + \big((-1)^iF_i- (-1)^{i-1}F_{i-1}+(-1)^{i+1}F_{i+1}\big)\delta\Big)\\
        &= \frac{1}{3}\Big(1+ \frac{1-(-1)^i}{2} + \big( (-1)^iF_i+ (-1)^iF_{i-1}-(-1)^iF_{i+1}\big)\delta\Big)\\
        &= \frac{1}{3}\Big(1+ \frac{1-(-1)^i}{2} + (-1)^i\big(F_i+ F_{i-1}-F_{i+1}\big)\delta\Big)\\
        &= \frac{1}{3}\Big(1+ \frac{1-(-1)^i}{2} \Big)
    \end{align*}
    Therefore, $f$ is unbiased with respect to $U_i$ for all $i \in [k]$.

    Next, for $i \in [k-1]$, we have $\frac{1}{2}\big(f(x_{2i-1}) + f(x_{2i+2})\big) = \frac{1}{2}\big(1-f(x_{2(i-1)})+f(x_{2(i+1)})\big)=\frac{1}{2}$. Similarly for $p^*$, we have 
    $\frac{1}{2}\big( p^*(x_{2i-1}) + p^*(x_{2i+2})\big) = \frac{1}{2}\big(p^*(x_{2i-1}) + 1- p^*(x_{2i-1}) \big) = \frac{1}{2}$, so $f$ is unbiased with respect to each $V_i$. 

    Note, however, that $f$ \emph{is biased} with respect to $W$: $\frac{1}{2}(f(x_0)+f(x_1)) = \frac{1}{2}$, but $\frac{1}{2}(p^*(x_0)-p^*(x_1))=\frac{1-\delta}{2}$. Therefore we have $\dAE_{\restr{\CD}{W}}(f) = \frac{\delta}{2}$, and $\wdMA_\CC(f) = \frac{\delta}{2(k+1)} = \eps$.

    \paragraph{Induced constraints on $\tilde f$.}
    Let $\tilde f$ be a predictor such that $\wdMA_\CC(\tilde f) \leq \eps/6$. We consider $\tilde f$ as a vector in $[0,1]^{2k+2} \cong [0,1]^\CX$ and denote its entries $\tilde f_i = \tilde f(x_i)$. Observe that $\tilde f$ must satisfy the following for all $i \in [k]$ and $j \in [k-1]$.
    \begin{align}
        \frac{1}{2(k+1)}|\tilde f_0 + \tilde f_1 - (1-\delta)| &\leq \eps/6 \label{eq:W-acc-windo}\\
        \frac{1}{2(k+1)}\left|\tilde f_{2i-1} + \tilde f_{2i} + \tilde f_{2i+1} - \left(1+\frac{1-(-1)^i}{2}\right)\right| &\leq \eps/6 \label{eq:U-acc-windo}\\
        \frac{1}{2(k+1)}|\tilde f_{2j-1} + \tilde f_{2j+2} - 1| &\leq \eps/6. \label{eq:V-acc-windo}
    \end{align}

    We prove the following claim inductively: \emph{for all $i \in [2k+1] \setminus \{0, 2\}$, the following inequalities hold.
        \begin{alignat*}{2}
            \tilde f_i - f(x_i) &\geq F_{i/2-1}\cdot\delta/3 + \delta/6 &\text{if $i \equiv 0\;\; (\text{mod}\; 4)$}\\
            f(x_i) - \tilde f_i &\geq F_{(i+1)/2}\cdot \delta/3 + \delta/3 &\quad\quad\text{if $i \equiv 1\;\; (\text{mod}\; 4)$}\\
            f(x_i) - \tilde f_i &\geq F_{i/2-1}\cdot\delta/3 + \delta/6 &\text{if $i \equiv 2\;\; (\text{mod}\; 4)$}\\
            \tilde f_i - f(x_i) &\geq F_{(i+1)/2}\cdot \delta/3 + \delta/3 &\text{if $i \equiv 3\;\; (\text{mod}\; 4)$}&.
        \end{alignat*}
    }
    We check the claim manually for $\tilde f_1$ and $\tilde f_3$, then induct.

    \textbf{Base Case:}
    From \Cref{eq:W-acc-windo}, we have 
    \begin{align*}
        \tilde f_1 \leq \tilde f_0 +
        \tilde f_1 &\leq 1 - \delta + \delta/6 = 1-5\delta/6\\
        1-\tilde f_1 &\geq 5\delta/6 \\
        1-\tilde f_1 &\geq F_1 \cdot \delta/3 + \delta/3 \\
    \end{align*}
    From $\eqref{eq:U-acc-windo}$, specifically by unbiasedness with respect to $U_1$, we have
    \begin{align*}
        \tilde f_{1} + \tilde f_{2} + \tilde f_{3} &\geq 2 - \delta/6\\
        1-5\delta/6 + \tilde f_{2} + \tilde f_{3} &\geq 2- \delta/6\\
        \tilde f_2 + \tilde f_3 & \geq 1 + 2\delta/3 \\
        \tilde f_3 &\geq 2\delta/3 \\
        \tilde f_3-0 &\geq F_2 \cdot \delta/3 + \delta/3.
    \end{align*}
    Therefore, the claim holds for $i = 1, 3$. For clarity, proceed with some $j \geq 4$.

    \textbf{Case 1: $j \equiv 0 \;\;(\text{mod}\; 4)$}: Note that $f(x_j) = 0$ and $f(x_{j-3}) = 1$. Since $j - 3 \geq 1$ and $j - 3\equiv 1 \; (\text{mod} \; 4)$, the inductive hypothesis gives $\tilde f_{j-3} \leq 1-F_{j/2-1}\cdot \delta/3 - \delta/3$. From \Cref{eq:V-acc-windo} specifically by unbiasedness with respect to $V_{j/2-1}$, we have
    \begin{align*}
        \tilde f_{j-3} + \tilde f_j &\geq 1 - \delta/6\\
        1-F_{j/2-1}\cdot \delta/3 - \delta/3 + \tilde f_j & \geq 1 - \delta/6\\
        \tilde f_j -0 &\geq F_{j/2-1}\cdot \delta/3 + \delta/6.
    \end{align*}

    \textbf{Case 2: $j \equiv 1 \;\;(\text{mod}\; 4)$}: Note that $f(x_{j-2})=f(x_{j-1}) = 0$ and $f(x_j) =1$. Since $j-2 \geq 3$, the inductive hypothesis gives $\tilde f_{j-2} \geq F_{(j-1)/2}\cdot \delta/3 + \delta/3$ and $\tilde f_{j-1} \geq F_{(j-1)/2-1} \cdot \delta/3 + \delta/6$. By \Cref{eq:U-acc-windo} on $U_{(j-1)/2}$, we have 
    \begin{align*}
        \tilde f_{j-2} + \tilde f_{j-1} + \tilde f_j &\leq \left(1+\frac{1-(-1)^{(j-1)/2}}{2}\right) + \delta/6\\
         \tilde f_{j-2} + \tilde f_{j-1} + \tilde f_j &\leq 1+\delta/6\\
         F_{(j-1)/2} \cdot \delta/3 + \delta/3 + F_{(j-1)/2-1} \cdot \delta/3 + \delta/6 + \tilde f_j &\leq 1+\delta/6\\
         \tilde f_j &\leq 1 - (F_{(j-1)/2} + F_{(j-1)/2-1})\cdot\delta/3- \delta/3\\
         1-\tilde f_j &\geq F_{(j-1)/2+1} \cdot \delta/3 + \delta/3\\
         1-\tilde f_j &\geq F_{(j+1)/2} \cdot \delta/3 + \delta/3.
    \end{align*}

    \textbf{Case 3: $j \equiv 2 \;\;(\text{mod}\; 4)$}: Note that $f(x_j) = 1$ and $f(x_{j-3}) = 0$. Since $j -3 \geq 3$, the inductive hypothesis gives $\tilde f_{j-3} \geq F_{(j-2)/2} \cdot \delta/3 + \delta/3$. From \Cref{eq:V-acc-windo}, specifically by unbiasedness with respect to $V_{j/2-1}$, we have
    \begin{align*}
        \tilde f_{j-3} + \tilde f_j &\leq 1 + \delta/6\\
        F_{(j-2)/2} \cdot \delta/3 + \delta/3 + \tilde f_j & \leq 1 + \delta/6\\
        \tilde f_j &\leq 1-F_{(j-2)/2} \cdot \delta/3-\delta/6\\
        1- \tilde f_j &\geq F_{j/2-1}\cdot \delta/3 + \delta/6.
    \end{align*}

    \textbf{Case 4: $j \equiv 3 \;\;(\text{mod}\; 4)$}: Note that $f(x_{j-2})=f(x_{j-1}) = 1$ and $f(x_j) =0$. Since $j-2 \geq 5$, the inductive hypothesis gives $\tilde f_{j-2} \leq 1-F_{(j-1)/2}\cdot \delta/3 - \delta/3$ and $\tilde f_{j-1} \leq 1-F_{(j-1)/2-1}\cdot \delta/3 - \delta/6$. By \Cref{eq:U-acc-windo} on $U_{(j-1)/2}$, we have 
    \begin{align*}
        \tilde f_{j-2} + \tilde f_{j-1} + \tilde f_j &\geq \left(1+\frac{1-(-1)^{(j-1)/2}}{2}\right) - \delta/6\\
         \tilde f_{j-2} + \tilde f_{j-1} + \tilde f_j &\geq 2-\delta/6\\
         2-F_{(j-1)/2}\cdot \delta/3 -\delta/3 - F_{(j-1)/2-1} \cdot \delta/3 - \delta/6 + \tilde f_j &\geq 2-\delta/6\\
         \tilde f_j &\geq  (F_{(j-1)/2} + F_{(j-1)/2-1})\cdot \delta/3 + \delta/3\\
         \tilde f_j - 0 &\geq  F_{(j+1)/2}\cdot \delta/3 + \delta/3.
    \end{align*}
    This completes the proof of our claim. 

    \paragraph{Lower bound on $\metric{1}{f}{\tilde f}$.}
    We use the claim above. Notice that 
    \begin{align*}
        \metric{1}{f}{\tilde f} \geq \frac{1}{2k+2}|f(x_{2k+1}) - \tilde f_{2k+1}| \geq \frac{F_{k+1} \cdot\delta/3}{2k+2} = \frac{1}{3}F_{k+1}\eps.
    \end{align*}
    Recall that we have the formula $F_i = \frac{\phi^i - (-\phi)^{-i}}{\sqrt{5}}$, where $\phi=\frac{1+\sqrt{5}}{2} \approx 1.618$. 
    It follows that $\metric{1}{f}{\tilde f} = \Omega(\eps  \cdot \phi^k)$.
\end{proof}

\subsection{A Linear Program for Computing $\dMA$}
\label{subsec:auditing-dma}

Notice that the \emph{bias} of a predictor is a linear constraint.
Furthermore, the set $\maccset(\CD)$ is always convex.
It is therefore natural to expect that we can compute $\dMA(f)$ using a linear program.

\begin{proposition}\label{prop:dMA_LP}
    Let $\CX = \{x_i\}_1^n$, $\CC = \{S_j\}_1^k$, and $\CD \in \Delta(\CX \times \CY)$. The following linear program computes $\dMA_\CC(f)$. Let $p := (\Pr[x_i])_i$, $f := (f(x_i))_i$, where $i \in [n]$, and define the matrix $A_j \in \R^{n \times n}$ so that its $(p, q)$th entry is 1 if $p=q$ and $x_p \in S_j$, and zero otherwise. 
    \begin{align}
        \min_{g, t} \quad &p^{\top} t\\
        \text{subject to} \quad &(A_jp)^{\top} g = \Pr[S_j] \cdot \E[\y | S_j], &\forall j \in [k]\label{eq:accurate_constraint}\\
        &t_i \geq g_i - f_i,\label{eq:abs_constraint_1}\\
        &t_i \geq f_i - g_i,\label{eq:abs_constraint_2}\\
        &g_i \geq 0,\label{eq:predictor_constraint_1}\\
        &g_i \leq 1, &\forall i \in [n].\label{eq:predictor_constraint_2}
    \end{align}
\end{proposition}

\begin{proof}
    By \Cref{eq:predictor_constraint_1}, \Cref{eq:predictor_constraint_2}, $g$ has its image in $[0,1]$. \Cref{eq:accurate_constraint} ensures $g$ is unbiased with respect to all sets. Conversely, a multiaccurate predictor $g$ satisfies all constraints on $g$ having its image in $[0,1]$ and satisfying \Cref{eq:accurate_constraint} for all $S\in \CC$.

    By \Cref{eq:abs_constraint_1}, \Cref{eq:abs_constraint_2} we have $t_i\geq|f_i-g_i|$, the difference between $f$ and $g$ at index $i$. Hence, by definition of $p$, it holds that $p^{\top}t \geq \metric{1}{f}{g}$ for any feasible $t, g$. Since the set of feasible $g$ is precisely the set of multicaccurate predictors, the optimum is $\dMA_\CC(f)$.
\end{proof}

\noindent \Cref{prop:dMA_LP} demonstrates that with perfect information on all $\Pr[S_i]$ and $\E[\y | S_i]$, and access to group membership indicators, one can $\dMA_\CC(f)$ with relative simplicity: one only needs $O(n)$ variables and $O(n + k)$ constraints. We point out that it is sensitive to error in these constraints, however. Specifically, recall the Fibonacci example from \Cref{lemma:improving-wdma}, in which $\wdMA(f) = \eps$ but $\metric{1}{f}{g} = \Omega(\eps c^k)$ for any perfectly multiaccurate $g$. In particular, we have \begin{align*}
    \left| \E_{\restr{\CD}{S_i}}[f(\x)] - \E_{\restr{\CD}{S_i}}[\y] \right| \leq \eps, \quad i \in [k].
\end{align*}
Under noisy estimates, one may select $g = f$ as a feasible point in the linear program above, and estimate $\dMA_\CC(f)$ to error $\Omega(\eps c^k)$.

\section{Conclusions and Future Work}

Our work broadly demonstrates barriers to auditing the distance to multicalibration notions.
However, it also reveals that in certain restricted settings which may be useful to practitioners, distance to multicalibration \emph{can} be audited (\Cref{subsec:auditability-cdMC}).
Indeed, the fact that the continuized distance to mutlicalibration error $\cdMC$ introduced and developed throughout \Cref{sec:cdmc} is exactly measuring an \emph{intersectional} \citep{himmelreich2024intersectionality} analogue of distance to multicalibration error is surprising and useful.
However, this also means that for any setting with many subgroups in the collection $\CC$, auditing the continuized distance is difficult.

The instance used to exhibit in-auditability of $\cdMC$ in \Cref{prop:no-audit-dimc} seems wholly due to the fact that the ground-truth has, in some sense, a ``maximal'' complexity.
In particular, one can ask if ``simpler'' ground truths $p^*$ give rise to positive auditability results.
One can imagine restricting the allowable ground truth distributions to, for example, those with low VC dimension.
This complexity restriction may demonstrate other settings where positive auditability results emerge.

In \Cref{appx:generalizing-dMC}, we demonstrate that information-theoretic barriers exist to auditing both distance to calibrated multiaccuracy \citep{casacuberta2025global} and low-degree multicalibration \citep{gopalan2022low}.
There are still yet unexplored ways to relax the full distance to multicalibration.
For example, we could consider the distance from $f$ to the set of \emph{nearly} perfectly multicalibrated predictors.
Alternatively, one could work with the ECE of restrictions of $f$ directly, as bucketing or discretization is common within the multicalibration community.
These approaches may allow for auditability in more general or straightforward ways.

Our preliminary foray into distance to \emph{multiaccuracy} also reveals interesting subtleties.
In particular, in \Cref{lemma:improving-wdma}, we show that the worst-group bias quantity $\wdMA$ exhibits the following interesting property: 
Even if $\wdMA_\CC(f) = \eps$, the predictor $f$ may need to change by some amount \emph{exponential} in the collection size $k$ in order to reach some $\tilde{f}$ with $\wdMA_\CC(\tilde f) \leq \eps/6$.
This holds in spite of all local minima of $\wdMA$ being global minima; this suggests that there can be large ``basins'' with low slope near the minima of the loss surface of $\wdMA$. 
Taken together, this demonstrates that $\dMA$ and $\wdMA$ are indeed different quantities, and suggests further motivation for finding a statistically and computationally efficient way to audit $\dMA$.

More generally, auditing for $\dMA$ may require a better understanding the geometry of the space of multiaccurate predictors.
It can be shown that the set $\maccset(\CD)$ is affine subspace of $[0,1]^\CX$, which lends some nice properties.
For example, if $f$ is perfectly multiaccurate with respect to the ground truth $p^*$, then $p^*$ is perfectly multiaccurate with respect to the distribution given by a ground truth with $f$. The same cannot be said for multicalibration.

Lastly, extending to settings beyond finite $\CX$ is especially interesting since the sets of calibrated or multicalibrated predictors would no longer be finite, an important assumption used throughout our proofs.

\subsection*{Acknowledgments and Funding}
We thank Parikshit Gopalan and Charlotte Peale for helpful discussions during early stages of this work.  This work was supported in part by an NSF CAREER Award CCF-2239265, an Amazon Research Award, a Google Research Scholar Award, and a Okawa Foundation Research Grant.

\newpage

\bibliography{refs}

\begin{thebibliography}{}

\bibitem[Ahdritz et~al., 2025]{ahdritz2024higherorder}
Ahdritz, G., Gollakota, A., Gopalan, P., Peale, C., and Wieder, U. (2025).
\newblock Provable uncertainty decomposition via higher-order calibration.
\newblock In {\em The Thirteenth International Conference on Learning Representations}.

\bibitem[Bequ{\'e} et~al., 2017]{beque2017approaches}
Bequ{\'e}, A., Coussement, K., Gayler, R., and Lessmann, S. (2017).
\newblock Approaches for credit scorecard calibration: An empirical analysis.
\newblock {\em Knowledge-Based Systems}, 134:213--227.

\bibitem[Bharti et~al., 2025]{bharti2025multiaccuracy}
Bharti, B., Clemens-Sewall, M.~V., Yi, P.~H., and Sulam, J. (2025).
\newblock Multiaccuracy and multicalibration via proxy groups.
\newblock In {\em Forty-second International Conference on Machine Learning}.

\bibitem[B{\l}asiok et~al., 2023]{blasiok2022unifying}
B{\l}asiok, J., Gopalan, P., Hu, L., and Nakkiran, P. (2023).
\newblock A unifying theory of distance from calibration.
\newblock In {\em Proceedings of the 55th Annual ACM Symposium on Theory of Computing}.

\bibitem[Br{\"o}cker, 2009]{brocker2009reliability}
Br{\"o}cker, J. (2009).
\newblock Reliability, sufficiency, and the decomposition of proper scores.
\newblock {\em Quarterly Journal of the Royal Meteorological Society: A journal of the atmospheric sciences, applied meteorology and physical oceanography}, 135(643):1512--1519.

\bibitem[Błasiok and Nakkiran, 2024]{blasiok2023smooth}
Błasiok, J. and Nakkiran, P. (2024).
\newblock Smooth ece: Principled reliability diagrams via kernel smoothing.
\newblock In {\em The Twelfth International Conference on Learning Representations}.

\bibitem[Casacuberta et~al., 2025]{casacuberta2025global}
Casacuberta, S., Gopalan, P., Kanade, V., and Reingold, O. (2025).
\newblock How global calibration strengthens multiaccuracy.
\newblock {\em arXiv preprint arXiv:2504.15206}.

\bibitem[Collina et~al., 2025]{collina2025collaborative}
Collina, N., Globus-Harris, I., Goel, S., Gupta, V., Roth, A., and Shi, M. (2025).
\newblock Collaborative prediction: Tractable information aggregation via agreement.
\newblock {\em arXiv preprint arXiv:2504.06075}.

\bibitem[Dahabreh et~al., 2017]{dahabreh2017review}
Dahabreh, I.~J., Chan, J.~A., Earley, A., Moorthy, D., Avendano, E.~E., Trikalinos, T.~A., Balk, E.~M., and Wong, J.~B. (2017).
\newblock A review of validation and calibration methods for health care modeling and simulation.
\newblock {\em Modeling and Simulation in the Context of Health Technology Assessment: Review of Existing Guidance, Future Research Needs, and Validity Assessment [Internet]}.

\bibitem[Detommaso et~al., 2024]{detommaso2024multicalibration}
Detommaso, G., Bertran, M., Fogliato, R., and Roth, A. (2024).
\newblock Multicalibration for confidence scoring in llms.
\newblock In {\em International Conference on Machine Learning}.

\bibitem[Devic et~al., 2024]{devic2024stability}
Devic, S., Korolova, A., Kempe, D., and Sharan, V. (2024).
\newblock Stability and multigroup fairness in ranking with uncertain predictions.
\newblock In {\em International Conference on Machine Learning}.

\bibitem[Dwork et~al., 2025]{dwork2024fairness}
Dwork, C., Hays, C., Immorlica, N., Perdomo, J.~C., and Tankala, P. (2025).
\newblock From fairness to infinity: Outcome-indistinguishable (omni) prediction in evolving graphs.
\newblock In {\em The Thirty Eighth Annual Conference on Learning Theory}.

\bibitem[Gohar and Cheng, 2023]{gohar2023survey}
Gohar, U. and Cheng, L. (2023).
\newblock A survey on intersectional fairness in machine learning: notions, mitigation, and challenges.
\newblock In {\em Proceedings of the Thirty-Second International Joint Conference on Artificial Intelligence}, pages 6619--6627.

\bibitem[Gollakota et~al., 2025]{gollakota2025loss}
Gollakota, A., Gopalan, P., Karan, A., Peale, C., and Wieder, U. (2025).
\newblock When does a predictor know its own loss?
\newblock In Bun, M., editor, {\em 6th Symposium on Foundations of Responsible Computing, {FORC} 2025, June 4-6, 2025, Stanford University, CA, {USA}}, volume 329 of {\em LIPIcs}, pages 22:1--22:22.

\bibitem[Gollakota et~al., 2024]{gollakota2024agnostically}
Gollakota, A., Gopalan, P., Klivans, A., and Stavropoulos, K. (2024).
\newblock Agnostically learning single-index models using omnipredictors.
\newblock In {\em Advances in Neural Information Processing Systems}, volume~36.

\bibitem[Gopalan et~al., 2023a]{gopalan2023lossmin}
Gopalan, P., Hu, L., Kim, M.~P., Reingold, O., and Wieder, U. (2023a).
\newblock Loss minimization through the lens of outcome indistinguishability.
\newblock In Kalai, Y.~T., editor, {\em 14th Innovations in Theoretical Computer Science Conference, {ITCS} 2023, January 10-13, 2023, MIT, Cambridge, Massachusetts, {USA}}.

\bibitem[Gopalan et~al., 2022a]{gopalan2021omnipredictors}
Gopalan, P., Kalai, A.~T., Reingold, O., Sharan, V., and Wieder, U. (2022a).
\newblock Omnipredictors.
\newblock In {\em Innovations in Theoretical Computer Science}.

\bibitem[Gopalan et~al., 2023b]{gopalan2023characterizing}
Gopalan, P., Kim, M.~P., and Reingold, O. (2023b).
\newblock Characterizing notions of omniprediction via multicalibration.
\newblock In {\em Advances in Neural Information Processing Systems}.

\bibitem[Gopalan et~al., 2022b]{gopalan2022low}
Gopalan, P., Kim, M.~P., Singhal, M.~A., and Zhao, S. (2022b).
\newblock Low-degree multicalibration.
\newblock In {\em Conference on Learning Theory}, pages 3193--3234. PMLR.

\bibitem[Guo et~al., 2017]{guo2017calibration}
Guo, C., Pleiss, G., Sun, Y., and Weinberger, K.~Q. (2017).
\newblock On calibration of modern neural networks.
\newblock In {\em International conference on machine learning}, pages 1321--1330. PMLR.

\bibitem[Guy et~al., 2025]{guy2025measuring}
Guy, I., Haimovich, D., Linder, F., Okati, N., Perini, L., Tax, N., and Tygert, M. (2025).
\newblock Measuring multi-calibration.
\newblock {\em arXiv preprint arXiv:2506.11251}.

\bibitem[Haghtalab et~al., 2023]{haghtalab2024unifying}
Haghtalab, N., Jordan, M., and Zhao, E. (2023).
\newblock A unifying perspective on multi-calibration: Game dynamics for multi-objective learning.
\newblock In {\em Advances in Neural Information Processing Systems}, volume~36.

\bibitem[Hansen et~al., 2024]{hansen2024multicalibration}
Hansen, D., Devic, S., Nakkiran, P., and Sharan, V. (2024).
\newblock When is multicalibration post-processing necessary?
\newblock In {\em Advances in Neural Information Processing Systems}.

\bibitem[H{\'e}bert-Johnson et~al., 2018]{hebert2018multicalibration}
H{\'e}bert-Johnson, U., Kim, M., Reingold, O., and Rothblum, G. (2018).
\newblock Multicalibration: Calibration for the (computationally-identifiable) masses.
\newblock In {\em International Conference on Machine Learning}, pages 1939--1948. PMLR.

\bibitem[Himmelreich et~al., 2024]{himmelreich2024intersectionality}
Himmelreich, J., Hsu, A., Lum, K., and Veomett, E. (2024).
\newblock The intersectionality problem for algorithmic fairness.
\newblock {\em arXiv preprint arXiv:2411.02569}.

\bibitem[Himmelreich et~al., 2025]{himmelreich2025intersectionality}
Himmelreich, J., Hsu, A., Veomett, E., and Lum, K. (2025).
\newblock The intersectionality problem for algorithmic fairness.
\newblock In {\em Workshop on Algorithmic Fairness Through the Lens of Metrics and Evaluation}, pages 68--95. PMLR.

\bibitem[Hu et~al., 2024a]{hu2024testing}
Hu, L., Jambulapati, A., Tian, K., and Yang, C. (2024a).
\newblock Testing calibration in nearly-linear time.
\newblock In Globersons, A., Mackey, L., Belgrave, D., Fan, A., Paquet, U., Tomczak, J.~M., and Zhang, C., editors, {\em Advances in Neural Information Processing Systems 38: Annual Conference on Neural Information Processing Systems 2024, NeurIPS 2024, Vancouver, BC, Canada, December 10 - 15, 2024}.

\bibitem[Hu et~al., 2024b]{hu2024multigroup}
Hu, L., Peale, C., and Shen, J.~H. (2024b).
\newblock Multigroup robustness.
\newblock In {\em Proceedings of the 41st International Conference on Machine Learning}, ICML'24. JMLR.org.

\bibitem[Jin et~al., 2025]{jin2025discretization}
Jin, H.~H., Ding, Z., Ngo, D.~D., and Wu, Z.~S. (2025).
\newblock Discretization-free multicalibration through loss minimization over tree ensembles.
\newblock {\em arXiv preprint arXiv:2505.17435}.

\bibitem[Jung et~al., 2023]{jung2022batch}
Jung, C., Noarov, G., Ramalingam, R., and Roth, A. (2023).
\newblock Batch multivalid conformal prediction.
\newblock In {\em International Conference on Learning Representations}.

\bibitem[Kakade and Foster, 2004]{kakade2004deterministic}
Kakade, S.~M. and Foster, D.~P. (2004).
\newblock Deterministic calibration and nash equilibrium.
\newblock In {\em International Conference on Computational Learning Theory}, pages 33--48. Springer.

\bibitem[Kim et~al., 2019]{kim2019multiaccuracy}
Kim, M.~P., Ghorbani, A., and Zou, J. (2019).
\newblock Multiaccuracy: Black-box post-processing for fairness in classification.
\newblock In {\em Proceedings of the 2019 AAAI/ACM Conference on AI, Ethics, and Society}, pages 247--254.

\bibitem[Kull et~al., 2017]{pmlr-v54-kull17a}
Kull, M., Filho, T.~S., and Flach, P. (2017).
\newblock {Beta calibration: a well-founded and easily implemented improvement on logistic calibration for binary classifiers}.
\newblock In {\em Proceedings of the 20th International Conference on Artificial Intelligence and Statistics}, volume~54 of {\em Proceedings of Machine Learning Research}, pages 623--631. PMLR.

\bibitem[Kull and Flach, 2015]{kull2015novel}
Kull, M. and Flach, P. (2015).
\newblock Novel decompositions of proper scoring rules for classification: Score adjustment as precursor to calibration.
\newblock In {\em Machine Learning and Knowledge Discovery in Databases: European Conference, ECML PKDD 2015, Porto, Portugal, September 7-11, 2015, Proceedings, Part I 15}, pages 68--85. Springer.

\bibitem[Kumar et~al., 2018]{kumar2018trainable}
Kumar, A., Sarawagi, S., and Jain, U. (2018).
\newblock Trainable calibration measures for neural networks from kernel mean embeddings.
\newblock In {\em International Conference on Machine Learning}, pages 2805--2814. PMLR.

\bibitem[Liu and Wu, 2025]{liu2024multi}
Liu, T. and Wu, S. (2025).
\newblock Multi-group uncertainty quantification for long-form text generation.
\newblock In {\em The 41st Conference on Uncertainty in Artificial Intelligence}.

\bibitem[Long et~al., 2025]{long2025kernel}
Long, C.~X., Alghamdi, W., Glynn, A., Wu, Y., and Calmon, F.~P. (2025).
\newblock Kernel multiaccuracy.
\newblock In {\em 6th Symposium on Foundations of Responsible Computing (FORC 2025)}, pages 7--1. Schloss Dagstuhl--Leibniz-Zentrum f{\"u}r Informatik.

\bibitem[Marx et~al., 2023]{marx2023calibration}
Marx, C., Zalouk, S., and Ermon, S. (2023).
\newblock Calibration by distribution matching: Trainable kernel calibration metrics.
\newblock {\em Advances in Neural Information Processing Systems}, 36:25910--25928.

\bibitem[Naeini et~al., 2015]{naeini2015obtaining}
Naeini, M.~P., Cooper, G., and Hauskrecht, M. (2015).
\newblock Obtaining well calibrated probabilities using bayesian binning.
\newblock In {\em Proceedings of the AAAI conference on artificial intelligence}, volume~29.

\bibitem[Niculescu-Mizil and Caruana, 2005]{niculescu2005predicting}
Niculescu-Mizil, A. and Caruana, R. (2005).
\newblock Predicting good probabilities with supervised learning.
\newblock In {\em International Conference on Machine Learning}, pages 625--632.

\bibitem[Okoroafor et~al., 2025]{okoroafor2025near}
Okoroafor, P., Kleinberg, R., and Kim, M.~P. (2025).
\newblock Near-optimal algorithms for omniprediction.
\newblock In {\em 66th Annual Symposium on Foundations of Computer Science}.

\bibitem[Shabat et~al., 2020]{shabat2020sample}
Shabat, E., Cohen, L., and Mansour, Y. (2020).
\newblock Sample complexity of uniform convergence for multicalibration.
\newblock In {\em Advances in Neural Information Processing Systems}, volume~33, pages 13331--13340.

\bibitem[Shen et~al., 2025]{shenalgorithms}
Shen, J.~H., Vitercik, E., and Wikum, A. (2025).
\newblock Algorithms with calibrated machine learning predictions.
\newblock In {\em Forty-second International Conference on Machine Learning}.

\bibitem[Silva~Filho et~al., 2023]{silva2023classifier}
Silva~Filho, T., Song, H., Perello-Nieto, M., Santos-Rodriguez, R., Kull, M., and Flach, P. (2023).
\newblock Classifier calibration: a survey on how to assess and improve predicted class probabilities.
\newblock {\em Machine Learning}, 112(9):3211--3260.

\bibitem[USA, 1976]{eco}
USA (1976).
\newblock {\em Equal Credit Opportunity Act (15 U.S.C. 1691 et seq.)}.
\newblock Federal Trade Commission.

\bibitem[Wang, 2023]{wang2023calibration}
Wang, C. (2023).
\newblock Calibration in deep learning: A survey of the state-of-the-art.
\newblock {\em arXiv preprint arXiv:2308.01222}.

\bibitem[Wu et~al., 2024]{wu2024bridging}
Wu, J., Liu, J., Cui, P., and Wu, S.~Z. (2024).
\newblock Bridging multicalibration and out-of-distribution generalization beyond covariate shift.
\newblock {\em Advances in Neural Information Processing Systems}, 37:73036--73078.

\end{thebibliography}
\bibliographystyle{apalike}

\newpage

\appendix
\section{Bounding the size of $\calset{(\CD)}$ and $\mcalset_\CC(\CD)$}
\label{appx:bounding-num-calibrated-predictors}

\begin{lemma}
    \label{lemma:calset-finite}
    If $|\CX| = n$, the set $\calset(\CD)$ is a finite set whose size is bounded above by $B(n)$, the $n$th Bell number (OEIS A000110), which grows exponentially but slower than $n!$
    \end{lemma}
\begin{proof}
    Fix $\equiv$ to be an equivalence relation on $\CX$. Let $[x] \subseteq \CX$ be the equivalence class $x$ is in. The number of possible equivalance relations on $\CX$ is given by the Bell numbers.
    Therefore, it suffices to prove that there is at most one predictor $f \in \calset(\CD)$ such that 
    \begin{align} 
    f(x_1) = f(x_2) \iff x_1 \equiv x_2\label{eq:calibration-eq-rel} 
    \end{align}

    If there is no predictor in $\calset(\CD)$ which satisfies \Cref{eq:calibration-eq-rel}, then we are done. 
    Otherwise, assume $f$ is a calibrated predictor which satisfies \Cref{eq:calibration-eq-rel}.

    Let $g: \CX \to [0,1]$ be the following predictor.
    \begin{align*}
        g(x) = \E_{(\x, \y) \sim \CD}\Big[\y\big| \x \in [x] \Big]
    \end{align*}

    We show that $f=g$. Fix $x \in \CX$, and let $v = f(x)$. By $f$ being calibrated, we have the following equation:
    \begin{align*}
        f(x) = \E_{(\x, \y) \sim \CD}\Big[\y\big| f(\x) = v\Big]
    \end{align*}
    By \Cref{eq:calibration-eq-rel}, we have $f(\x) = v$ if and only if $\x \in [x]$. Therefore, we have the following:
    \begin{align*}
        f(x) = \E_{(\x, \y) \sim \CD}\Big[\y\big| \x \in [x]\Big]= g(x)
    \end{align*}
\end{proof}

\begin{lemma}\label{lemma:mcalset-finite}
    If the set $\CC = \{S_1, \cdots, S_k\}$ covers $\CX$, then the set $\mcalset(\CD)$ is finite and bounded by above $\prod_{i=1}^k B(|S_i|)$ where $B(\cdot)$ denotes the Bell Numbers (OEIS A000110).
\end{lemma}

\begin{proof}
    This proof has a similar structure to \Cref{lemma:calset-finite}. For each $i \in [k]$, fix equivalence relations $\equiv_i$ on the set $S_i$. Let $[x]_i \subseteq S_i$ be the equivilance class $x$ is in with respect to the equivilance relation $\equiv_i$. There are $\prod_{i=1}^k B(|S_i|)$ ways to choose these equivalence relations. It suffices to show the following:

    \emph{There is at most one predictor $f \in \mcalset(\CD)$ such that for all $i \in [k]$ and for all $x_1, x_2 \in S_i$}
    \begin{align}
        f(x_1) = f(x_2) \iff x_1 \equiv_i x_2 \label{eq:multicalibration-eq-rel}
    \end{align}

    One might want to note that this criteria on $f$ is a much stronger condition than the cooresponding criteria in \Cref{lemma:calset-finite} because it requires \Cref{eq:multicalibration-eq-rel} to be satisfied for each set.

    If there is no predictor $f \in \mcalset(\CD)$ which satisfies \Cref{eq:multicalibration-eq-rel} for all $i$, then we are done. Otherwise, assume $f$ is a multicalibrated predictor which satisfies \Cref{eq:multicalibration-eq-rel} for all $i \in [k]$.

    For each $i \in [k]$ let $g_i: \CX \to [0,1]$ be defined as follows:
    \begin{align*}
        g_i(x) = \mathbbm{1}\left(x \in S_i \right) \cdot \E_{(\x, \y) \sim \restr{\CD}{S_i}}\Big[\y| \x \in [x]_i \Big]
    \end{align*}
    Define $g: \CX \to [0,1]$ as $g(x) = \max\limits_{i \in [k]} g_i(x)$.
    It suffices to show that $f=g$.

    Fix $x \in \CX$. Let $i \in [k]$ such that $x \in S_i$ and for all $j \in [k]$, $g_i(x) \geq g_j(x)$. It is therefore the case that $g(x) = g_i(x)$. \footnote{We can actually show something stronger: For all $i \in [k]$ and for all $x \in S_i$ $g(x)=g_i(x)$, however this fact is not needed for the proof.}

    Let $v = f(x)$. By calibration on $f$ with respect to $S_i$, we have the following equation:
    \begin{align*}
        f(x) = \E_{(\x, \y) \sim \restr{\CD}{S_i}}\Big[\y| f(\x) = v\Big]
    \end{align*}
    Since $f$ satisfies \Cref{eq:multicalibration-eq-rel}, if $\x$ is sampled from $\restr{\CD}{S_i}$, then $f(\x)=v$ if and only if $\x \in [x]_i$. Therefore we have the following:
    \begin{align*}
        f(x) = \E_{(\x, \y) \sim \restr{\CD}{S_i}}\Big[\y| \x \in [x]_i\Big] = g_i(x) = g(x)
    \end{align*}
\end{proof}

\section{Deferred Proofs}

\subsection{Proof of \Cref{lemma:improving-wdmc}}
\label{appx:proof-wdMC}
\begin{proof}
    Let $\CX = \{x_1, x_2, x_3\}$ and $S_1 = \{x_1, x_2\}$, $S_2 = \{x_2, x_3\}$. Fix $\delta \in [0,0.5)$, $C \geq 6$, $D = C / 2$, and $\alpha, \beta > 0$ such that $\delta + C \cdot (\alpha + \beta) \leq 0.5$ and $\alpha + \beta \leq 2\delta / (3C)$. Take $\CD_{\x}$ as the uniform distribution over $\CX$ and \begin{align*}
        p^* = (p^*(x_i))_i &= \Big( 0.5 + \delta - C\cdot(\alpha + \beta), 0.5 - \delta, 0.5 + \delta + C\cdot (\alpha + \beta) \Big)\\
        f = (f(x_i))_i &= \Big( 0.5 - D\cdot (\alpha + \beta), 0.5, 0.5 + D\cdot (\alpha + \beta) \Big).
    \end{align*}
    By \cref{lemma:calset-finite}, $|\calset(\restr{\CD}{S_1})|, |\calset(\restr{\CD}{S_2})| \leq 2$; inspecting the partitions of $\CX$, one finds that $\calset(\restr{\CD}{S_1}) = \{\restr{p^*}{S_1}, g_1\}$ and $\calset(\restr{\CD}{S_2}) = \{\restr{p^*}{S_2}, g_2\}$ for constant maps $g : S_1 \to [0,1]$ and $g_2 : S_2 \to [0,1]$ defined by \begin{align*}g_1(x) = 0.5 - D\cdot (\alpha + \beta), \ \ g_2(x) = 0.5 + D\cdot (\alpha + \beta).\end{align*} Via direct computation, one obtains 
    
    \begin{align*}
        \Big\|\restr{f}{S_1} - \restr{p^*}{S_1}\Big\|_{1,S_1} &\geq \delta / 2,\\
        \Big\|\restr{f}{S_2} - \restr{p^*}{S_2}\Big\|_{1,S_2} &\geq \delta / 2\\
        \Big\|\restr{f}{S_1} - g_1\Big\|_{1,S_1} = \Big\|\restr{f}{S_2} - g_2\Big\|_{1,S_2} &= \frac{C}{4}(\alpha + \beta).
    \end{align*}

    Since $\alpha + \beta \leq \delta / C$, reweighting gives $\wdMC(f) = \frac{C}{6}(\alpha + \beta)$.

    We first prove that $f$ is a local minimum. Let $\gamma > 0$ (to be chosen later), and let $B_\gamma(g)$ denote the closed $\gamma$-ball (\textit{w.r.t.} $\normnoarg{1}$) around some function $g \in \R^\CX$. Consider the point $v \in B_\gamma(0)$ such that $f + v \in [0,1]^\CX$ for \begin{align*}
        f + v = \Big( 0.5 - D\cdot (\alpha + \beta) + v_1, 0.5 + v_2, 0.5 + D\cdot (\alpha + \beta) + v_3 \Big).
    \end{align*}
    We first get lower bounds on the conditional distance to $p^*$. By the triangle inequality we have, \begin{align*}
        \Big\|\restr{f}{S_i} + \restr{v}{S_i} - \restr{p^*}{S_i} \Big\|_{1,S_i}-\frac{3}{2}\|v\|_1 \leq \Big\|\restr{f}{S_i} - \restr{p^*}{S_i}\Big\|_{1,S_i} - \Big\|\restr{v}{S_i}\Big\|_{1,S_i} \leq \Big\|\restr{f}{S_i} + \restr{v}{S_i} - \restr{p^*}{S_i}\Big\|_{1,S_i}
    \end{align*}
    Since $\|\restr{f}{S_i} - \restr{p^*}{S_i}\|_1 \geq \delta/2$, it holds for small enough $\gamma$ that \begin{align*}
        \Big\|\restr{f}{S_i} + \restr{v}{S_i} - \restr{p^*}{S_i}\Big\|_{1,S_i} \geq \frac{1}{2}\delta - \frac{3}{2}\gamma \geq \frac{1}{3}{\delta}.
    \end{align*}
    Next, note that \begin{align*}
        \Big\| \restr{f}{S_1} + \restr{v}{S_1} - g_1 \Big\|_{1,S_1} &= \frac{1}{2}|v_1| + \frac{1}{2} \left|v_2 + D\cdot (\alpha + \beta)\right|\\
        \Big\|\restr{f}{S_2} + \restr{v}{S_2} - g_2\Big\|_{1,S_2} &= \frac{1}{2}|v_3|+ \frac{1}{2}\left|v_2 - D\cdot (\alpha + \beta)\right|.
    \end{align*}
    Note that for small enough $\gamma$, both quantities are less than or equal to $\delta / 3$, since $C \cdot (\alpha+\beta)/4 \leq \delta/4$ and hence, \begin{align*}
        \inf_{\substack{v \in B_\gamma(0) :\\f + v \in [0,1]^\CX}} \wdMC(f + v) = \frac{1}{3} \inf_{\substack{v \in B_\gamma(0) :\\f + v \in [0,1]^\CX}} \max \left\{ |v_1| + \left|v_2 + \frac{C}{2}(\alpha + \beta)\right|, |v_3| + \left|-v_2 + \frac{C}{2}(\alpha + \beta)\right|\right\} = \frac{C}{6}(\alpha + \beta).
    \end{align*}
    That is, we have some $\gamma > 0$ such that $\wdMC(\tilde{f}) \geq \wdMC(f)$ for all predictors in $\tilde{f} \in B_\gamma(f) \cap [0,1]^\CX$.

    Finally, assume there exists $\tilde{f} \in [0,1]^\CX$ such that $\wdMC(\tilde{f}) \leq C\alpha/6$. We will argue that $\tilde{f}$ must be close to $p^*$ and hence far from $f$. As a first case, suppose that \begin{align*}
        \frac{2}{3} \Big\|\restr{\tilde{f}}{S_1} - g_1\Big\|_{1,S_1} \leq \frac{C}{6}\alpha, \ \ \ \ \frac{2}{3} \Big\|\restr{\tilde{f}}{S_2} - \restr{p^*}{S_2}\Big\|_{1,S_2} \leq \frac{C}{6}\alpha.
    \end{align*}
    It follows that $|\tilde{f}(x_2) - g_1(x_2)|, |\tilde{f}(x_2) - p^*(x_2)|  \leq D \cdot \alpha$. Direct computation also gives $|p^*(x_2) - g_1(x_2)| = \delta - D\cdot (\alpha + \beta)$, so 
    \begin{align*}
        \delta - D(\alpha + \beta) \leq |p^*(x_2) - g_1(x_2)| \leq |\tilde{f}(x_2) - g_1(x_2)| + |\tilde{f}(x_2) - p^*(x_2)| &\leq C \cdot \alpha < C \cdot (\alpha + \beta)\\
        \delta &< 1.5 C\cdot(\alpha + \beta).
    \end{align*}
    However, this is a contradiction. This same argument works in the case where \begin{align*}
        \frac{2}{3} \Big\|\restr{\tilde{f}}{S_1} - \restr{p^*}{S_1}\Big\|_{1,S_1} \leq \frac{C}{6}\alpha, \ \ \ \ \frac{2}{3} \Big\|\restr{\tilde{f}}{S_2} - g_2\Big\|_{1,S_2} \leq \frac{C}{6}\alpha.
    \end{align*}
    
    As a third case, suppose that \begin{align*}
        \frac{2}{3} \Big\|\restr{\tilde{f}}{S_1} - g_1\Big\|_{1,S_1} \leq \frac{C}{6}\alpha, \ \ \ \ \frac{2}{3} \Big\|\restr{\tilde{f}}{S_2} - g_2\Big\|_{1,S_2} \leq \frac{C}{6}\alpha.
    \end{align*}
    Then similarly, $|\tilde{f}(x_2) - g_1(x_2)|, |\tilde{f}(x_2) - g_2(x_2)| \leq D \cdot \alpha$, so we have \begin{align*}
        C\cdot (\alpha + \beta) = |g_1(x_2) - g_2(x_2)| \leq |g_1(x_2) - \tilde{f}(x_2)| + |g_2(x_2) - \tilde{f}(x_2)| \leq C \cdot \alpha.
    \end{align*}
    This is a contradiction by the assumption that $\beta \neq 0$. It therefore must hold that \begin{align*}
        \frac{2}{3} \Big\|\restr{\tilde{f}}{S_1} - \restr{p^*}{S_1}\Big\|_{1,S_1} \leq \frac{C}{6}\alpha, \ \ \ \ \frac{2}{3} \Big\|\restr{\tilde{f}}{S_2} - \restr{p^*}{S_2}\Big\|_{1,S_2} \leq \frac{C}{6}\alpha.
    \end{align*}
    It follows immediately that $\|\tilde{f} - p^*\|_1 \leq \frac{C}{3}\alpha$. Note, however, that \begin{align*}
        \|f - p^*\|_1 = \frac{1}{3}(\delta - D\cdot (\alpha + \beta)) + \frac{1}{3}\delta + \frac{1}{3}(\delta + D\cdot (\alpha + \beta)) = \delta.
    \end{align*}
    Moreover by the triangle inequality, we have, \begin{align*}
        \|f - \tilde{f}\|_1 \geq \|f - p^*\|_1 - \|\tilde{f} - p^*\|_1 \geq \delta - \frac{C}{3}\alpha.
    \end{align*}
    By our choice of $\alpha$, we have $\|f - \tilde{f}\|_1 \geq \frac{2}{3}\delta \in \Omega(1)$. Taking $C = 6$ and $\alpha = \beta = \eps / 2$ gives the proposition.
\end{proof}

\subsection{Generalized \Cref{thm:dMC-global-minima}}
\label{appx:generalized-dmc-global-minima}

In what follows, we present a more general proof for any $\ell_p$ metric.

\begin{proposition} \label{prop:generalized-global-minima}
    Let $p \geq 1$, and $U \subset [0,1]^\CX$. Then all local minima of $\ell^p(\cdot, U) : [0,1]^\CX \to \R_{\geq 0}$ are global minima. 
\end{proposition}

\begin{proof}
    First consider some $f \in [0,1]^\CX$ such that $\metric{p}{f}{U} > 0$. We claim that $f$ cannot be a local minimum. First note that $\ell^p(f, U) = \metric{p}{f}{\on{cl}(U)}$, where $\on{cl}(U)$ denotes the closure of $U$, so we may assume without loss of generality that $U$ is closed. Next, since all finite-dimensional normed spaces are Banach and locally compact, the metric space ($\R^d$, $\normnoarg{p}$) satisfies Heine-Borel, and $U$ is compact since it is a subset of the bounded set $[0,1]^\CX$. 
    By the extreme-value theorem and continuity of $\metric{p}{f}{\cdot} : U \to \R$, there is some $u \in U$ such that $\metric{p}{f}{u} = \metric{p}{f}{U}$. Letting $t \in (0,1]$, we have that 
    \begin{align*}
        \metric{p}{(1-t)f + tu}{u} = \norm{p}{(1-t)f + tu - u} = (1-t) \cdot \norm{p}{f-u} < \metric{p}{f}{U}.
    \end{align*}
    Moreover, for any $\eps > 0$, we can choose $t$ so that $(1-t)f + tu \in B_\eps(f)$. Therefore, all local minima of $\metric{p}{\cdot}{U}$ evaluate to zero, and are hence global.
\end{proof}

\noindent Omitting proof, we note that this holds more generally for any norm-induced metric and subset $U \subset \R^\CX$. To handle this latter abstraction, one can restrict focus to an intersection $(f-U) \cap B_r(0)$ in order to invoke Heine-Borel.

\subsection{Proof of \Cref{lemma:dCE-Lipschitz}} \label{appx:dCE-Liphcitz}

\begin{proof}
    Fix a predictor $f: \CX \to [0,1]$. 
    
    Without loss of generality, assume $\dC{\CD}{p_1}(f) \leq \dC{\CD}{p_2}(f)$ It suffices to show that $\dC{\CD}{p_2}(f)-\dC{\CD}{p_1}(f) \leq \metric{1}{p_1}{p_2}$.

    By \Cref{lemma:calset-finite}, there are finitely many calibrated predictors when $p_1$ is the ground truth predictor. 
    This means there is some calibrated predictor $g:\CX \to [0,1]$ such that $\dC{\CD}{p_1}(f) = \metric{1}{f}{g}$. 
    Let $\equiv$ be an equivalence relation on $\CX$ such that $x_1 \equiv x_2$ if and only if $g(x_1)=g(x_2)$. For each $x \in \CX$, there is an equivalence class $[x]$ which is the set of all $x' \in \CX$ such that $x' \equiv x$.

    Since $g$ is calibrated with respect to the distribution $(\CD_\x, p_1)$, we have that for all $x \in \CX$:
    \begin{align*}
        g(x) = \E_{\x \sim \CD_\x}\Big[ p_1(\x) \big| g(\x) = g(x) \Big]
        = \E_{\x \sim \CD_\x}\Big[ p_1(\x) \big| \x \in [x] \Big].
    \end{align*}

    Notice that the equivalence relation $\equiv$ can be considered without $g$ since it simply partitions the space $\CX$.
    We then use $\equiv$ to construct a new predictor $h:\CX \to [0,1]$. 
    \begin{align*}
        h(x) = \E_{\x \sim \CD_\x}\Big[ p_2(\x) \big| \x \in [x] \Big]
    \end{align*}
    That is, for any $x \in \CX$, $h(x)$ is defined as the weighted average value of $p_2$ over the equivalence class $[x]$.
    To show $h$ is calibrated with respect to $p_2$, fix $x_1 \in \CX$ and let $v = h(x_1)$. Note that $h^{-1}(v)$ does not necessarily equal $[x_1]$, but we do have $[x_1] \subseteq h^{-1}(v)$.
    It is possible for there to be an $x_2 \in \CX$ such that $h(x_1)=h(x_2)=v$ despite $x_1 \not \equiv x_2$. It follows that we can find some $x_1, \cdots, x_m$ such that $h^{-1}(v) = \mathop{\sqcup}\limits_{i=1}^m [x_i]$ 
    where $\sqcup$ is the disjoint union of sets.
    Using this fact, we can show that $h$ meets the criteria for being calibrated with respect to the ground truth predictor $p_2$:
    \begin{align*}
        \E_{\x \sim \CD_\x}\Big[p_2(\x) \big| h(\x) = v\Big] &= \E_{\x \sim \CD_\x}\Big[p_2(\x) \big| \x \in h^{-1}(v) \Big]\\
        &= \sum_{i=1}^m \E_{\x \sim \CD_\x}\Big[p_2(\x) \big| \x \in [x_i] \Big] \cdot \Pr\big[[x_i]\big]/\Pr\big[h^{-1}(v)\big]\\
        &= \sum_{i=1}^m v \cdot \Pr\big[[x_i]\big]/\Pr\big[h^{-1}(v)\big]\\
        &=v
    \end{align*}

By construction of $g$ being the closest calibrated predictor to $f$, and using the fact that $h$ is a calibrated predictor on the ground truth $p_2$, we have that:
\begin{align}
\label{eq:dce-upper-bound}
    \metric{1}{f}{g} = \dC{\CD}{p_1}(f) \leq \dC{\CD}{p_2}(f) \leq \metric{1}{f}{h}.
\end{align}

Furthermore, by the triangle inequality, we have $\metric{1}{g}{h} \geq \metric{1}{f}{h}- \metric{1}{f}{g}$. 
This fact, combined with subtracting $\metric{1}{f}{g}$ from each part of \Cref{eq:dce-upper-bound}, gives us that
\begin{align*}
    0 \leq \dC{\CD}{p_2}(f) - \dC{\CD}{p_1}(f) \leq \metric{1}{f}{h} - \metric{1}{f}{g} \leq \metric{1}{g}{h}.
\end{align*}

Thus, it suffices to show $\metric{1}{g}{h} \leq \metric{1}{p_1}{p_2}$. 
\begin{align}
    \metric{1}{g}{h} &= \E_{\x_0 \sim \CD_\x}\Big[|g(\x_0)-h(\x_0)|\Big] \nonumber \\
    &= \E_{\x_0 \sim \CD_\x}\bigg[\bigg|\E_{\x \sim \CD_\x}\Big[ p_1(\x) \big| \x \in [\x_0] \Big]-\E_{\x \sim \CD_\x}\Big[ p_2(\x) \big| \x \in [\x_0] \Big]\bigg|\bigg] \nonumber\\
    &= \E_{\x_0 \sim \CD_\x}\bigg[\bigg|\E_{\x \sim \CD_\x}\Big[ p_1(\x)-p_2(\x) \big| \x \in [\x_0] \Big]\bigg|\bigg] \nonumber\\
    & \leq \E_{\x_0 \sim \CD_\x}\bigg[\E_{\x \sim \CD_\x}\Big[ |p_1(\x)-p_2(\x)| \big| \x \in [\x_0] \Big]\bigg] \nonumber\\
    &= \sum_{x_0 \in \CX} \Big( \sum_{x \in [x_0]} |p_1(x)-p_2(x)|\Pr\big[x\big]/\Pr\big[[x_0]\big]\Big) \Pr\big[x_0\big] \label{eq:dce-last-eq}
\end{align}

Notice that the inner summation only depends on the \emph{equivalence class} $[x_0]$, and not on the particular value of $x_0$ itself.
Therefore, we can group the terms in the outer summation by equivalence class, varying over all equivalence classes in $\CX$ (given by $\CX / \equiv$):

\begin{align*}
    \eqref{eq:dce-last-eq} 
    = \sum_{[x_0] \in \CX / \equiv } \Big( \sum_{x \in [x_0]} |p_1(x)-p_2(x)|\Pr\big[x\big]/\Pr\big[[x_0]\big]\Big) \Pr\big[[x_0]\big]
    = \sum_{x \in \CX} |p_1(x)-p_2(x)|\Pr\big[x\big] = \metric{1}{p_1}{p_2}.
\end{align*}

\end{proof}

\subsection{Relation Between $\wdMC$ and $\dMC$}
\label{appx:wdmc-bounded-by-dmc}
\begin{lemma}
    For any distribution $\CD \in \Delta(\CX \times \CY)$, subgroups $\CC$, and predictor $f \in \predictors$,  
    $\wdMC_{\CC}(f)\leq \dMC_{\CC}(f)$.
    \label{prop:wdmc-leq-dmc}
\end{lemma}
\begin{proof}
    Take any $S \in \CC$ such that 
        $\wdMC_{\CC}(f) = \Pr[S]\cdot \dCE_{\restr{\CD}{S}}(\restr{f}{S})$. We have the following.
    \begin{align*}
        \dMC_{\CC}(f) &=\inf_{g\in \mcalset_\CC(\CD)}\sum_{x\in \CX}[\Pr[x] \cdot|f(x)-g(x)|] \\
        &\geq \inf_{g\in\mcalset_\CC(\CD)}\sum_{x \in S}[\Pr[x]\cdot|f(x)-g(x)|] 
        \end{align*}
        For any $g \in \mcalset_{\CC}(\CD)$, $\restr{g}{S} \in \calset(\restr{\CD}{S})$. Hence,
        \begin{align*} 
        \inf_{g\in\mcalset_\CC(\CD)}\sum_{x \in S}[\Pr[x] \cdot |f(x)-g(x)|]&\geq \inf_{{h} \in \calset(\restr{\CD}{S})} \sum_{x \in S}[\Pr[x]\cdot|f(x)-{h}(x)|]\\
        &=\Pr[S] \cdot \dCE_{\restr{\CD}{S}}(\restr{f}{S})\\
        &= \wdMC_{\CC}(f).
    \end{align*}
\end{proof}

\subsection{Proof of \Cref{prop:cdMC-not-dist-to-p-star}}
\label{proof:cdMC-not-dist-to-p-star}
\begin{proof}
    We let $\CX = \{x_1, x_2, x_3, x_4\}$, and let $S_1 = \{x_1, x_2, x_3\}$ and $S_2 = \{x_2, x_3, x_4\}$. Let the marginal distribution $\CD_\x$ be uniform with $\Pr[x_i] = 1/4$ for each $i \in [4]$. We define $p^*$ as follows:
    \begin{align*}p^*(x_1)=0.3, p^*(x_2)=0.2, p^*(x_3) = 0.8,  p^*(x_4)=0.8 \end{align*}

    We define the predictor $f \neq p^*$, which is also perfectly calibrated as follows:
    \begin{align*}f(x_1)=0.3, f(x_2)=0.5, f(x_3) = 0.5,  f(x_4)=0.8\end{align*}.

    Note that $\E_{\x \sim \restr{\CD_\x}{S_1}}\Big[p^*(\x) \big| f(\x)=0.3\Big] = p^*(x_1) = 0.3$. Also, $\E_{\x \sim \restr{\CD_\x}{S_2}}\Big[p^*(\x) \big| f(\x)=0.8\Big] = p^*(x_4) = 0.8$. Finally, we have $\E_{\x \sim \restr{\CD_\x}{S_1}}\Big[p^*(\x) \big| f(\x)=0.5\Big] = \E_{\x \sim \restr{\CD_\x}{S_2}}\Big[p^*(\x) \big| f(\x)=0.5\Big]= \frac{p^*(x_2) + p^*(x_3)}{2} = 0.5$, meaning $f$ is multicalibrated. 

    Since $f$ is also calibrated with respect to $\{x_2, x_3\}=S_1 \cap S_2$, $f$ is intersection multicalibrated, meaning we have $\dME{\CC}{p^*}(f) = 0$ by \Cref{thm:dimc-continuized-dmc}.
\end{proof}

\subsection{Proof of \Cref{lemma:PIE-dIMC-definition}}
\label{appx:dIME-equivalence}
Before we prove \Cref{lemma:PIE-dIMC-definition}, we will first need some auxilary lemmas.

\begin{lemma} \label{lemma:disjoint-union-calibrated}
    If $f$ is calibrated with respect to $S_1$ and $S_2$, and $S_1$ and $S_2$ are disjoint, then $f$ is calibrated with respect to $S_1 \sqcup S_2$.
\end{lemma}
\begin{proof}
    Fix $v \in f(S_1 \sqcup S_2)$. Let $V_1 = S_1 \cap f^{-1}(v)$ and let $V_2 = S_2 \cap f^{-1}(v)$. Let $p^*$ be the ground truth predictor. If either $V_1$ or $V_2$ are empty, by $f$ being calibrated with respect to $S_1$ and $S_2$, We have $\E_{\x \sim \CD_\x}\Big[p^*(\x) \big| \x \in V_1 \sqcup V_2 \Big] =v$

    Otherwise,
    \begin{align*}
        \E_{\x \sim \CD_\x}\Big[p^*(\x) \big| \x \in V_1 \sqcup V_2 \Big]
        &= \frac{\Pr[V_1]}{\Pr[V_1 \sqcup V_2]} \cdot \sum_{x \in V_1} p^*(x) \Pr[x]/\Pr[V_1] + \frac{\Pr[V_2]}{\Pr[V_1 \sqcup V_2]} \cdot \sum_{x \in V_2} p^*(x) \Pr[x]/\Pr[V_1]
    \end{align*}
    By $f$ being calibrated with respect to $S_1$ and $S_2$, we have $\sum\limits_{x \in V_1} p^*(x) \Pr[x]/\Pr[V_1] = v = \sum\limits_{x \in V_2} p^*(x) \Pr[x]/\Pr[V_2]$. Therefore, we can conclude that $f$ is calibrated because we have $
        \E_{\x \sim \CD_\x}\Big[p^*(\x) \big| \x \in V_1 \sqcup V_2 \Big]
        = v.$
\end{proof}

\begin{lemma} \label{lemma:compliment-calibrated}
If $f$ is calibrated with respect to $S_1$ and $S_2$, and $S_1 \subseteq S_2$, then $f$ is calibrated with respect to $S_2-S_1$. 
\end{lemma}
\begin{proof}
    Fix $v \in S_2-S_1$. $V_1 = S_1 \cap f^{-1}(v)$ and let $V_2 = S_2 \cap f^{-1}(v)$. Let $p^*$ be the ground truth predictor. We have
    \begin{align*}
        v &= \E_{\x \sim \CD_\x}\Big[ p^*(\x) \big| \x \in V_2\Big]
        &= \frac{\Pr[V_1]}{\Pr[V_2]} \cdot \sum_{x \in V_1} p^*(x) \Pr[x]/\Pr[V_1] + \frac{\Pr[V_2-V_1]}{\Pr[V_2]} \cdot \sum_{x \in V_2-V_1} p^*(x) \Pr[x]/\Pr[V_2-V_1]
    \end{align*}

    By $f$ being calibrated with respect to $S_1$, we have $\sum\limits_{x \in V_1} p^*(x) \Pr[x]/\Pr[V_1] = v$. Since $\Pr[V_2]=\Pr[V_1]+\Pr[V_2-V_1]$, and $\Pr[V_2-V_1] > 0$ we can conclude with,
    \begin{align*}
        v &= \frac{\Pr[V_1]}{\Pr[V_2]} \cdot v + \frac{\Pr[V_2-V_1]}{\Pr[V_2]} \cdot \sum_{x \in V_2-V_1} p^*(x) \Pr[x]/\Pr[V_2-V_1]\\
        \frac{\Pr[V_2-V_1]}{\Pr[V_2]} \cdot v &= \frac{\Pr[V_2-V_1]}{\Pr[V_2]} \cdot \sum_{x \in V_2-V_1} p^*(x) \Pr[x]/\Pr[V_2-V_1]\\
        v &=  \sum_{x \in V_2-V_1} p^*(x) \Pr[x]/\Pr[V_2-V_1]\\
        &= \E_{\x \sim \CD_\x}\Big[p^*(\x) \big| \x \in V_2-V_1\Big]
    \end{align*}
\end{proof}

We are now ready to prove \Cref{lemma:PIE-dIMC-definition}.
\begin{proof}[Proof of \Cref{lemma:PIE-dIMC-definition}]

    We prove this by showing that $\mcalset_{\intcl(\CC)}(\CD) = \mcalset_{\CJ(\CC)}(\CD)$.

    For non-empty $I \subseteq [k]$, let $A_I = \cap_{i \in I}S_i$. Let $B_I = A_I-\left(\cup_{i \in [k]-I} S_i\right)$. It is important to note that $\CJ(\CC)$ covers $\CX$ with disjoint sets. To show this let $x \in \CX$. Let $I \subseteq [k]$ such that $i \in I$ if and only of $x \in S_i$. It follows that $x \in B_I$. Let $J \subseteq [k]$ be nonempty such that $J \neq I$. If there is some $j$ such that $j \in J$ but $j \not \in I$, then $B_J \subseteq A_J$, and $x \not \in A_J$ because $x \not \in S_j$. If there is some $j$ such that $j \in I$ but $j \not \in J$, then $x \in S_j \subseteq \cup_{i \in [k]-J} S_i$, thus $x \not \in B_J$. Therefore, the sets in $\CJ(\CC)$ are disjoint, and they cover $\CX$.

    \textbf{Part I: $\mcalset_{\CJ(\CC)}(\CD) \subseteq \mcalset_{\intcl(\CC)}(\CD)$}

    Let $g \in \mcalset_{\CJ(\CC)}(\CD)$. Let $A \in \intcl(\CC)$. Note that we can write $A$ as $A_I$ for some $I \subseteq [k]$. Also note that for all $J \supseteq I$, we have $B_J \subseteq A_J \subseteq A_I$. Also for all $J \not \supseteq I$, we have $B_J \cap S_i = \emptyset$ for $i \in I - J$. Thus $B_J \cap A_I = \emptyset$. 

    It follows from the fact that $\CJ(\CC)$ is a collection of disjoint sets which cover $\CX$ that $A_I = \bigsqcup\limits_{J \supseteq I} B_J$. Since $g$ is calibrated with respect to each $B_J$, by \Cref{lemma:disjoint-union-calibrated}, we know that $g$ is calibrated with respect to $A_I$.

    \textbf{Part II: $\mcalset_{\intcl(\CC)}(\CD) \subseteq \mcalset_{\CJ(\CC)}(\CD)$}
    
    Let $g \in \mcalset_{\intcl(\CC)}(\CD)$. To show that $g$ is calibrated for all $B_I \in \CJ(\CC)$, we do induction on $|I|$. As a base case, assume $|I| = k$. Then we must have $I = [k]$, thus $B_{[k]} = \cap_{i \in [k]}S_i = A_{[k]}$. Since $B_{[k]} \in \intcl(\CC)$, we know that $g$ is calibrated with respect to $B_I$ if $|I| = k$. 

    Fix $\ell < k$, and assume by inductive hypothesis that $g$ is calibrated with respect to $B_I$ whenever $|I| > \ell$. Let $I \subseteq [k]$ such that $|I| = \ell$. We know by definition that $g$ is calibrated with respect to $A_I$. We also know that $A_I = \bigsqcup\limits_{J \supseteq I} B_J$. By the inductive hypothesis, we know that $g$ is calibrated with respect to $B_J$ for all $J \supsetneq I$. By \Cref{lemma:compliment-calibrated}, we can subtract each $B_J$ from $A_I$, so long as $J \supsetneq I$, and we would still get a set which $g$ is calibrated with respect to. Since we have $A_I - \bigsqcup\limits_{J \supsetneq I} B_J = B_I$, it follows that $g$ is calibrated with respect to $B_I$, completing the proof by induction.
\end{proof}

\subsection{Proof of \Cref{lem:dCE-sample-guarentee}} \label{appx:dCE-auditability}
\begin{proof}
    We refer to the notion of \emph{Lower Distance to Multicalibration Error} from \citet{blasiok2022unifying}, which we denote as $\ul{\dCE}_{\restr{\CD}{S}}(f)$. By \citet[Theorem~9.10]{blasiok2022unifying}, $O(\eps^{-2})$ \emph{i.i.d.} samples from $\restr{\CD}{S}$ suffice to obtain an estimate $\hat \theta$ such that $\ul{\dCE}_\restr{\CD}{S}(f) \in [\hat \theta-\eps, \hat \theta+\eps]$ with probability at least $2/3$. 
    
    Let $\hat \theta \in [0,1]$ be the median of $m$ such estimates $\{\hat \theta_1, \cdots \hat \theta_m\}$. 
    If $\ul{\dCE}_\CD(f) \not \in [\hat \theta-\eps, \hat \theta+\eps]$, it must be the case that at least half of the values $\hat \theta_i$ do not lie within $\eps$ of $\ul{\dCE}_{\restr{\CD}{S}}(f)$. Let $\sigma_i = \mathbbm{1}(\ul{\dCE}_\restr{\CD}{S}(f) \not \in [\theta_i-\eps, \theta_i+\eps])$, and let $s = \sum_{i=1}^m \sigma_i$. Note that $\E[s] \leq m/3$. By Hoeffding, we have
    \begin{align*}
        \Pr\big[\ul{\dCE}_\restr{\CD}{S}(f) \not \in [\hat \theta-\eps, \hat \theta+\eps]\big] \leq \Pr\big[s \geq m/2\big] = \Pr\big[s-m/3 \geq m/6\big] \leq \exp(-m/18).
    \end{align*}
    For $m = O(\log(1/\delta))$, we have $\ul{\dCE}_\restr{\CD}{S}(f) \in [\hat \theta-\eps, \hat \theta+\eps]$ with probability $1-\delta$. Note that by \citet[Corollary~6.4]{blasiok2022unifying}, it holds that $\ul{\dCE}_\CD \leq \dCE_\CD \leq 4 \sqrt{\ul{\dCE}_\CD}$. It follows that with probability $1-\delta$, \begin{align*}
        \hat\theta - \eps \leq \dCE_\CC(f) \in \leq \sqrt{\hat\theta + \eps}.
    \end{align*}
\end{proof}

\subsection{Lemma for Alternative Definition of Bias}
\label{appx:wdma-proof}
\begin{lemma}
    \label{prop:dAE-alternative-def}
    $\dAE_\CD(f) = \metric{1}{f}{\accset(\CD)}$
\end{lemma}
\begin{proof}
    Fix a predictor $f$ and a ground truth predictor $p^*$. By definition, we have $\dAE_\CD(f) = \Big| \E\limits_{\x \sim \CD_\x}\Big[f(\x)-p^*(\x)\Big]\Big|$. Without loss of generality $\E\limits_{\x \sim \CD_\x}\Big[f(\x)-p^*(\x)\Big]$ is positive, thus we can remove the absolute value.

    We can partition $\CX$ into three sets: 
    \begin{align*}S_- = \{x: f(x) < p^*(x)\}; \; S_0 = \{x: f(x)=p^*(x)\}; \; S_+ = \{x: f(x) > p^*(x)\}\end{align*}

    Let $\alpha = \metricres{1}{S_+}{f}{p^*} \cdot \Pr[S_+]; \; \beta = \metricres{1}{S_-}{f}{p^*} \cdot \Pr[S_-]$
    Now we can wewrite $\dAE_\CD(f)$ as follows:
    \begin{align*}
        \dAE_\CD(f) &= \Big( \E_{\x \sim \CD_\x}\Big[|f(\x)-p^*(\x)|\big|\x \in S_+ \Big] \cdot \Pr[S_+] \Big) - \Big( \E_{\x \sim \CD_\x}\Big[|f(\x)-p^*(\x)|\big|\x \in S_- \Big] \cdot \Pr[S_-] \Big)\\
        &= \alpha - \beta
    \end{align*}

    Let $t = \beta/\alpha \in [0,1]$. Let $g: \CX \to [0,1]$ such that 
    \begin{align*}
        g(x) = \begin{Bmatrix} t \cdot f(x) + (1-t) \cdot p^*(x) & x \in S_+ \\
        f(x) & x \not \in S_+
        \end{Bmatrix}
    \end{align*}

    We have $\metricres{1}{S_+}{g}{p^*} = t \cdot \metricres{1}{S_+}{f}{p^*}$, and $\metricres{1}{S_-}{g}{p^*} = \metricres{1}{S_-}{f}{p^*}$. Therefore
    $ \dAE_\CD(g) = t \alpha - \beta = 0 $
    .

     Computing $\metric{1}{f}{g}$, we get the following:
     \begin{align*}
        \metric{1}{f}{g} &= \metricres{1}{S_+}{f}{g} \cdot \Pr[S_+]\\
        &= 
        (1-t) \cdot \metricres{1}{S_+}{f}{p^*} \cdot \Pr[S_+]\\
        &= (\alpha-\beta)/\alpha \cdot \alpha \\
        &= \dAE_\CD(f)
     \end{align*}

     Since $g \in \accset(\CD)$, we know that $\metric{1}{f}{\accset(\CD)} \leq \metric{1}{f}{g} = \dAE_\CD(f)$.

     It remains to show that for all $h \in \accset(\CD)$, we have $\dAE_\CD(f) \leq \metric{1}{f}{h}$. Let $h \in \accset(\CD)$. We have 
     \begin{align*}
     \metric{1}{f}{h} &= \E_{\x \sim \CD_\x}\Big[|f(\x) - h(\x)|\Big]\\
     &\geq \Big | \E_{\x \sim \CD_\x}\Big[f(\x) - h(\x)\Big] \Big| \\
     &= \Big | \E_{\x \sim \CD_\x}\Big[f(\x) - p^*(\x)\Big]  - \E_{\x \sim \CD_\x}\Big[p^*(\x) - h(\x)\Big] \Big|\\
     &= \Big | \E_{\x \sim \CD_\x}\Big[f(\x) - p^*(\x)\Big]  \Big|\\
     &= \dAE_\CD(f)
    \end{align*}
    
\end{proof}

\section{Inauditability of Distance to Low Degree Multicalibration and Calibrated Multiaccuracy}
\label{appx:generalizing-dMC}

In this section, we show negative results about distance to low degree multicalibration and calibrated multiaccuracy, namely that neither of these notions are Lipschitz in $p^*$.
The following definition uses the \emph{weight-function} viewpoint of multicalibration, developed in \citep{gopalan2022low}.
\begin{definition}
    Let $\polys{r}$ be the set of all polynomials of degree $<r$. A predictor is degree-$r$ multicalibrated with respect to $\CC$ if for all $w \in \polys{r}$ and for all $S \in \CC$ we have $\E\limits_{\x \sim \CD_\x}\Big[w(f(\x)) \cdot(f(\x)-p^*(\x))\big| \x \in S\Big] = 0$. Note that multiaccuracy is equivalent to degree-one multicalibration.

    Let $\mcalset_{\CC}^r(\CD)$ be the set of degree-$r$ multicalibrated predictors, and define \emph{low degree distance to multicalibration} as follows: \begin{align*}\dMC_\CC^r(f) := \inf\limits_{g \in \mcalset_\CC^r(\CD)}\metric{1}{f}{g} = \metric{1}{f}{\mcalset_\CC^r(\CD)}.\end{align*}
\end{definition}

\begin{lemma}
    \label{lem:dmc-stronger-than-low-deg} Multicalibration is a stronger notion than low degree multicalibration. That is if a predictor $f$ is multicalibrated, then it is degree-$r$ multicalibrated for all $r \in \N$.
\end{lemma}
\begin{proof}
    Assume $f$ is multicalibrated, fix a ground truth $p^*$, and let $w:[0,1] \to \R$ be any function; in particular, we can let $w \in \polys{r}$. We have the following:

    \begin{align*}
        \E_{\x \sim \CD_\x}\Big[(w(f(\x)) \cdot (f(x) - p^*(x))\big| \x \in S\Big] &= \sum_{v \in f(S)} \sum_{x \in f^{-1}(v) \cap S}\big(w(v) \cdot (v - p^*(x))\big)\frac{\Pr[x]}{\Pr[S]}\\
        &= \sum_{v \in f(S)} w(v) \cdot \sum_{x \in f^{-1}(v) \cap S} (v - p^*(x)) \frac{\Pr[x]}{\Pr[S]} \\
        &= \sum_{v \in f(S)} w(v) \cdot 0 && \text{(By $f$ being multicalibrated)}\\
        &= 0
    \end{align*}
\end{proof}

\begin{proposition}
    \label{prop:no-audit-low-deg-dmc} For all $r > 1$,  $\dMC_\CC^r(f)$ is not continuous in $p^*$.\footnote{We also remark that we do not get an analogue of \Cref{thm:dimc-continuized-dmc} for low degree multicalibration, meaning \emph{continuized} low degree multicalibration is not as attractive of a notion. However, further discussion of this issue is beyond the scope of this paper.}
\end{proposition}
\begin{proof}
    We prove that $\dMC_\CC^r$ is not Lipsctitz by giving a counterexample. Particularly, we use the exact example from \Cref{prop:no-audit-dMC}. By \Cref{lem:dmc-stronger-than-low-deg}, all multicalibrated predictors are also degree-$r$ multicalibrated. Also since $\polys{r} \subseteq \polys{s}$ for $r<s$, if we can show that in the example in \Cref{prop:no-audit-dMC}, the only degree-$2$ multicalibrated predictors are the perfectly multicalibrated predictors, then we show that for all $r > 1$, $\dMC_\CC^r(f)$ shares the same inaudibility result as $\dMC$, as shown in \Cref{prop:no-audit-dMC}.

    The following are necessary conditions for a predictor $f = (f_1, f_2, f_3)$ to be degree-$2$ multicalibrated. Specifically, these are the degree-$2$ multicalibration conditions for $w(x)=1$ and $w(x)=x$ in each set $S_1$ and $S_2$.
    \begin{align}
        f_1+f_2-1 &= 0 \label{eq:accuracy-s1} \\
        f_2+f_3-1-\alpha &= 0\label{eq:accuracy-s2}\\
        f_1\cdot (f_1 - 0.8) + f_2 \cdot (f_2 - 0.2) &= 0 \label{eq:deg-2-mcal-s1}\\
        f_2\cdot (f_2 - 0.2) + f_3 \cdot (f_3 - 0.8-\alpha) &= 0 \label{eq:deg-2-mcal-s2}
    \end{align}
    Using \Cref{eq:accuracy-s1} we can rewrite \Cref{eq:deg-2-mcal-s1} with only $f_2$. 
    \begin{align*}
        (1-f_2)\cdot(0.2-f_2) + f_2\cdot(f_2 - 0.2) &= 0\\
        2f_2^2 -1.4 f_2 + 0.2 &= 0\\
        f_2 = 0.2 \; \text{or} \; f_2 = 0.5
    \end{align*}

    If $f_2 = 0.2$, then solving for $f_1$ and $f_3$ with\Cref{eq:accuracy-s1} and \Cref{eq:accuracy-s2}, we have $f=p^*$, which is multicalibrated. 

    If $f_2 = 0.5$, then using \Cref{eq:accuracy-s2} we can rewrite \Cref{eq:deg-2-mcal-s2} with only $f_2 = 0.5$
    \begin{align*}
        0.5 \cdot 0.3 + (0.5 - \alpha)\cdot (-0.3) &= 0\\
        0.3 \alpha &= 0
    \end{align*}
    \Cref{eq:deg-2-mcal-s2} only admits a solution for $f_2 = 0.5$ when $\alpha = 0$, in which case $f=(0.5,0.5,0.5)$ is multicalibrated.

    By the fact that $f_1$ and $f_3$ are completely determined by $f_2$, as seen in \Cref{eq:accuracy-s1} and \Cref{eq:accuracy-s2}, there is at most one unique degree-$2$ multicalibrated predictors for each value of $f_2$, and since all values of $f_2$ which admitted degree-$2$ multicalibrated predictors were also perfectly multicalibrated, in \Cref{prop:no-audit-dMC}, all degree-$r$ multicalibrated predictors are perfectly multiaccurate, thus auditing distance to degree-$r$ multicalibration is discontinuous w.r.t. $p^*$ and information-theoretically hard.
\end{proof}

\begin{definition}[Distance to Calibrated Multiaccuracy]
    Fix a distribution $\CD$ with ground truth $p^*$ and groups $\CC = \{S_1 \cdots S_k\}$. Let $\on{cmac}_\CC(\CD)$ be the set of multiaccurate predictors which are also globally calibrated. We define distance to calibrated multiaccuracy as follows.
    \begin{align*}
        \on{dCMA}_{\CC, p^*}(f) := \inf_{g \in \on{cmac}_\CC(\CD)}\metric{1}{f}{g} = \metric{1}{f}{\on{cmac}_\CC(\CD)}
    \end{align*}
\end{definition}

\begin{proposition}
    \label{prop:no-audit-dCMA} $\on{dCMA}_{\CC, p^*}(f)$ is not continuous in $p^*$.
\end{proposition}
\begin{proof}
 We construct a counterexample. Let $\CX = \{x_1 \cdots x_6\}$. Let $\CC = \{S_1, S_2, S_3\}$ where $S_1 = \{x_1, x_2, x_3\}$, $S_2 = \{x_3, x_4, x_5\}$, $S_3 = \{x_1, x_5, x_6\}$. Let $\CD_\CX$ be the uniform distribution where for all $x$ in $\CX$, we have $\Pr[x] = 1/6$. 
    Define $f$, the predictor we will audit as follows:
    \begin{align*}
        f(x_1)=f(x_3)=f(x_5)&=0.6\\
        f(x_2)=f(x_4)=f(x_6)&=0.3
    \end{align*}
    Let $p^*(x)$ be defined as follows: 
    \begin{align*}
        p^*(x_1) = 0.6&; p^*(x_2) = 0.2; p^*(x_3) = 0.7\\
        p^*(x_4) = 0.3&; p^*(x_5) = 0.5; p^*(x_6) = 0.4
        \end{align*}
    Notice that $f$ is multi-calibrated, and for each $i \in [3]$, we have $\E\limits_{\x \sim \CD_\x}\Big[p^*(x) \big| x \in S_i\Big] = \E\limits_{\x \sim \CD_\x}\Big[f(x) \big| x \in S_i\Big]= 0.5$, meaning $f$ is multi-accurate, thus we have $\on{dCMA}_{\CC, p^*}(f) = 0$

    For some irrational $\eps > 0$, let $q^*$ be defined as follows \footnote{This construction also works if $\eps \in \Q$, which can be less-easily shown by checking at most $203$ calibrated predictors.} 
    \begin{align*}
        q^*(x_i) = \begin{cases}
            0.2-\eps & \text{ if } i = 2\\
            p^*(x_i) & \text{ if } i \neq 2
        \end{cases}.
    \end{align*}
    Notice that as we take $\eps \to 0$, $q^*$ gets arbitratily close to $p^*$. We show $\on{dCMA}_{\CC, q^*}(f) > C$ for some constant $C \geq \frac{1}{60}$.

    Let $g$ be a calibrated multiaccurate predictor with respect to the ground truth $q^*$. Let $v = g(x_2)$. By the fact that $g$ is calibrated, we have \begin{align*}v = \E\limits_{\x \sim \CD_\x}\Big[q^*(\x)\big| \x \in g^{-1}(v)\Big] = \frac{1}{|g^{-1}(v)|}\sum\limits_{x \in g^{-1}(v)}q^*(x)\end{align*}

    Assume, for the sake of contradiction, that $g^{-1}(v) \not \subseteq S_1$. By the assumption that $g$ is unbiased on $S_1$, we have
    \begin{align}1.5-\eps = \sum\limits_{x \in S_1-g^{-1}(v)}q^*(x) + |g^{-1}(v) \cap S_1| \cdot v. \label{eq:cmac-rationality-contradiction} \end{align} 
    For all $x \in \CX - g^{-1}(v)$, we know $q^*(x)$ is rational, thus there is some $a \in \Q$ such that $a = \sum\limits_{x \in S_1-g^{-1}(v)}q^*(x)$. By $g$ being calibrated, we have $v = \frac{1}{|g^{-1}(v)|}\sum\limits_{x \in g^{-1}(v)}q^*(x)$. Since $q^*(x)$ is rational for all $x \neq x_2$, we have $v = \frac{-\eps}{|g^{-1}(v)|} + b$ for some $b \in \Q$. Since we assume $g^{-1}(v) \not \subseteq S_1$, but $x_2 \in g^{-1}(v)$ we have $0 < |g^{-1}(v) \cap S| < |g^{-1}(v)|$. Let $c = \frac{|g^{-1}(v) \cap S|}{|g^{-1}(v)|}$. Note $c \in \Q$ and $0 < c < 1$.
    Therefore we can rewrite \Cref{eq:cmac-rationality-contradiction} as follows
    \begin{align*}
        1.5-\eps &= a+|g^{-1}(v) \cap S| \cdot b - c \eps\\
        1.5-a-|g^{-1}(v) \cap S| \cdot b &= (1-c)\eps
    \end{align*}
    Which is a contradiction because the left hand side is rational and the right hand side is not.

    Therefore, it must be the case that $g^{-1}(v) \subseteq S_1$. Then there are only four possibilities for what $g^{-1}(v)$ could equal, $\{x_2\}$, $\{x_1, x_2\}$, $\{x_2, \x_3\}$, or $\{x_1, x_2, x_3\}$. In the first case, $v=0.2-\eps$, meaning $\metric{1}{f}{g} \geq |f(x_2)-g(x_2)|/6 > 0.1/6 = \frac{1}{60}$. In the second case, $v = g(x_1)=g(x_2) = 0.4-\frac{\eps}{2}$, in which case $\metric{1}{f}{g} \geq |f(x_1)-g(x_1)|/6 > 0.2/6 = \frac{1}{30}$. In the third case, $v = g(x_2)=g(x_3) = 0.45-\frac{\eps}{2}$, in which case $\metric{1}{f}{g} \geq |f(x_3)-g(x_3)|/6 > 0.25/6 = \frac{1}{24}$. Finally, in the fourth case, we have $v = g(x_1)=g(x_2)=g(x_3) = 0.5-\frac{\eps}{3}$ in which case $\metric{1}{f}{g} \geq (|f(x_1)-g(x_1)|+|f(x_3)-g(x_3)|)/6 > 0.3/6 = \frac{1}{20}$. Therefore, we have $\on{dCMA}_{\CC, q^*}(f) > \frac{1}{60}$.

\end{proof}

\end{document}